\documentclass[lettersize,journal]{IEEEtran}
\usepackage{amsmath,amsfonts}
\usepackage{algorithmic}
\usepackage{algorithm}
\usepackage{array}
\usepackage[caption=false,font=normalsize,labelfont=sf,textfont=sf]{subfig}
\usepackage{textcomp}
\usepackage{stfloats}
\usepackage{url}
\usepackage{verbatim}
\usepackage{graphicx}
\usepackage{cite}
\usepackage{subfiles}
\usepackage{graphicx}
\usepackage{amsmath}
\usepackage{amssymb}
\usepackage{multirow}
\usepackage{makecell}
\usepackage{color}
\usepackage{bm}
\usepackage[pagebackref,breaklinks,colorlinks]{hyperref}
\usepackage{marvosym}
\usepackage{xcolor}
\usepackage{booktabs}
\usepackage{colortbl}
\usepackage{threeparttable}

\usepackage[capitalize]{cleveref}
\crefname{section}{Sec.}{Secs.}
\Crefname{section}{Section}{Sections}
\Crefname{table}{Table}{Tables}
\crefname{table}{Tab.}{Tabs.}

\newcommand{\FIG}[1]{\cref{#1}}
\newcommand{\TABLE}[1]{\cref{#1}}
\newcommand{\EQUA}[1]{\cref{#1}}
\newcommand{\SECTION}[1]{\cref{#1}}

\newcommand{\MODEL}{Unnamed Model}

\newcommand{\xmark}{$\times$}
\newcommand{\xm}{\xmark}
\newcommand{\cm}{\checkmark}

\newcommand{\ETAL}{{\emph{et al.}}}

\newcommand{\IE}{{\emph{i.e.}}}

\newcommand{\SIMAUGCITE}{\cite{liang2020simaug}}

\newcommand{\GARDENCITE}{\cite{liang2020garden}}

\newcommand{\TRAJECTRONPP}{Trajectron++}

\newcommand{\TRAJECTRONPPCITE}{\cite{salzmann2020trajectron}}
\newcommand{\YNET}{$\textsf{Y}$-net}
\newcommand{\YNETCITE}{\cite{mangalam2020s}}

\newcommand{\SPECTGNN}{SpecTGNN}
\newcommand{\SPECTGNNCITE}{\cite{cao2021spectral}}

\newcommand{\AGENTFORMER}{AgentFormer}
\newcommand{\AGENTFORMERCITE}{\cite{yuan2021agentformer}}

\newcommand{\MSN}{MSN}
\newcommand{\MSNCITE}{\cite{wong2021msn}}

\newcommand{\MID}{MID}
\newcommand{\MIDCITE}{\cite{gu2022stochastic}}
\newcommand{\SHENET}{SHENet}
\newcommand{\SHENETCITE}{\cite{meng2022forecasting}}

\newcommand{\MUSEVAE}{MUSE-VAE}
\newcommand{\MUSEVAECITE}{\cite{lee2022muse}}

\newcommand{\EVMODEL}{E-V$^2$-Net}

\newcommand{\EVCITE}{\cite{wong2023another}}

\newcommand{\FLOWCHAIN}{FlowChain}
\newcommand{\FLOWCHAINCITE}{\cite{maeda2023fast}}
\newcommand{\IMP}{IMP}
\newcommand{\IMPCITE}{\cite{shi2023representing}}
\newcommand{\SCMODEL}{SocialCircle}
\newcommand{\SCCITE}{\cite{wong2023socialcircle}}
\newcommand{\MSRL}{MSRL}
\newcommand{\MSRLCITE}{\cite{wu2023multi}}

\newcommand{\LGTRAJ}{LG-Traj}
\newcommand{\LGTRAJCITE}{\cite{chib2024lg}}
\newcommand{\RAN}{RAN}
\newcommand{\RANCITE}{\cite{dong2024recurrent}}

\newcommand{\UPDD}{UPDD}
\newcommand{\UPDDCITE}{\cite{liu2024uncertainty}}
\newcommand{\SCPMODEL}{SocialCircle+}
\newcommand{\SCPCITE}{\cite{wong2024socialcircle+}}
\newcommand{\UEN}{UEN}
\newcommand{\UENCITE}{\cite{su2024unified}}
\newcommand{\MSTIP}{MS-TIP}
\newcommand{\MSTIPCITE}{\cite{chib2024ms}}
\newcommand{\SMEMO}{SMEMO}
\newcommand{\SMEMOCITE}{\cite{marchetti2024smemo}}
\newcommand{\DICE}{Dice}
\newcommand{\DICECITE}{\cite{choi2024dice}}
\newcommand{\PPT}{PPT}
\newcommand{\PPTCITE}{\cite{lin2024progressive}}
\newcommand{\AFFLN}{AgentFormer-FLN}
\newcommand{\AFFLNCITE}{\cite{xu2024adapting}}
\newcommand{\SOPERMODEL}{SoperModel}
\newcommand{\SOPERMODELCITE}{\cite{yang2025sopermodel}}
\newcommand{\REMODEL}{\emph{Resonance}}
\newcommand{\RECITE}{\cite{wong2024resonance}}

\renewcommand{\MODEL}{\emph{Reverberation}}
\newcommand{\MODELSHORT}{\emph{Rev}}
\newcommand{\APPENDIX}[1]{{Appendix {#1}}}

\newcommand{\C}[1]{{\textcolor[HTML]{0000FF}{#1}}}
\newcommand{\CC}[1]{{\underline{\C{#1}}}}

\newcommand{\TODO}[1]{\colorbox{yellow}{TODO}}
\newcommand{\hatbf}[1]{\hat{\mathbf{#1}}}

\newcommand{\NEWHALFLINE}{\vspace{0.5em}\noindent}

\newcommand{\LIN}{\mathrm{lin}}
\newcommand{\NON}{\mathrm{non}}
\newcommand{\SOC}{\mathrm{soc}}

\begin{document}

\title{
    Reverberation: Learning the Latencies Before Forecasting Trajectories
}

\author{
    Conghao Wong,
    Ziqian Zou,
    Beihao Xia,
    and~Xinge You,~\IEEEmembership{Senior Member,~IEEE}
    \thanks{
        This work was supported by the National Natural Science Foundation of China under Grant 62172177, 62575116.
        \emph{(Corresponding author: Xinge You.)}
    }
    \thanks{
        Conghao Wong, Ziqian Zou, and Beihao Xia are with the School of Electronic Information and Communications, Huazhong University of Science and Technology, Wuhan 430074, China
        (Email: conghaowong@icloud.com,
        ziqianzoulive@icloud.com,
        xbh\_hust@hust.edu.cn).
    }
    \thanks{
        Xinge You is with the School of Electronic Information and Communications, Huazhong University of Science and Technology, Wuhan 430074, China, and also with National Anti-Counterfeit Engineering Research Center, Wuhan 430074, China
        (Email: youxg@mail.hust.edu.cn).
    }
    \thanks{
        Code is available at \url{https://github.com/cocoon2wong/Rev}.
    }
}

\markboth{Journal of \LaTeX\ Class Files,~Vol.~14, No.~8, August~2021}%
{Shell \MakeLowercase{\textit{et al.}}: A Sample Article Using IEEEtran.cls for IEEE Journals}


\maketitle

\begin{abstract}

    Bridging the past to the future, connecting agents both spatially and temporally, lies at the core of the trajectory prediction task.
    Despite great efforts, it remains challenging to explicitly learn and predict \emph{latencies}, \IE, response intervals or temporal delays with which agents respond to various trajectory-changing events and adjust their future paths, whether on their own or interactively.
    Different agents may exhibit distinct latency preferences for noticing, processing, and reacting to a specific trajectory-changing event.
    The lack of consideration of such latencies may undermine the temporal continuity of forecasting systems, leading to implausible or unintended trajectories.
    Inspired by reverberation in acoustics, we propose a new reverberation transform and the corresponding \MODEL~(\MODELSHORT~for short) trajectory prediction model, which predicts both individual latency preferences and their stochastic variations accordingly, by using two explicit and learnable reverberation kernels, enabling latency-conditioned and controllable trajectory prediction under both non-interactive and social latencies.
    Experiments on multiple datasets, whether of pedestrians or vehicles, demonstrate that \MODELSHORT~achieves competitive accuracy while revealing interpretable latency dynamics across agents and scenarios.
    Qualitative analyses further verify the properties of the reverberation transform, highlighting its potential as a general latency modeling approach.

\end{abstract}



\section{Introduction}
\label{sec_introduction}

\emph{To live is to imagine ourselves in the future, and there we inevitably arrive.}
Human cognition and planning fundamentally rely on anticipating future outcomes from past experiences.
Similarly, trajectory prediction seeks to forecast future positions for all agents based on their observed temporal patterns and potential interactions \cite{alahi2016social}.
It can be used and applied in various intelligent systems and applications, including but not limited to behavior analyses \cite{zhang2023pedestrian,yue2023human,zhang2023bip}, transportation systems \cite{golchoubian2023pedestrian,wong2021msn,zhou2024edge}, multi-agent decision systems \cite{wang2022stepwise,guo2022pedestrian,liu2023stagp}, thus attracting increasing attention from more and more researchers.

As one of the time-series forecasting tasks, the modeling of temporal dependencies and relationships \cite{mao2019learning,pei2019human} between every observation and prediction step has been widely researched in the field of trajectory prediction.
Prior researchers tried to capture temporal dependencies within trajectories using recurrent networks like RNNs \cite{jain2016structural} and LSTMs \cite{alahi2016social,gupta2018social}, discussing how the next step would be determined by the current state of an agent.
Further, advanced neural networks like Transformers \cite{yu2020spatio,giuliari2020transformer}, diffusion models \cite{gu2022stochastic,mao2023leapfrog}, and even knowledge transferred from large language models \cite{bae2024can} have been introduced to this task to further explore how to bridge past observations to the future, not only simply from the time domain but additionally considering the spectral properties \cite{wong2022view,wong2024resonance} of trajectories.

\begin{figure}[tbp]
    \centering
    \includegraphics[width=1.0\linewidth]{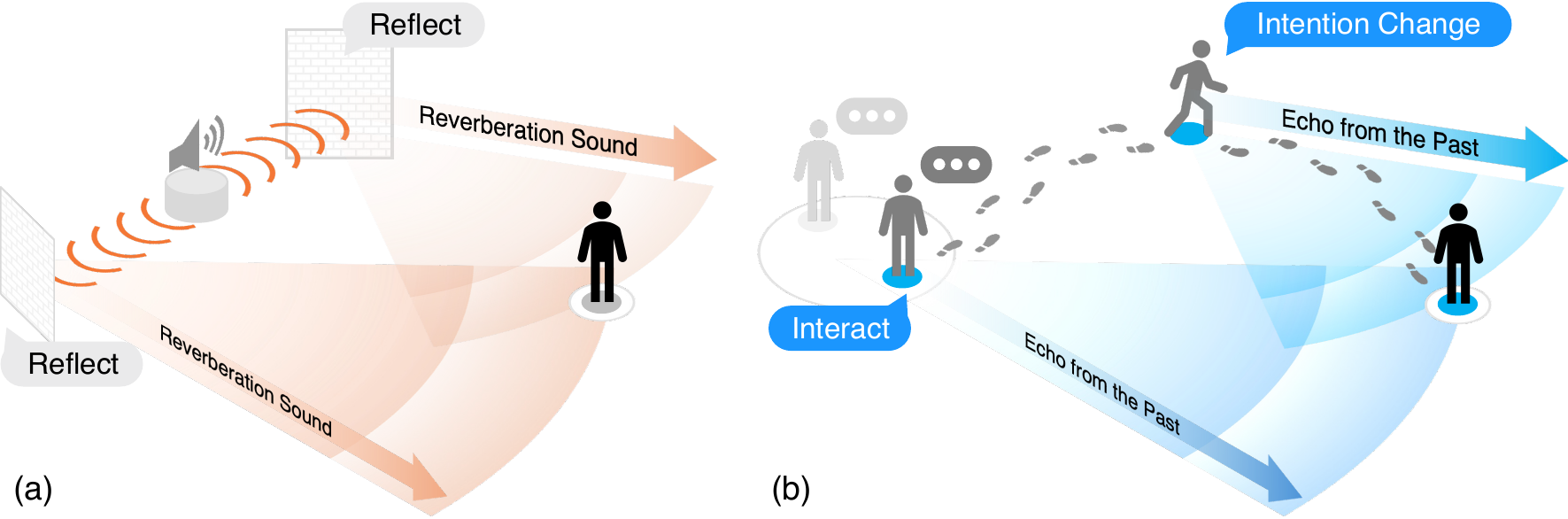}
    \caption{
        Motivation illustration.
        Inspired by the acoustic reverberation that describes how the echo sound decays in the space after reflecting, we model how past trajectory-changing events (like intention changes or interactions) change trajectories as well as their latencies as ``echoes from the past''.
    }
    \label{fig_intro}
\end{figure}

Despite great efforts, directly instructing neural networks to analyze and identify potential trajectory-changing events within observed trajectories and to predict the impact of these events at some specific future steps remains a challenging problem.
For example, it might be hard for these methods to answer questions such as how long it takes for an event to begin affecting the trajectory, how long such an effect persists, and when its influence will dissipate.
However, for the agents we are concerned with, whether pedestrians, vehicles, or robots, this is quite intuitive.
For instance, we always subconsciously calculate these intervals: how long after an event (whether self-sourced or socially interactive) occurs before it will be noticed by us, how long we wait before initiating a response and altering our trajectory, and finally how long it takes for us to finish modifying trajectories correspondingly.
While current methods can bridge the past and the future to some extent, like capturing interactions or intention changes then making the corresponding trajectory modifications when forecasting, the span and precise location of these ``bridges'' remain unknown.

Formally, existing forecasting models mainly assume instantaneous propagation of effects induced by trajectory-changing events.
This assumption has been structurally embedded in RNN recurrence, Transformer attention, and diffusion step-wise denoising, limiting their ability to model latency as an independent variable in the forecasting process.
In other words, although these methods have been able to characterize the temporal dependence between different observed and predicted moments, they still do not explicitly model the temporal \textbf{\emph{latencies}} between an event's occurrence and the moment its influence begins to shape the future trajectory.
This prevents the predictions from fully aligning with temporal continuity and real-world human intuition, leading to unexpected outcomes and diminished interpretability.
Specifically, in this work, we define trajectory latencies\footnote{
    We have provided comparisons among concepts like latencies, dependencies, temporal attention, and temporal convolution in \APPENDIX{E}.
} as temporal offsets between the steps where trajectory-changing events are encoded and the steps where their influence peaks or vanishes for the future trajectories, which differs from conventional temporal dependence that focuses primarily on correlations rather than temporal directions or temporal mappings.
Also, these latencies measure the temporal properties of how each agent's future trajectories will be modified as responses to these events.
In other words, they describe the temporal response characteristics of agents' trajectories, rather than agents' real-world reaction times or behavioral delays that could be mostly unobservable and hard to label.
Therefore, explicitly modeling and predicting these temporal latencies, \IE, temporal offsets for responding to different events, and then conditioning the forecasting process on them, becomes our central focus.

\emph{Though the future is a product of every present that precedes it, tomorrow does not belong to today.}
As we described above, building bridges from the past to the future is a centerpiece of the trajectory prediction task.
However, these bridges are more likely to be \emph{drawbridges}, since such connections between specific past and future could be cut off as time goes by.
Therefore, it remains challenging for current models to not only clearly locate these bridges, but estimate when they will close in the future.
Here, the first problem we have to address is that the past and the future are not equivalent because of the causality \cite{granger1969investigating,heise1975causal} of the system, where only events occurring in the past can affect the future, and this effect is accompanied by a specific latency.
Similar to our intuitive experience, the future (\emph{tomorrow}) is a product of the past (\emph{present}), but such production presents specific temporal properties, or more accurately, specific time-frequency responses.

In the field of signal processing, the Fourier transform and its variations provide valuable insights into describing such time-frequency properties.
Just as the Fourier transform decomposes a signal into the superposition of a number of sinusoids and analyzes its spectral properties by different superposition coefficients, a natural thought is that we can decompose a future trajectory, or at least its representation, into the superposition of the representations of past events, and simulate latencies within trajectories by the corresponding superposition coefficients.
Specifically, these superposition coefficients will be made trainable, thus further unlocking the potential of advanced neural networks for learning and simulating latencies in these trajectories in different scenarios.
As with reverberation in acoustics (illustrated in \FIG{fig_intro} (a)), which describes the intensity decay of a sound wave as it is reflected and reaches a specific destination, we describe future trajectories as superpositions of ``echo ripples'' from the past (\FIG{fig_intro} (b)), thus directly modeling the latencies for agents that respond to different trajectory-changing events.
We call this transform the \emph{Reverberation Transform}, which explicitly represents the latencies within the trajectory to respond to different trajectory-changing events at any past moment, and combines the representation of future trajectories as a shifted superposition of such past representations.

Specifically, inspired by the matrix form of the Fourier transform, we use two learnable coefficient matrices $\{\mathbf{R}, \mathbf{G}\}$, referred to as reverberation kernels, to further decouple latencies into an event-level latency basis and a set of agent-specific variation modes.
Denote the observation and prediction lengths as $T_h$ and $T_f$.
The transform operates on the $T_h \times T_h$ similarity matrix of encoded observations for each feature dimension, which captures pairwise temporal relationships among observed events.
First, the Reverberation Kernel $\mathbf{R} \in \mathbb{R}^{T_h \times T_f}$ captures the temporal latency response and provides the event-level latency basis.
The similarity matrix is right-multiplied by kernel $\mathbf{R}$, thus decomposing trajectory representations on all future $T_f$ steps as a combination of $T_h$ observed (past) features.
The second kernel $\mathbf{G}$ is introduced to further capture agent-dependent variability of latency preferences.
Similar to previous multi-style approaches \cite{wong2021msn}, the Generating Kernel $\mathbf{G} \in \mathbb{R}^{T_h \times K_g}$ generates $K_g$ variation modes over the single latency basis.
Correspondingly, the above representation will be further left-multiplied by $\mathbf{G}^\top$ to simultaneously generate diverse latency variations conditioned on the latency basis provided by kernel $\mathbf{R}$.
With the reverberation transform, each agent's unique latency preference will be predicted by kernels, thus explicitly enabling the prediction of latencies for different trajectory-changing events, and further using these latencies as a prior to enable interpretable trajectory forecasting.

In summary, our main contributions are as follows:
\begin{itemize}
    \item
    We propose the reverberation transform to explicitly model agents' latencies in handling trajectory-changing events, consisting of two trainable reverberation kernels: the reverberation kernel $\mathbf{R}$ that models event-level latency responses and the generating kernel $\mathbf{G}$ that simulates variations and alternatives of such latencies.

    \item
    We design the \MODEL~(\MODELSHORT~for short) trajectory prediction model, which builds upon the proposed reverberation transform, to simultaneously learn non-interactive and social latencies, thereby enabling latency-aware and interpretable trajectory prediction.

    \item
    We validate the proposed \MODELSHORT~both quantitatively and qualitatively on multiple trajectory datasets, and extensively analyze the theoretical properties of the proposed reverberation transform.
    Extensive ablations and discussions further verify both the quantitative improvements and the interpretability of the learned latency dynamics.
\end{itemize}

\section{Related Works}

\subsection{Trajectory Prediction and Social Interactions}

Trajectory prediction aims at predicting agents' future movements based on their observed states and potential interactions \cite{alahi2016social}.
Researchers have increasingly focused attention on the modeling of social interactions when forecasting.
Early works mainly use handcrafted rules to simulate interactions as forces among agents or environments.
Social Force \cite{helbing1995social} regards pedestrian motion as a system of attractive and repulsive forces, while the velocity obstacle \cite{fiorini1998motion} and reciprocal velocity obstacle \cite{van2008reciprocal} frameworks model collision avoidance via geometric velocity constraints.
With the development of data-based methods, sequential models like RNNs \cite{kim2017probabilistic,sun2020recursive} and LSTMs \cite{alahi2016social,zhang2019sr,zhang2020social} are then introduced to capture implicit dynamic interactions among agents.
Social-LSTM \cite{alahi2016social} first uses social pooling to model the target agent's local interactions with its surroundings, inspiring diverse follow-up works \cite{deo2018convolutional,pei2019human,liu2023stagp} in the same social-interaction-modeling fashion.
Researchers further leverage attention mechanisms \cite{vemula2018social,fernando2018soft} or Transformers \cite{giuliari2020transformer,yuan2021agentformer,zhao2020tnt}, as well as structured information in GNNs \cite{ivanovic2019trajectron,cao2020spectral,mohamed2020social,li2020dynamic,sang2025review}, to globally capture complex interactions and group behaviors among agents.
Furthermore, recent advancements in GNNs have introduced deformable spatial-temporal modeling \cite{chen2025dstigcn}, multiscale hypergraphs \cite{sang2025msdhgnn}, and physically constrained higher-order graphs \cite{chen2025pchgcn} to further filter redundant interactions and handle complex dynamics.

Although countless great ideas or works have been proposed, it is still difficult for current approaches to cover all potential connections or interactions between intelligent agents, especially in more and more complex scenarios.
Some researchers have recently introduced more interpretable theories to simulate such interactions from a more human-centric point of view.
Inspired by echolocation of marine animals, Wong \ETAL \cite{wong2023socialcircle,wong2024socialcircle+} use angle-based representations to model social interactions with surroundings by considering their velocity, distance, and direction.
Further, considering the knowledge embedded within language models and their potential for drawing interactions, Bae \ETAL \cite{bae2025social,bae2024can} introduce large language models with specially designed numeric tokens to reveal the spatial interactions from a brand-new view.
Emulating the resonance phenomenon, \REMODEL~\cite{wong2024resonance} views social interactions as co-vibrations of agents' spectrums, thus decomposing and interpreting randomness among agents when interacting.
GPCC \cite{zou2024who} also takes group preferences of pedestrians into account, using the long-term kernel function to capture multi-level group interactions.
Despite their remarkable efforts, most current models neglect the consideration of latencies for noticing, handling, or finally terminating any specific social events over the trajectory-decision process, leading to the underestimation of interactions temporally.

\subsection{Temporal Dependencies and Latencies in Trajectories}

Apart from social interactions, temporal dependencies in trajectories have been widely studied in the field of trajectory prediction.
Researchers try to capture these dependencies through diverse theoretical approaches to deep neural networks, from the Social Force \cite{helbing1995social} to recurrent structures like RNNs \cite{jain2016structural,kim2017probabilistic,martinez2017human,sun2020recursive} or LSTMs \cite{alahi2016social,xue2018ss,manh2018scene,bisagno2018group}, or further attention-based networks like Transformers \cite{yu2020spatio,ngiam2021scene,zhou2022hivt,shi2023trajectory,zhou2024edge}, and more recently, diffusion models \cite{gu2022stochastic,rempe2023trace,bae2024singulartrajectory,li2024bcdiff,choi2024dice} and knowledge transferred from large language models \cite{bae2025social,bae2024can}.

These models could effectively bridge agents' representations from the observation period to the prediction period, yet they lack explicit mechanisms to model potential temporal latencies.
For example, each pedestrian may present different latency preferences for starting to notice something, starting to process an event, and starting to modify their trajectories.
Note that this latency is not exactly equivalent to time dependence.
Temporal dependence indicates how future representations will be represented through the current moment, whereas latency indicates how long after an interval the effects of a particular event will be apparent.
Unfortunately, most of these current methods cannot predict this latency relationship.

Our considered latencies only describe the temporal offsets of how agents' trajectories are or will be finally modified in response to different events.
Notably, the wider concepts of temporal offsets or temporal delays have already been studied in the fields of signal processing and control systems, such as delay compensation \cite{krstic2009delay,cortes2011delay}.
Human behavioral studies \cite{grice1982human,bradshaw2002dissociation} also reveal diverse kinds of reaction times, \IE, diverse biological behavioral preferences or delays, in the form of temporal offsets.
Nevertheless, most current trajectory prediction approaches still assume instantaneous propagation of events and do not account for our considered latencies as explicit temporal offsets when forecasting, not to mention agents' different biological reaction times or behavioral delays.
By introducing the new reverberation transform, the proposed \MODELSHORT~model explicitly introduces learnable latency controls, thus extending trajectory prediction networks to be conditioned by such temporal response properties inherent in such latencies.

~\\
In summary, existing trajectory prediction models focus more on the modeling of spatial interactions or implicit temporal dependencies, without explicitly modeling when and how long past events influence agents' future motions.
Prior temporal models are better at capturing sequential dependencies within trajectories but not latencies for agents to plan their trajectories.
Our work differs by explicitly formulating and learning latencies as trainable temporal responses, implemented through the proposed Reverberation Transform and the \MODELSHORT~trajectory prediction model, making it possible to take into account latencies while forecasting trajectories.

\section{Method}

\subsection{Reverberation Transform}

The foundation of the proposed \MODEL~(\MODELSHORT)~model is the reverberation transform.
Like the acoustic reverberation curves, it is designed to describe how agents would instinctively modify their trajectory plans in anticipation of the upcoming prediction period, according to the trajectory-changing events they saw or have already experienced during observation.
Here, the reverberation time indicates how long the ego will finally notice a trajectory-changing event and start to modify its own trajectory plan, as well as how long such modification will last or vanish in the future.
These times could be all considered as the \textbf{latencies} to handle existing or possible trajectory-changing events for changing their own trajectory plans, depending on their unique behavioral preferences.

\begin{figure}[tbp]
    \centering
    \includegraphics[width=1.0\linewidth]{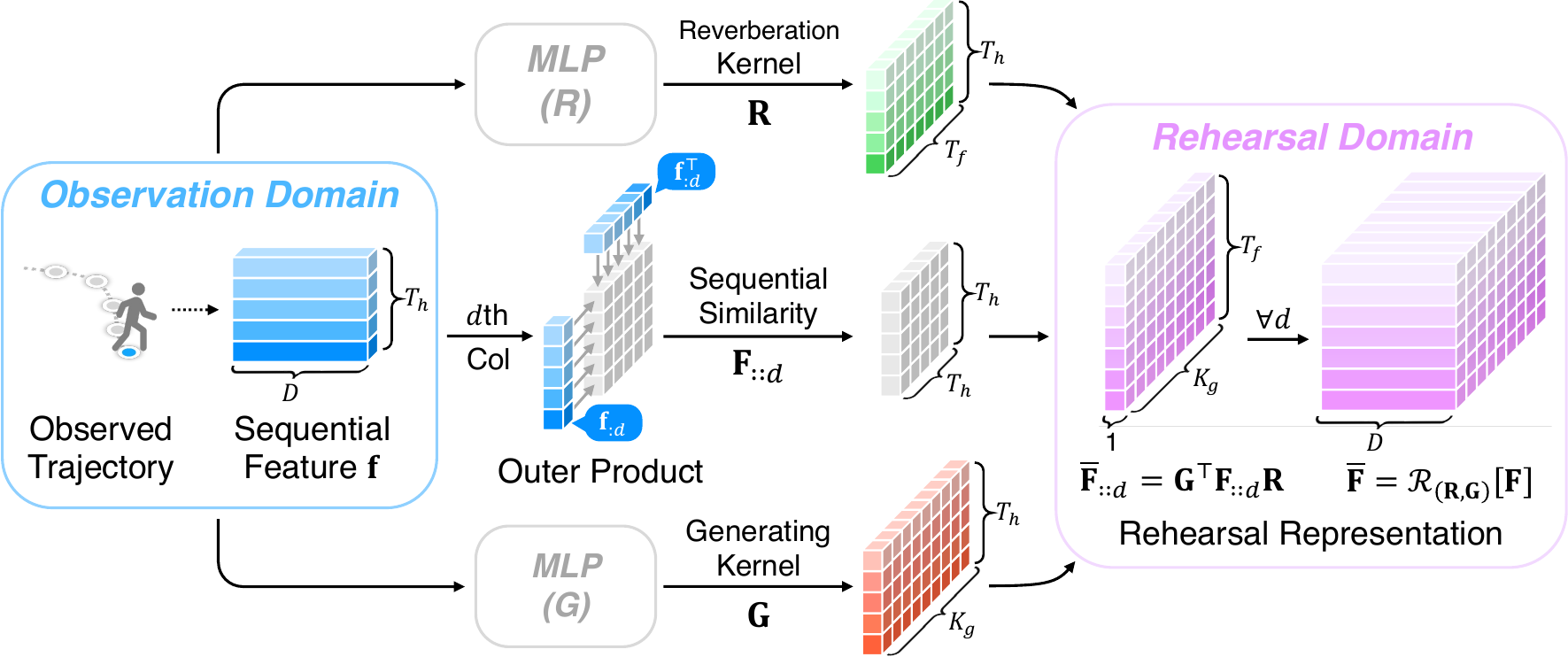}
    \caption{
        Computation pipelines of the proposed reverberation transform.
        It aims at learning two reverberation kernels $\mathbf{R}$ and $\mathbf{G}$, thus mapping sequential feature $\mathbf{f}$ from the observation domain into the imagined rehearsal domain. 
    }
    \label{fig_transform}
\end{figure}

The main focus of the reverberation transform is to learn and forecast such latencies when agents plan their trajectories.
Intuitively, people tend to imagine those upcoming events in their minds before they actually happen, as if they were doing a \emph{dress rehearsal} before the live show.
These trajectory plans are hypothesized and may not exactly be the same as their real future paths.
Thus, as shown in \FIG{fig_transform}, this transform aims at connecting ego agents' observations (the \emph{observation} domain) and their immediate planning of future trajectories (the \emph{rehearsal} domain), reflecting the instinctive latency of how ego agents respond to different trajectory determinants.

\NEWHALFLINE
\subsubsection{Events}
To extend the applicability of the reverberation transform, we do not assign specific physical meanings or further classifications to trajectory-changing events when defining this transform.
We use the term \emph{events} to refer to those idealized input events (\IE, the \emph{excitation} to a system), which could finally change how the system performs and responds after their occurrence.
These events are represented with the temporal-sequence form, and they may last for either a single temporal step or live across multiple ones.
We use representation $\mathbf{f} \in \mathbb{R}^{T_h \times D}$ to denote such events as observed sequential input (on $T_h$ past steps, $D$ is the feature dimension) to the system.
Here, the \textbf{minimum unit} for distinguishing and analyzing these events has become one single step.

\NEWHALFLINE
\subsubsection{Sequential Similarity}
We first construct the sequence-wise similarity $\mathbf{F} = \sigma(\mathbf{f}) \in \mathbb{R}^{T_h \times T_h \times D}$ to characterize differences among temporal steps, enhancing feature-level differences among those events.
For the $d$th feature dimension ($d = 1, 2, ..., D$), we have its slice
\begin{equation}
    \mathbf{F}_{::d} = \sigma(\mathbf{f}_{:d}) :=  \mathbf{f}_{:d} \mathbf{f}_{:d}^\top.
\end{equation}
This helps emphasize the differences among temporal steps within the sequence, thereby highlighting salient observations or distinctive events, allowing subsequent operations to focus on the most informative events rather than on uniform feature patterns when modeling latencies.

\NEWHALFLINE
\subsubsection{Reverberation Kernels}
The reverberation transform considers how the system responds to any events from two perspectives: the latencies of agents in handling events or modifying trajectories, and the randomness or variations within such latencies.
Its basic thought is that the system's future responses could be directly represented by the linear combinations of its historical states.
Such combination coefficients may present specific temporal-spectral properties, playing a role similar to the time-variant \emph{unit impulse responses}\footnote{
    See interpretations from the system-theory view in \APPENDIX{B}.
} to characterize this system's response properties.
Especially, we regard that latencies for handling any events could also be represented by such combination coefficients explicitly rather than indirectly, just like Fourier transforms decompose signals as direct combinations of sinusoids.

For each agent, this combination depends on a trainable \emph{Reverberation Kernel} $\mathbf{R} \in \mathbb{R}^{T_h \times T_f}$ and a trainable \emph{Generating Kernel} $\mathbf{G} \in \mathbb{R}^{T_h \times K_g}$.
We collectively refer to them as \emph{Reverberation Kernels}.
Elements in both reverberation kernels are limited to the range of $[-1, 1]$ via the $\tanh$ activation, limiting the range of transformed features while forcing the network to learn and distinguish these two kernels by their unique structures rather than through massive data-fitting.
Also, we regard that different agents may have different reverberation kernels, thus reflecting their preference differences.
 
A reverberation transform, denoted as $\mathcal{R}_{(\mathbf{R}, \mathbf{G})}[\cdot]$, transforms similarity $\mathbf{F}$ from the \emph{observation} domain (corresponding to $T_h$ past steps) into the \emph{rehearsal} domain (plan-previews for future $T_f$ steps).
For the $d$th feature dimension, the transformed slice $\overline{\mathbf{F}}_{::d} \in \mathbb{R}^{K_g \times T_f}$ is computed as the linear combination
\begin{equation}
    \label{eq_rev_transform}
    \overline{\mathbf{F}}_{::d} = 
        \mathcal{R}_{(\mathbf{R}, \mathbf{G})} \left[
            \mathbf{F}_{::d}
        \right] :=
        \mathbf{G}^\top \mathbf{F}_{::d} \mathbf{R}.
\end{equation}
For clarity, the tensor slicing operator $_{::d}$ will be omitted hereafter, while the matrix multiplications for all feature dimensions will default to being broadcasted.

\NEWHALFLINE
\subsubsection{Interpretations of Reverberation Kernels}
Reverberation transform can be directly interpreted using the trainable reverberation kernels.
The reverberation kernel $\mathbf{R}$ characterizes event-level latencies, while the generating kernel $\mathbf{G}$ models global latency preferences and their stochastic variations.
Together, these kernels can span a representation space for the input events, enabling an \emph{approximate completeness}\footnote{
    See discussions of their theoretical properties in \APPENDIX{A}.
} for simulating latencies when planning trajectories.
Correspondingly, this transform can be separated into two steps:

(a) The reverberation kernel $\mathbf{R}$ first linearly combines columns in each slice $\mathbf{F}_{::d}$ to directly establish connections from every observation (spectral) step $t_\mathrm{obs} \in \{1, ..., T_h\}$ to every future $t_\mathrm{fut} \in \{T_h + 1, ..., T_h + T_f\}$.
Each element $\mathbf{R}_{uv}$ represents how events at an observed step $t_\mathrm{obs} = u$ affect or contribute to the future step $t_\mathrm{fut} = T_h + v$, reflecting how the planned time-sensitive events are sourced during observation and scheduled after specific latencies.
This formally enforces temporal directions and temporal consistencies between observations and future trajectory changes, making sure agents' future steps are all sourced from their former states.
Also, each row in kernel $\mathbf{R}$ is supposed to be computed individually, without further mixings or normalizations, representing how the system would directly and separately respond to each input event at a specific observation step.
For any feature dimension $d$, it outputs a slice $\mathbf{F}_{::d}\mathbf{R} \in \mathbb{R}^{T_h \times T_f}$, which shares the same temporal scales and sequential length as the final predicted trajectory, without further recurrent computations, meaning that the reverberation transform does not introduce additional sequential decoding or recurrent unrolling.

(b) The generating kernel $\mathbf{G}$ re-weights each row in the slice $(\mathbf{F}_{::d}\mathbf{R})$ in $K_g$ ways, simulating $K_g$ different ways to handle such latencies in the rehearsal domain.
This is inspired by the previous work MSN \MSNCITE~that employs multiple style channels when forecasting, thus generating better stochastic trajectories.
Similarly, the generating kernel $\mathbf{G}$ provides more alternatives for combining all observed events with different latency preferences.
Unlike the style-channel-based MSN, such re-weighting through the explicit kernel decomposition may not significantly diversify trajectory-level outputs, but provide more ways to simultaneously take into account latencies of past events.
Formally, each of the $K_g$ rows in $\mathbf{G}$ corresponds to a distinct latency-handling mode, which can be treated as minor but characteristic modifications on the reverberation kernel $\mathbf{R}$ to \emph{alter} latency preferences.
For any feature dimension, this step outputs the slice $\overline{\mathbf{F}}_{::d} \in \mathbb{R}^{K_g \times T_f}$.

Overall, the reverberation transform is a feature-level mapping $\mathcal{R}_{(\mathbf{R}, \mathbf{G})}: \mathbb{R}^{T_h \times T_h \times D} \to \mathbb{R}^{K_g \times T_f \times D}$.
It maps the observed sequential representation $\mathbf{f}$ (or similarity $\mathbf{F} = \sigma(\mathbf{f})$) into the rehearsal domain to simulate agents' instant preview of their scheduled plans in the following $T_f$ steps by explicitly taking into account latencies of different input events, also their $K_g$ potential variations of combinations.
This design enables explicit modeling of heterogeneous latency preferences, which distinguishes the reverberation transform from existing feature-aggregation or style-based methods.

\subsection{The \MODEL~(\MODELSHORT) Trajectory Prediction Model}

\begin{figure*}[tbp]
    \centering
    \includegraphics[width=1.0\linewidth]{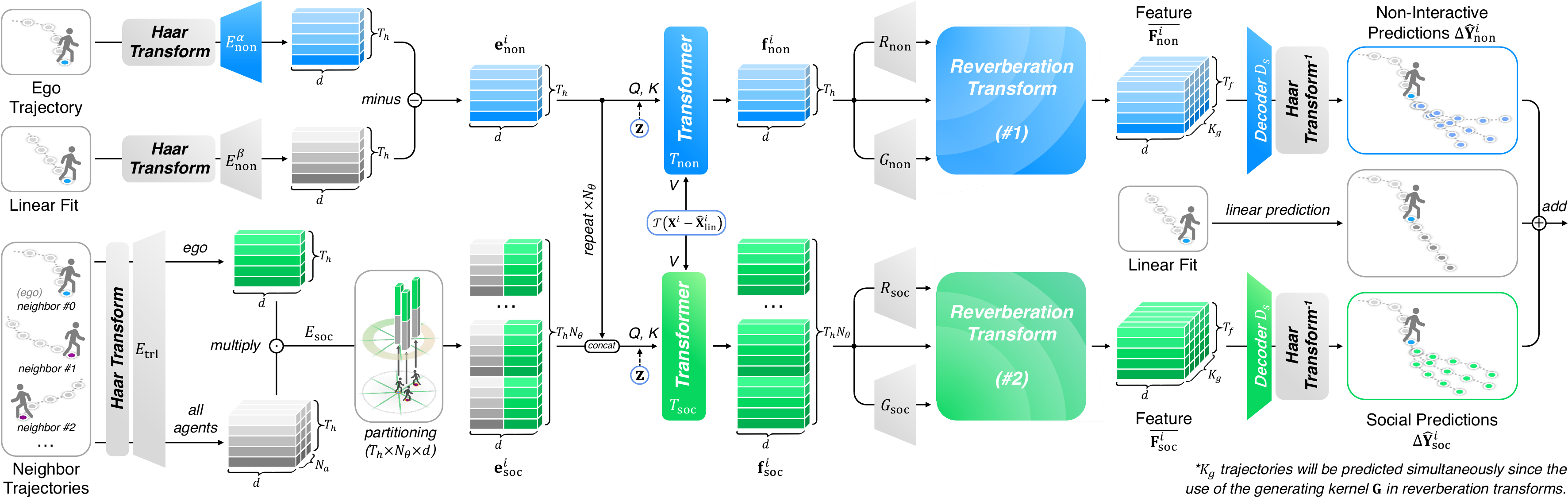}
    \caption{
        Overview of the proposed \MODEL~(\MODELSHORT~for short) trajectory prediction model.
        It predicts trajectories as two learnable parts for any ego $i$, the non-interactive $\Delta \hatbf{Y}^i_\NON$ and the social $\Delta \hatbf{Y}^i_\SOC$, with specific non-interactive latencies and social latencies, added upon the linear prediction $\hatbf{Y}^i_\LIN$ as reference.
    }
    \label{fig_method}
\end{figure*}

\begin{table}[tbp]
    \caption{
        Symbols and Network Structures of the \MODELSHORT~Model
    }
    \label{tab_symbol}
    \centering
    \begin{tabular}{cllc}
        \toprule
        Sym. & Descriptions & Shapes \\

        \midrule
        $\Delta t$          & Temporal interval to sample trajectories          & - \\
        $\mathbf{p}^i_t$    & Agent-$i$'s 2D position at time step $t$          & $(m:=2)$          \\
        $\mathbf{X}^i$      & $i$'s observed trajectory ($t_h$ past steps)      & $(t_h, m)$        \\
        $\mathbf{Y}^i$      & $i$'s groundtruth trajectory ($t_f$ future steps) & $(t_f, m)$        \\
        
        \midrule
        $d$         & Feature dimensions in most networks   & - \\
        $N_\theta$  & The number of social partitions       & - \\
        $K_g$       & The number of columns in generating kernels & - \\
        $\hatbf{Y}^i_R$     & \MODELSHORT~predictions ($K_g$ generations each)  & $(K_g, t_f, m)$ \\

        \bottomrule
    \end{tabular}

\end{table}

\begin{table}[tbp]
    \caption{
        Network Structures and Computations of the \MODELSHORT~Model
    }
    \label{tab_structures}
    \centering
    \begin{tabular}{cllc}
        \toprule
        Symbols & Descriptions \\

        \midrule
        $\mathcal{T}(\cdot)$    & Haar transform, $(t_h, m) \to (T_h, M)$ \\
        $\mathcal{T}^{-1}(\cdot)$ & Inverse Haar transform, $(T_f, M) \to (t_f, m)$ \\
        $T_\NON$ & Transformer (4 encoder-decoder layers, 8 heads) \\
        $T_\SOC$ & Transformer (2 encoder-decoder layers, 8 heads) \\
        $E_\NON^{\alpha (\beta)}$ / $E_\mathrm{trl}$ & $\to \mathrm{fc}(d, \mathrm{ReLU}) \to \mathrm{fc}(d, \tanh) $ \\
        $E_\SOC$                & $\to \mathrm{fc}(d, \mathrm{ReLU}) \to \mathrm{fc}(d, \mathrm{ReLU}) \to \mathrm{fc}(d/2, \mathrm{ReLU})$ \\
        $R_\NON$ or $R_\SOC$    & $\to \mathrm{fc}(d, \mathrm{ReLU}) \to \mathrm{fc}(d, \mathrm{ReLU}) \to \mathrm{fc}(T_f, \tanh)$ \\
        $G_\NON$ or $G_\SOC$    & $\to \mathrm{fc}(d, \mathrm{ReLU}) \to \mathrm{fc}(d, \mathrm{ReLU}) \to \mathrm{fc}(K_g, \tanh)$ \\
        $D_\NON$ or $D_\SOC$    & $\to \mathrm{fc}(M, \mathrm{none})$\\

        \bottomrule
    \end{tabular}

\end{table}

\textbf{Formulations.}
Building upon the reverberation transform, we now formulate the complete \MODEL~(\MODELSHORT~for short) trajectory prediction model.
Our main focus is to learn and forecast agents' latencies to handle trajectory-changing events when forecasting their trajectories, whether non-interactive or social events.
Denote the 2D position (we use $m = 2$ to represent this dimensionality) of any $i$th ($i \in \{1, 2, ..., N_a\}$) agent at a time step $t$ as $\mathbf{p}^i_t = (x^i_t, y^i_t)^\top \in \mathbb{R}^m$.
Our focused trajectory prediction aims at forecasting possible future trajectory(ies) $\hatbf{Y}^i = (\hatbf{p}^i_{t_h + 1}, ..., \hatbf{p}^i_{t_h + t_f})^\top \in \mathbb{R}^{t_f \times m}$ on $t_f$ discrete future steps, according to the observed trajectories of all these agents $\mathcal{X} = \{\mathbf{X}^i = (\mathbf{p}^i_1, ..., \mathbf{p}^i_{t_h})^\top \in \mathbb{R}^{t_h \times m} | i = 1, 2, ..., N_a\}$ on the currently observed $t_h$ steps (all temporal steps are sampled with the interval $\Delta t$).
Formally, it aims at constructing the network $N$, such that $\{\hatbf{Y}^i\} = N (\mathcal{X})$.
Summaries of symbols and network structures are listed in \TABLE{tab_symbol,tab_structures}.

\textbf{Method Details.}
Inspired by the decomposition in \REMODEL~\RECITE, \MODELSHORT~predicts trajectories as the linear superposition of three decoupled parts: linear, non-interactive, and social trajectories, thereby representing distinct self-sourced or social latency patterns separately.
This decomposition is a structural consequence of the linearity of the reverberation transform (\APPENDIX{A}), which guarantees that temporal responses from different kinds of events could superpose without introducing cross-component coupling during forecasting.
As illustrated in \FIG{fig_method}, for any ego agent $i$, prediction $\hatbf{Y}^i_R$ is forecasted by
\begin{equation}
    \hatbf{Y}^i_R = \hatbf{Y}^i_\LIN + \Delta \hatbf{Y}^i_\NON + \Delta \hatbf{Y}^i_\SOC.
\end{equation}
Note that due to the generating kernel $\mathbf{G}$, \MODELSHORT~will predict $K_g$ stochastic predictions in parallel for each ego agent.

\NEWHALFLINE
\subsubsection{Linear Predictions}
Inspired by the \REMODEL~model \cite{wong2024resonance}, the uniform (linear) movement of ego agents without external forces will be considered as a reference for considering and modeling either non-interactive or social latencies and corresponding predictions.
Such \textbf{linear fit} $\hatbf{X}^i_\LIN \in \mathbb{R}^{t_h \times m}$ and \textbf{linear prediction} $\hatbf{Y}^i_\LIN \in \mathbb{R}^{t_f \times m}$ are computed by
\begin{align}
    \hatbf{X}^i_\LIN
        &= \mathbf{A}_h \mathbf{w}^i_\LIN
        = \left(\begin{matrix}
            1 & 1 & ... & 1 \\
            1 & 2 & ... & t_h \\
        \end{matrix}\right)^\top \mathbf{w}^i_\LIN, \\
    \hatbf{Y}^i_\LIN
        &= \mathbf{A}_f \mathbf{w}^i_\LIN
        = \left(\begin{matrix}
            1 & ... & 1 \\
            t_h + 1 & ... & t_h + t_f \\
        \end{matrix}\right)^\top \mathbf{w}^i_\LIN.
\end{align}
Here, $\mathbf{w}^i_\LIN \in \mathbb{R}^{2 \times m}$ is the linear weight matrix, solved by
\begin{equation}
    \mathbf{w}^i_\LIN = \left(
        \mathbf{A}^\top_h \mathbf{A}_h
    \right)^{-1} \mathbf{A}^\top_h \mathbf{X}^i.
\end{equation}

\NEWHALFLINE
\subsubsection{Non-interactive Latencies}
We now focus on the latencies of (ego) agents themselves to handle their own trajectory-changing events, including the intention changes or random behaviors during observation.
We first construct representations of such non-interactive parts of any ego-$i$'s observed trajectory $\mathbf{X}^i$.
This residual term describes how agents' observed trajectories differ from their initial plans (linear movements) after self-sourced trajectory-changing events.
Given two mirrored MLPs $E_\NON^\alpha$ and $E_\NON^\beta$ (structures in \TABLE{tab_structures}), we have
\begin{equation}
    \mathbf{e}^i_\NON = \frac{1}{2} \left[
        E_\NON^\alpha \left(\mathcal{T}[\mathbf{X}^i]\right) -
        E_\NON^\beta \left(\mathcal{T}[\hatbf{X}^i_\LIN]\right)
    \right].
\end{equation}
Here, $\mathcal{T}$ denotes the Haar transform that maps trajectories into the frequency domain, since previous work \RECITE~has indicated its capability in characterizing sequences from the time-frequency joint distribution perspective, which may help us model latency preferences.
Furthermore, we adopt the Haar transform for its superior capability in capturing high-frequency signal variations with the minimum compact support, which enables the finest localization of trajectory-changing events both temporally and spectrally, as analyzed and verified in \EVMODEL\EVCITE\footnote{
    See further discussions of trajectory transforms in \APPENDIX{D}.
}.
Considering that Haar spectrums have different sequence lengths from the original sequences, we denote the shape of Haar spectrums with capitalized symbols, \IE, the mapping of shapes $(t_h, m) \to (T_h := t_h/2, M := 2m)$.
Denote the feature dimension as $d$, we have $\mathbf{e}^i_\NON \in \mathbb{R}^{T_h \times d}$.
Therefore, the \textbf{minimum unit} for learning, encoding, and processing these self-sourced trajectory-changing events has become each sequential step in the Haar spectrum, \IE, each temporal-spectral step, corresponding to 2 adjacent real-world observation frames in the time domain.

A Transformer \cite{vaswani2017attention} backbone $T_\NON$ will be used to establish relations between different spectral steps within the embedded $\mathbf{e}^i_\NON$.
Here, $\mathbf{e}^i_\NON$ itself acts as both keys and queries in Transformer encoder and decoder, while the differential spectrum $\mathcal{T}[\mathbf{X}^i - \hatbf{X}^i_\LIN]$ is used as value sequences in Transformer decoder.
Noise $\mathbf{z}_\NON \sim \mathbf{N}(\mathbf{0}, \mathbf{I})$ will then be sampled to simulate random choices.
It finally encodes a feature $\mathbf{f}^i_\NON \in \mathbb{R}^{T_h \times d}$:
\begin{equation}
    \label{eq_transformer_non}
    \mathbf{f}^i_\NON = T_\NON \left(
        \mathrm{Concat}\left(
            \mathbf{e}^i_\NON,
            \mathbf{z}_\NON
        \right),
        \mathcal{T}[\mathbf{X}^i - \hatbf{X}^i_\LIN]
    \right).
\end{equation}

Then, the reverberation transform $\mathcal{R}_{(\mathbf{R}^i_\NON, \mathbf{G}^i_\NON)}$ will be applied to transform feature $\mathbf{f}^i_\NON$ into the rehearsal domain, denoted as $\overline{\mathbf{F}^i}_\NON$, with two trainable reverberation kernels $\mathbf{R}_\NON^i$ and $\mathbf{G}_\NON^i$ assigned for each agent.
This reverberation transform is a feature-level map $\mathbb{R}^{T_h \times T_h \times d} \to \mathbb{R}^{K_g \times T_f \times d}$.
Given two MLPs $R_\NON$ and $G_\NON$ (see \TABLE{tab_structures}), we have
\begin{gather}
    \mathbf{R}_\NON^i = R_\NON \left(
        \mathbf{f}^i_\NON
    \right), \quad
    \mathbf{G}_\NON^i = G_\NON \left(
        \mathbf{f}^i_\NON
    \right), \\
    \label{eq_selfkernels1}
    \overline{\mathbf{F}^i_\NON} = 
        \mathcal{R}_{(\mathbf{R}^i_\NON, \mathbf{G}^i_\NON)}
        \left[
            \sigma \left(\mathbf{f}^i_\NON\right)
        \right] \in \mathbb{R}^{K_g \times T_f \times d}.
\end{gather}
Notably, both such MLPs treat the temporal dimension ($T_h$) in $\mathbf{f}^i_\NON$ as the ``data batch'', which ensures that different rows in these kernels are computed separately without further temporal mixings, therefore making these reverberation kernels learn unique system responses (more specifically \emph{unit impulse responses}) to events at different observation steps.

The final \textbf{non-interactive prediction} $\Delta \hatbf{Y}^i_\NON$ is inferred from such rehearsal representation $\overline{\mathbf{F}^i_\NON}$ via a simple decoder MLP $D_\NON$ (an FC layer without activations, see \TABLE{tab_structures}).
After applying the inverse Haar transform $\mathcal{T}^{-1}$, we have
\begin{equation}
    \Delta \hatbf{Y}^i_\NON = \mathcal{T}^{-1} \left[
        D_\NON \left(
            \overline{\mathbf{F}^i_\NON}
        \right)
    \right] \in \mathbb{R}^{K_g \times t_f \times m}.
\end{equation}
This process could be treated as the learning of potential biases between the scheduled trajectories during rehearsal and the actual trajectories observed after the prediction period.
These trajectories share the same temporal scale, and no additional extrapolations are required during forecasting.

It can also be seen that Transformer backbones only compute temporal features over the observation period, rather than both observation and prediction periods.
Compared to former works that utilize recurrent networks or Transformers to directly compute features from the observation to the whole prediction period, the reverberation transform not only straightforwardly provides a ``buffer pool'' (the rehearsal representations) that aligns the prediction period, but also keeps the continuity during training and testing while avoiding further recurrent computations.
Most importantly, the latencies with which agents react to different trajectory determinants, as well as their randomness, could still be captured and learned by two trainable reverberation kernels.
This is why the rehearsal domain and the reverberation transform have been introduced.

\NEWHALFLINE
\subsubsection{Social Latencies}
Our considered social latencies indicate how agents schedule their responses to social behaviors, like how long it takes them to start noticing, to handle, and to finally finish a social event.
The corresponding social prediction is obtained using a similar pipeline as the non-interactive branch above.
The only difference is that representations of social interactions will be considered during forecasting.

Different from non-interactive latencies, it is in line with our intuition that agents may present distinctive latency preferences for handling social events that occur in different \emph{angular orientations}.
Inspired by works \cite{wong2024resonance,wong2024socialcircle+} that represent social interactions in an angle-based space, we use a similar social-interaction-modeling approach as proposed in \REMODEL~\RECITE~to represent the context of social interactions when forecasting.
It divides the whole $2\pi$ angle into $N_\theta$ angle-based social partitions, and computes the average dot product between the features of each neighbor and the ego within every partition to represent social interactions.
For any agent $u$, its feature is first embedded into the $\mathbf{e}^u_\mathrm{trl} \in \mathbb{R}^{T_h \times d}$ through the translated trajectory $\mathbf{X}^u - \mathbf{p}^u_{t_h}$, via MLP $E_\mathrm{trl}$ in \TABLE{tab_structures}:
\begin{equation}
    \mathbf{e}^u_\mathrm{trl} = E_{\mathrm{trl}} \left(
        \mathcal{T}\left[
            \mathbf{X}^u - \mathbf{p}^u_{t_h}
        \right]
    \right).
\end{equation}
Then, for the ego $i$, considering a neighbor $j$, their interactions are encoded and represented as spectral similarity \RECITE
\begin{equation}
    \mathbf{e}^{i \leftarrow j}_\SOC = E_\SOC \left(
        \mathbf{e}^i_\mathrm{trl} \odot \mathbf{e}^j_\mathrm{trl}
    \right),
\end{equation}
where $\odot$ denotes the element-wise multiply ($\mathbf{e}^{i \leftarrow j}_\SOC \in \mathbb{R}^{T_h \times d}$).

After that, all ego-neighbor-pairs' representations will be averaged within each social partition.
The following computations are the same as those used in the \REMODEL~model \RECITE, by replacing their ``$\mathbf{f}^{i \leftarrow j}$'' with the above ``$\mathbf{e}^{i \leftarrow j}_\SOC$'' in their original Eqs. (11) - (12).
Similar to their resonance matrix, we have our social representation $\mathbf{e}^i_\SOC \in \mathbb{R}^{T_h \times N_\theta \times d}$.

Compared to the original computations \RECITE, the main difference is that $\mathbf{e}^i_\SOC$ maintains both temporal-spectral ($T_h$ historical steps) and spatial ($N_\theta$ angular positions) information of all social events during observation, which ensures the locating and modeling of latencies to handle social events at the partition-level dynamically.
In other words, the \textbf{minimum unit} for handling social trajectory-changing events has become one partition at each temporal-spectral step.
This ensures that the reverberation kernels could account for angle-specific social interactions and their latencies, like different delays induced by neighbors in front versus behind the ego agent.

Then, social representation $\mathbf{e}_\SOC^i$ and the former embedded $\mathbf{e}_\NON^i$ will be further encoded together using another Transformer backbone $T_\SOC$.
To make such two representations compatible, the non-interactive $\mathbf{e}_\NON^i$ ($T_h \times d$) will be repeated for $N_\theta$ times at its first dimension, turning it into the new representation $\tilde{\mathbf{e}}_\NON^i \in \mathbb{R}^{T_h N_\theta \times d}$.
Meanwhile, the first two dimensions of the social representation $\mathbf{e}_\SOC^i$ ($T_h \times N_\theta \times d$) will be flattened, turning into the compatible $\tilde{\mathbf{e}}_\SOC^i \in \mathbb{R}^{T_h N_\theta \times d}$.
Like \EQUA{eq_transformer_non}, we then sample noise $\mathbf{z}_\SOC \sim \mathbf{N}(\mathbf{0}, \mathbf{I})$, and
\begin{equation}
    \mathbf{f}^i_\SOC = T_\SOC \left(
        \mathrm{Concat}\left(
            \tilde{\mathbf{e}}^i_\NON,
            \tilde{\mathbf{e}}^i_\SOC,
            \mathbf{z}_\SOC
        \right),
        \mathcal{T}[\mathbf{X}^i - \hatbf{X}^i_\LIN]
    \right).
\end{equation}

Similar to \EQUA{eq_selfkernels1}, reverberation transform $\mathcal{R}_{(\mathbf{R}^i_\SOC, \mathbf{G}^i_\SOC)}$ is then applied on feature $\mathbf{f}^i_\SOC$ along with its two trainable reverberation kernels $\mathbf{R}_\SOC^i$ and $\mathbf{G}_\SOC^i$.
Differently, this reverberation transform is a map: $\mathbb{R}^{T_h N_\theta \times T_h N_\theta \times d} \to \mathbb{R}^{K_g \times T_f \times d}$, aiming at locating social events both temporally and spatially (\IE, $T_h N_\theta$ social events in total) and then learning their latencies.
These kernels are inferred from MLPs $R_\SOC$ and $G_\SOC$ (\TABLE{tab_structures}):
\begin{gather}
    \label{eq_socialkernels2}
    \mathbf{R}_\SOC^i = R_\SOC \left(
        \mathbf{f}^i_\SOC
    \right), \quad
    \mathbf{G}_\SOC^i = G_\SOC \left(
        \mathbf{f}^i_\SOC
    \right), \\
    \label{eq_socialkernels1}
    \overline{\mathbf{F}^i_\SOC} = 
        \mathcal{R}_{(\mathbf{R}^i_\SOC, \mathbf{G}^i_\SOC)}
        \left[
            \sigma (\mathbf{f}^i_\SOC)
        \right] \in \mathbb{R}^{K_g \times T_f \times d}.
\end{gather}
Given the decoder $D_\SOC$ (same structured as $D_\NON$), the batch ($K_g$ predictions) of \textbf{social prediction} $\Delta \hatbf{Y}^i_\SOC$ is forecasted by
\begin{equation}
    \Delta \hatbf{Y}^i_\SOC = \mathcal{T}^{-1} \left[
        D_\SOC \left(\overline{\mathbf{F}^i_\SOC}\right)
    \right] \in \mathbb{R}^{K_g \times t_f \times m}.
\end{equation}

\NEWHALFLINE
\subsubsection{Loss Function}
Similar to previous works \MSNCITE, to forecast better trajectories while maintaining its generalizability, \MODELSHORT~model is trained end-to-end with the \emph{best-of-$K_g$} $\ell_2$ loss.
For the ego agent $i$, considering the batch of predictions ($\hatbf{Y}^i$ is output during a single forward pass and noise-sampling, containing $K_g$ stochastic trajectories in total) and the ground truth trajectory $\mathbf{Y}^i$, this loss function $L$ is computed as
\begin{equation}
    L \left(\hatbf{Y}^i, \mathbf{Y}^i\right) = 
        \min_{1 \leq k \leq K_g} \frac{1}{t_f} \left\|
            \hatbf{Y}^i_{k::} - \mathbf{Y}^i
        \right\|_2.
\end{equation}

\section{Experiments}

\subsection{Experimental Settings}

\subsubsection{Datasets}
We use three public trajectory datasets\footnote{
    We also use a motion (human skeleton) prediction dataset to verify the generalizability of \MODELSHORT~beyond trajectory prediction.
    See \APPENDIX{F}.
} to evaluate the proposed \MODELSHORT~model, including two pedestrian sets (ETH-UCY and SDD) and one vehicle dataset (nuScenes):

(a) \textbf{ETH\cite{pellegrini2009youll}-UCY\cite{lerner2007crowds}} contains several pedestrian walking scenes, including five main test scenes: eth, hotel, univ, zara1, zara2.
Models are validated with \emph{leave-one-out} \cite{alahi2016social,gupta2018social} strategy, under the common $\{t_h, t_f, \Delta t\} = \{8, 12, 0.4\}$ settings.

(b) \textbf{Stanford Drone Dataset (SDD) \cite{robicquet2016learning}} contains 60 drone videos captured in the Stanford campus.
Multiple kinds of agents, like pedestrians, bikers, and carts, have been labeled, presenting more complex interactions.
We randomly select 36 videos to train, 12 to validate, and the left 12 to test \SIMAUGCITE\GARDENCITE, also using the common $\{t_h, t_f, \Delta t\} = \{8, 12, 0.4\}$ settings.

(c) \textbf{nuScenes} \cite{caesar2019nuscenes} is a large-scale dataset collected from urban driving scenes, including multiple kinds of vehicles and pedestrians.
We use 550 scenes to train, 150 to validate, and 150 to test \cite{saadatnejad2022pedestrian}.
Only vehicles' trajectories are used \cite{yuan2021agentformer}, with the $\{t_h, t_f, \Delta t\} = \{4, 12, 0.5\}$ settings \cite{lee2022muse}.

\NEWHALFLINE
\subsubsection{Metrics}
We use the minimum Average/Final Displacement Error over $K$ generated trajectories \cite{alahi2016social,pellegrini2009youll,gupta2018social} to evaluate models' performance, known as ``minADE$_K$/minFDE$_K$'', where $K$ is usually set to 20 when evaluating.
For agent $i$, using subscript $_k$ to denote the $k$th prediction, we have
\begin{align}
    \mbox{minADE}_{20}(i) &= \min_{k \in \{1, 2, ..., K\}} \frac{1}{t_f} \sum_{t = t_h + 1}^{t_h + t_f} \left\|
        \hatbf{p}^i_{t_k} - \mathbf{p}^i_t
    \right\|_2, \\
    \mbox{minFDE}_{20}(i) &= \min_{k \in \{1, 2, ..., K\}} \left\|
        \hatbf{p}^i_{{t_h + t_f}_k} - \mathbf{p}^i_{t_h + t_f}
    \right\|_2.
\end{align}
The final metrics are averaged over every agent in each set, and we will refer to them as ``ADE/FDE'' for clarity.

In addition, we have proposed a qualitative metric, named \emph{reverberation strength}, to characterize the learned latencies for handling trajectory-changing events.
See detailed formulations and discussions in \SECTION{sec_discussion_rev}.

\NEWHALFLINE
\subsubsection{Baselines}
We choose several recent state-of-the-art methods as our baselines, including
\SHENET\SHENETCITE
, \MID\MIDCITE
, \RAN\RANCITE
, \MSTIP\MSTIPCITE
, \SMEMO\SMEMOCITE
, \LGTRAJ\LGTRAJCITE
, \MSRL\MSRLCITE
, \PPT\PPTCITE
, \EVMODEL\EVCITE
, \AGENTFORMER\AGENTFORMERCITE
, \SCMODEL\SCCITE
, \SCPMODEL\SCPCITE
, \YNET\YNETCITE
, \UPDD\UPDDCITE
, \REMODEL\RECITE
, \FLOWCHAIN\FLOWCHAINCITE
, \IMP\IMPCITE
, \SPECTGNN\SPECTGNNCITE
, \UEN\UENCITE
, \MUSEVAE\MUSEVAECITE
, \TRAJECTRONPP\TRAJECTRONPPCITE
, \SOPERMODEL\SOPERMODELCITE
, \DICE\DICECITE
, and \AFFLN\AFFLNCITE.

\NEWHALFLINE
\subsubsection{Implementation Details}
\MODELSHORT~model is trained on one NVIDIA RTX 3090, using an Adam optimizer with a learning rate of 3e-4 and a batch size of 1000 for up to 200 epochs.
When generating 20 trajectories, the number of columns of generating kernels $\mathbf{G}$ is set to $K_g = 20$.
Feature dimension of most subnetworks is set to $d = 128$.
According to conclusions of previous research \cite{wong2023socialcircle,wong2024socialcircle+,wong2024resonance}, the number of social partitions is set to $N_\theta = 8$.
Trajectories of ego agents have been translated to the origin as a preprocessing step \EVCITE~before the network implementation.
All reported metrics are averaged across five runs.
Model weights, training code, and detailed settings are provided in our code repo.
See additional theoretical properties, efficiency analyses, extended experiments (motion prediction), statistical verifications, and further conceptual interpretations and comparisons in Appendix.

\begin{table*}
    \caption{
        Comparisons with Other State-of-the-Arts on ETH-UCY (left) and SDD (right)
    }
    \label{tab_ethucy}
    \centering
    \begin{threeparttable}
        \begin{tabular}{@{} c @{\hspace{0.5em}} c @{}}
            \begin{tabular}{lcccccccccccccc}
                \toprule
                Method (ETH-UCY) & eth $\downarrow$ & hotel $\downarrow$ & univ $\downarrow$ & zara1 $\downarrow$ & zara2 $\downarrow$ & Average $\downarrow$ \\

                \midrule
                \SHENET\SHENETCITE~(2022)           & 0.41/0.61         & 0.13/0.20         & 0.25/0.43         & 0.21/0.32         & 0.15/0.26         & 0.23/0.36 \\
                \MID\MIDCITE~(2022)                 & 0.39/0.66         & 0.13/0.22         & \C{0.22}/0.45     & \C{0.17}/0.30     & \C{0.13}/0.27     & 0.21/0.38 \\
                \RAN\RANCITE~(2024)                 & 0.41/0.69         & 0.13/0.21         & 0.25/0.46         & 0.22/0.41         & 0.16/0.31         & 0.23/0.42 \\
                \MSTIP\MSTIPCITE~(2024)             & 0.39/0.57         & 0.13/0.22         & 0.24/\C{0.40}     & 0.20/0.34         & 0.17/0.29         & 0.22/0.36 \\
                \SMEMO\SMEMOCITE~(2024)             & 0.39/0.59         & 0.14/0.20         & \C{0.23/0.41}     & 0.19/0.32         & 0.15/0.26         & 0.22/0.35 \\
                \LGTRAJ\LGTRAJCITE~(2024)           & 0.38/0.56         & \C{0.11}/0.17     & \C{0.23}/0.42     & 0.18/0.33         & \C{0.14}/0.25     & 0.20/0.34 \\
                \MSRL\MSRLCITE~(2023)               & 0.28/0.47         & 0.14/0.22         & 0.24/0.43         & \C{0.17}/0.30     & \C{0.14/0.23}     & \C{0.19}/0.33 \\
                \PPT\PPTCITE~(2024)                 & 0.36/0.51         & \C{0.11/0.15}     & \C{0.22/0.40}     & \C{0.17}/0.30     & \C{0.12/0.21}     & 0.20/0.31 \\
                \EVMODEL\EVCITE~(2025)              & \C{0.25}/0.38     & \C{0.11/0.16}     & \C{0.23}/0.42     & 0.19/0.30         & \C{0.13}/0.24     & \C{0.18}/0.30 \\
                \AGENTFORMER\AGENTFORMERCITE~(2021) & 0.26/0.39         & \C{0.11/0.14}     & 0.26/0.46         & \C{0.15/0.23}     & \C{0.14}/0.23     & \C{0.18/0.29} \\
                \SCMODEL\SCCITE~(2024)              & \C{0.25}/0.38     & \C{0.12/0.14}     & \C{0.23}/0.42     & 0.18/0.29         & \C{0.13/0.22}     & \C{0.18/0.29} \\
                \SCPMODEL\SCPCITE~(2024)            & \C{0.25}/0.39     & \C{0.10/0.15}     & 0.24/0.42         & 0.18/\C{0.28}     & \C{0.13/0.22}     & \C{0.18/0.29} \\
                \YNET\YNETCITE~(2021)               & 0.28/\C{0.33}     & \C{0.10/0.14}     & 0.24/\C{0.41}     & \C{0.17/0.27}     & \C{0.13/0.22}     & \C{0.18/0.27} \\
                \UPDD\UPDDCITE~(2024)               & \C{0.22}/0.42     & 0.17/0.30         & \C{0.14/0.28}     & \C{0.16}/0.30     & \C{0.14}/0.31     & \C{0.17}/0.32 \\
                \REMODEL\RECITE~(2025)              & \C{0.23/0.35}     & \C{0.10/0.15}     & 0.24/\C{0.41}     & \C{0.17}/0.29     & \C{0.13/0.22}     & \C{0.17/0.28} \\ 

                \midrule
                \MODELSHORT~(Ours)                  & \C{0.23/0.34}     & \C{0.10/0.15}     & \C{0.23/0.40}     & \C{0.17/0.28}     & \C{0.13/0.22}     & \C{0.17/0.28} \\ 

                \bottomrule
            \end{tabular}

            &

            \begin{tabular}{lc}
                \toprule
                Method (SDD) & ADE/FDE $\downarrow$ \\

                \midrule
                \RAN\RANCITE~(2024) & 10.97/19.95 \\
                \FLOWCHAIN\FLOWCHAINCITE~(2023) & 9.93/17.17 \\
                \IMP\IMPCITE~(2023) & 8.98/15.54 \\
                \SPECTGNN\SPECTGNNCITE~(2021) & 8.21/12.41 \\
                \SMEMO\SMEMOCITE~(2024) & 8.11/13.06 \\
                \LGTRAJ\LGTRAJCITE~(2024) & 7.80/12.79 \\
                \YNET\YNETCITE~(2021) & 7.85/11.85 \\
                \UEN\UENCITE~(2024) & 7.30/10.40 \\
                \PPT\PPTCITE~(2024) & 7.03/10.65 \\
                \UPDD\UPDDCITE~(2024) & 6.59/13.90 \\
                \EVMODEL\EVCITE~(2025) & 6.57/10.49 \\
                \SCMODEL\SCCITE~(2024) & 6.54/10.36 \\
                \SCPMODEL\SCPCITE~(2024) & 6.44/\C{10.22} \\
                \MUSEVAE\MUSEVAECITE~(2022) & \C{6.36}/11.10 \\
                \REMODEL\RECITE~(2025) & \C{6.27/10.02} \\
                
                \midrule
                \MODELSHORT~(Ours) & \C{6.14/9.87} \\

                \bottomrule
            \end{tabular}
        \end{tabular}

        \begin{tablenotes}[flushleft]
            \footnotesize
            \item[]
                Metrics are reported as ``ADE/FDE'' (\emph{best-of-20}).
                Lower metrics indicate better prediction performance.
                \C{Blue markers} denote best 3 results on each set.
        \end{tablenotes}
    \end{threeparttable}
\end{table*}

\begin{table}[tbp]
    \caption{
        Comparisons with Other State-of-the-Arts on nuScenes
    }
    \label{tab_nbanuscenes}
    \centering
    \begin{threeparttable}
        \begin{tabular}{lcc}
            \toprule
            Method (nuScenes) & \emph{best-of-5} $\downarrow$ & \emph{best-of-10} $\downarrow$ \\

            \midrule
            \TRAJECTRONPP\TRAJECTRONPPCITE~(2020) & 3.14/7.45 & 2.46/5.65 \\
            \YNET\YNETCITE~(2020) & 2.46/5.15 & 1.88/3.47 \\
            \SOPERMODEL\SOPERMODELCITE~(2025) & 2.21/4.58 & 1.79/3.40 \\
            \AGENTFORMER\AGENTFORMERCITE~(2021) & 1.86/3.89 & 1.45/2.86 \\
            \DICE\DICECITE~(2024) & 1.76/3.70 & 1.44/2.67 \\
            \AFFLN\AFFLNCITE~(2024) & 1.83/3.78 & 1.32/2.73 \\
            \EVMODEL\EVCITE~(2025) & 1.46/3.18 & 1.15/2.37 \\
            \SCMODEL\SCCITE~(2024) & 1.44/3.10 & 1.13/2.30 \\
            \MUSEVAE\MUSEVAECITE~(2022) & \C{1.38/2.90} & \C{1.09/2.10} \\
            \REMODEL\RECITE~(2025) & \C{1.26/2.72} & \C{1.03/2.12} \\

            \midrule
            \MODELSHORT~(Ours) & \C{1.25/2.73} & \C{1.00/2.03} \\

            \bottomrule
        \end{tabular}
        \begin{tablenotes}[flushleft]
            \footnotesize
            \item[]
                Metrics reported are ``ADE/FDE'' in meters under \emph{best-of-5} and \emph{best-of-10}.
                \C{Blue markers} denote best 3 results under each setting.
        \end{tablenotes}
    \end{threeparttable}
\end{table}

\subsection{Comparisons to State-of-the-Arts}

This section quantitatively compares \MODELSHORT~with several recent state-of-the-art methods.
See results in \TABLE{tab_ethucy,tab_nbanuscenes}.

\textbf{Pedestrian Datasets:}
It should be noted that the prediction improvements on these two pedestrian datasets have become increasingly marginal in recent years, approaching a potential bottleneck.
Under these conditions, the proposed \MODELSHORT~model still demonstrates competitive performance.
In \TABLE{tab_ethucy}, it can be seen that \MODELSHORT~could obtain top-3 prediction performance on all five ETH-UCY sets, presenting its adaptivity across different scenes.
Interestingly, on univ and zara2, \MODELSHORT~outperforms most competing methods.
On univ, it performs similarly to recent baselines such as \SCPMODEL~and \PPT, while on zara2 it achieves about 29\% lower FDE than \UPDD, highlighting its strength in dense pedestrian scenes.
Also, its performance gains are balanced in terms of both ADE and FDE, indicating that \MODELSHORT~can simultaneously capture global motion trends and final destinations.
Notably, \MODELSHORT~outperforms the state-of-the-art \REMODEL~by about 2.1\% lower ADE and 1.5\% lower FDE, and further improves over \MUSEVAE~by about 3.5\% in ADE, where such improvements are rarely seen even on SDD.
This demonstrates its adaptability to diverse pedestrian scenarios, whether on streets or campuses.

\textbf{Vehicle Dataset:}
\TABLE{tab_nbanuscenes} shows that \MODELSHORT~also achieves noticeable performance gains on the nuScenes vehicle dataset.
When generating $K = 10$ trajectories, it outperforms \REMODEL~by about 4.2\% FDE.
Our above observations also hold on this vehicle dataset, \IE, \MODELSHORT~balances improvements in both global trends (ADE) and destination considerations (FDE).
For example, comparing \REMODEL~and \MUSEVAE, the former achieves lower ADE while the other yields lower FDE.
Meanwhile, \MODELSHORT~has about 2.9\% lower ADE and 3.3\% lower FDE than both methods.
Note that \MODELSHORT~performs slightly worse than \REMODEL~(0.8\%/0.4\% worse ADE/FDE) when only generating $K = 5$ trajectories, revealing potential shortcomings when generating a limited number of trajectories.
Nevertheless, it still outperforms \MUSEVAE~by about 9.4\%/9.3\% of ADE/FDE under this situation, presenting its competitiveness.

\begin{table*}[tbph]
    \caption{
        Ablation Studies on ETH-UCY and SDD
    }
    \label{tab_ab}
    \centering
    \begin{threeparttable}
        \begin{tabular}{lcccccccccccccc}
            \toprule
            ID &
            $\mathcal{R}$ & $\mathbf{G}$ & $\mathbf{R}$ &
            $\LIN$~~$\NON$~~$\SOC$ &
            eth & hotel & univ & zara1 & zara2 & ETH-UCY
            & SDD \\

            \midrule
            a1 & \xm & -- & --      & \emph{Forecast together}  & 0.265/0.424           & 0.109/0.167      & 0.344/0.620        & 0.194/0.336       & 0.138/0.240       & 0.210/0.357       & 6.477/10.460 \\
            a2 & \cm & \xm & \cm    & \emph{Forecast together}  & 0.269/0.395           & 0.114/0.171      & 0.387/0.692        & 0.196/0.315       & 0.142/0.238       & 0.222/0.362       & 6.381/10.095 \\
            a3 & \cm & \cm & \xm    & \emph{Forecast together}  & 0.243/0.373           & 0.117/0.164      & 0.415/0.779        & 0.194/0.316       & 0.140/0.240       & 0.222/0.374       & 6.337/10.176 \\
            a4 & \cm & \cm & \cm    & \emph{Forecast together}  & 0.265/0.401           & 0.110/0.161      & 0.310/0.541        & 0.191/0.314       & 0.141/0.231       & 0.203/0.329       & 6.386/\C{10.022} \\

            a5 & \cm & \cm & \cm    & \xm~~~~\cm~~~~\cm         & 0.273/0.380           & \C{0.108}/0.160  & 0.337/0.593        & 0.190/0.299       & 0.138/\C{0.227}   & 0.209/0.332       & 6.296/10.064 \\
            a6 & \cm & \cm & \cm    & \cm~~~~\xm~~~~\cm         & 0.238/0.377           & 0.110/0.164      & \C{0.249}/0.440    & \C{0.174/0.294}   & 0.136/0.236       & \C{0.181}/0.302   & 6.286/10.247 \\
            a7 & \cm & \cm & \cm    & \cm~~~~\cm~~~~\xm         & \CC{0.228}/\C{0.363}  & 0.109/0.176      & \C{0.241/0.426}    & 0.176/0.297       & 0.134/\C{0.224}   & \C{0.178/0.297}   & \C{6.227}/10.046 \\

            a8 & \xm & -- & --      & \cm~~~~\cm~~~~\cm         & 0.255/0.400           & \C{0.107}/0.160  & 0.280/0.521        & 0.178/0.305       & \C{0.133}/0.230   & 0.191/0.323       & 6.446/10.672 \\
            a9 & \cm & \xm & \cm    & \cm~~~~\cm~~~~\cm         & \C{0.235/0.360}       & 0.112/\C{0.158}  & 0.250/\C{0.439}    & \C{0.175/0.296}   & \C{0.132/0.227}   & \C{0.181/0.296}   & \C{6.176/9.896} \\
            a10 & \cm & \cm & \xm   & \cm~~~~\cm~~~~\cm         & 0.238/0.366           & \C{0.107/0.155}  & 0.260/0.473        & 0.176/0.301       & \C{0.133}/0.238   & 0.182/0.306       & 6.326/10.388 \\
            
            \midrule
            full & \cm & \cm & \cm  & \cm~~~~\cm~~~~\cm         & \C{0.229}/\CC{0.347}  & \CC{0.105/0.154} & \CC{0.237/0.409}   & \CC{0.172/0.283}  & \CC{0.131/0.220}  & \CC{0.174/0.282}  & \CC{6.144/9.874} \\

            \bottomrule
        \end{tabular}
        \begin{tablenotes}[flushleft]
            \footnotesize
            \item[]
                ``$\mathcal{R}$'' denotes whether or how the reverberation transform is applied.
                Alternative basic mapping networks with similar number of parameters will be used as replacements when reverberation kernels $\mathbf{R}$ and $\mathbf{G}$ are disabled (marked with \xm) to map features from observation to prediction periods and to generate stochastic predictions, respectively.
                ``$\LIN$'', ``$\NON$'', and ``$\SOC$'' denotes whether linear, non-interactive, and social events (predictions) are considered when training and forecasting.
                Metrics reported are ``ADE/FDE'' using \emph{best-of-20}.
                \C{Blue} numbers mark the top-3 results, and \CC{underline} denote the best metrics.
        \end{tablenotes}
    \end{threeparttable}
\end{table*}

\begin{table}[tbph]
    \caption{
        Ablation Studies on nuScenes (only vehicles, 4 $\to$ 12 frames)
    }
    \label{tab_ab2}
    \centering
    \begin{threeparttable}
        \begin{tabular}{lcccccccccccccc}
            \toprule
            ID &
            $\mathcal{R}$ & $\mathbf{G}$ & $\mathbf{R}$ &
            $\LIN$~~$\NON$~~$\SOC$ & \emph{best-of-5} & \emph{best-of-10} \\

            \midrule
            a1 & \xm & --  & --     & \emph{Forecast together}  & 1.571/3.354       & 1.342/2.930 \\
            a2 & \cm & \xm & \cm    & \emph{Forecast together}  & 1.339/2.897       & 1.107/2.236 \\
            a3 & \cm & \cm & \xm    & \emph{Forecast together}  & 1.455/3.054       & 1.087/2.312 \\
            a4 & \cm & \cm & \cm    & \emph{Forecast together}  & 1.320/\C{2.836}   & 1.078/\C{2.160} \\

            a5 & \cm & \cm & \cm    & \xm~~~~\cm~~~~\cm         & \CC{1.257/2.734}  & \CC{1.002/2.025} \\
            a6 & \cm & \cm & \cm    & \cm~~~~\xm~~~~\cm         & 1.315/2.928       & 1.061/2.218\\
            a7 & \cm & \cm & \cm    & \cm~~~~\cm~~~~\xm         & \C{1.307}/2.889   & 1.065/2.219\\

            a8 & \xm & -- & --      & \xm~~~~\cm~~~~\cm         & 1.344/2.893       & 1.111/2.281 \\
            a9 & \cm & \xm & \cm    & \xm~~~~\cm~~~~\cm         & 1.308/2.897       & \C{1.050/2.170} \\
            a10 & \cm & \cm & \xm   & \xm~~~~\cm~~~~\cm         & 1.309/\C{2.861}   & 1.054/2.188 \\

            \midrule
            full & \cm & \cm & \cm  & \cm~~~~\cm~~~~\cm         & \C{1.306}/2.901   & \C{1.040}/2.173  \\

            \bottomrule
        \end{tabular}
    \end{threeparttable}
    \begin{tablenotes}[flushleft]
        \footnotesize
        \item[]
            Metrics are reported in the form of ``ADE/FDE'' in meters.
    \end{tablenotes}
\end{table}

\subsection{Ablation Studies}

This section conducts ablations to verify how each proposed component contributes to the \MODELSHORT~model.
Detailed ablation settings and results are reported in \TABLE{tab_ab,tab_ab2}.
Reported results are averaged over 5 runs to reduce randomness.

\NEWHALFLINE
\subsubsection{Overall Verifications of Reverberation Transform}
We use a series of minimal variations \{a1, a2, a3, a4\} to verify how the reverberation transform helps achieve better performance.
These variations only include a single Transformer backbone for prediction.
They do not include any further operations like decoupling prediction into non-interactive or social portions, with only social-interaction-modeling components reserved.
We observe from \TABLE{tab_ab,tab_ab2} that the use of reverberation transform could directly lead to better quantitative performance, especially by using its two kernels $\mathbf{R}$ and $\mathbf{G}$ simultaneously.
In particular, this transform may bring lower FDEs.
For example, variation a4 outperforms a1 by up to 5.7\% FDE on eth and 12.7\% FDE on univ.
Variation a4 also brings noticeable ADE enhancements, including about 3.3\% on the average of ETH-UCY, 1.4\% on SDD, and 19.6\% on nuScenes (\emph{best-of-10}), although such enhancements are not as \emph{strong} as those in FDE (7.8\%, 4.2\%, and 26.3\%, respectively).
In addition, consistent with the findings in the previous \REMODEL~\RECITE, the decoupling of non-interactive and social components is crucial for handling the high stochasticity of pedestrian trajectories, whereas the proposed reverberation transform can independently achieve significant gains on the more deterministic nuScenes (\TABLE{tab_ab2}).
These results validate its usefulness, especially in forecasting trajectories in the further future, where all latencies have been posted intuitively for better FDE gains than ADE.

\NEWHALFLINE
\subsubsection{Verifications of Reverberation Kernel $\mathbf{R}$}
This kernel serves as the main component to map features and learn initial latencies for handling trajectory-changing events.
When this kernel is disabled in variations \{a3, a10\}, a simple but parameter-equivalent MLP will be used to take over its duty of mapping features from the observation period to the prediction period.
Comparing variation-pairs \{a3, a4\} or \{a10, full\} in \TABLE{tab_ab,tab_ab2}, consistent with the above conclusions, it can be seen that FDEs have been far more than ADEs, like 8.6\%/12.0\% gains on ETH-UCY (a3 and a4), indicating the potential relationship between this reverberation kernel and latencies in the further future.
Variations \{a10, full\} show similar trends, including 4.4\%/7.8\% gains on ETH-UCY and 2.9\%/4.9\% on SDD.
It shows that the kernel $\mathbf{R}$ plays the core role to consider how trajectories could grow in the further future, also proving the usefulness of directly modeling latencies of trajectory-changing events when forecasting.

\NEWHALFLINE
\subsubsection{Verifications of Generating Kernel $\mathbf{G}$}
This kernel provides randomness and alternatives to $\mathbf{R}$, simulating agents' unique stochastic properties of latencies.
In variations a2 and a9, we use the \MSN-like~\MSNCITE~strategy for comparison when this kernel is disabled.
Comparing variations \{a9, full\} in \TABLE{tab_ab,tab_ab2}, we see that kernel $\mathbf{G}$ achieves larger ADE improvements than kernel $\mathbf{R}$, gaining 3.9\%/4.7\% on ETH-UCY and even 1.0\%/-1.8\% on nuScenes (10).
These comparisons indicate that the kernel $\mathbf{G}$ is better at improving the predicted trajectories globally (due to its ADE gains), presenting a totally different role compared to the reverberation kernel $\mathbf{R}$.
Especially, we see that the use of kernel $\mathbf{G}$ even leads to a little worse FDE on nuScenes, which could be a \emph{compromise} of prediction accuracy and diversity (See \SECTION{sec_gen}).

\NEWHALFLINE
\subsubsection{Verifications of Non-interactive and Social Latencies}
We use two parallel networks to learn latencies of agents for handling non-interactive and social trajectory-changing events and making the corresponding predictions.
Comparing variations \{a6, a7, full\} in \TABLE{tab_ab,tab_ab2}, it can be seen that the use of both branches mostly leads to the best prediction performance.
Specifically, removing the non-interactive branch from the full variation may lead to more performance drops than removing the social branch, like -4.0\%/-7.0\% versus -2.3\%/-5.3\% on ETH-UCY.
Such a difference is consistent with our intuition that social behaviors may not significantly modify our scheduled trajectory, as discussed from the energy perspective in \REMODEL~\RECITE.
Nevertheless, both such branches could still provide comparable performance gains compared to the ``forecast-together'' variation a4, with up to 14.3\% gains on ETH-UCY.
Similar trends could also be observed for the nuScenes in \TABLE{tab_ab2}, presenting an interesting phenomenon that the linear part could make the performance slightly worse in this vehicle dataset, verifying the adaptability of these branches in adjusting roles in different datasets dynamically.

\subsection{Discussions of Reverberations and Latencies}
\label{sec_discussion_rev}

\begin{figure}[tbp]
    \centering
    \includegraphics[width=1.0\linewidth]{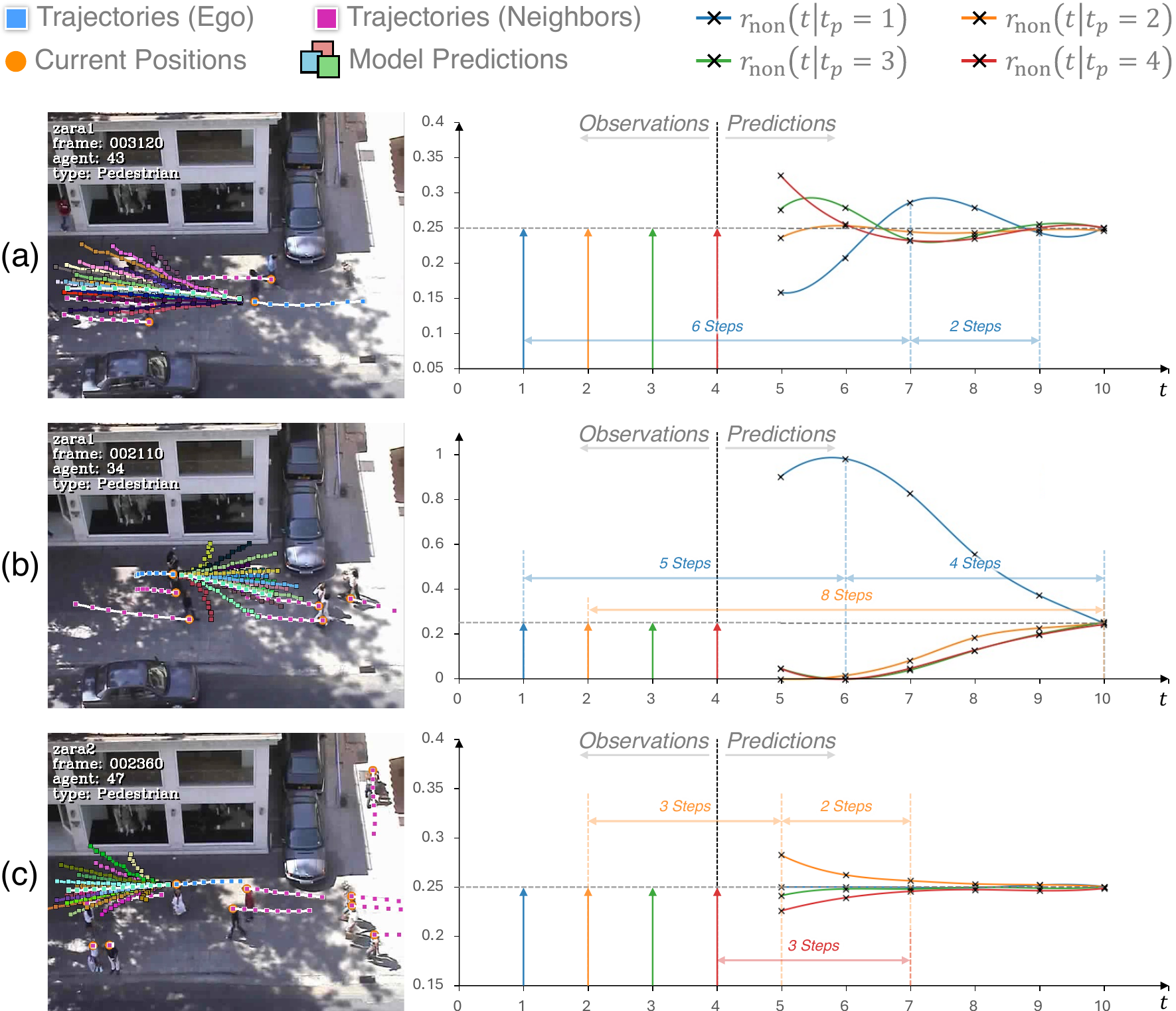}
    \caption{
        Examples of non-interactive reverberation strength $r_\NON (t|t_p)~(t_p \in \{1, 2, 3, 4\})$ of kernel $\mathbf{R}_\NON$ in the proposed \MODELSHORT~model.
        ($\{T_h, T_f\} = \{4, 6\}$)
    }
    \label{fig_curve_R_non}
\end{figure}

\begin{figure*}[tbp]
    \centering
    \includegraphics[width=1.0\linewidth]{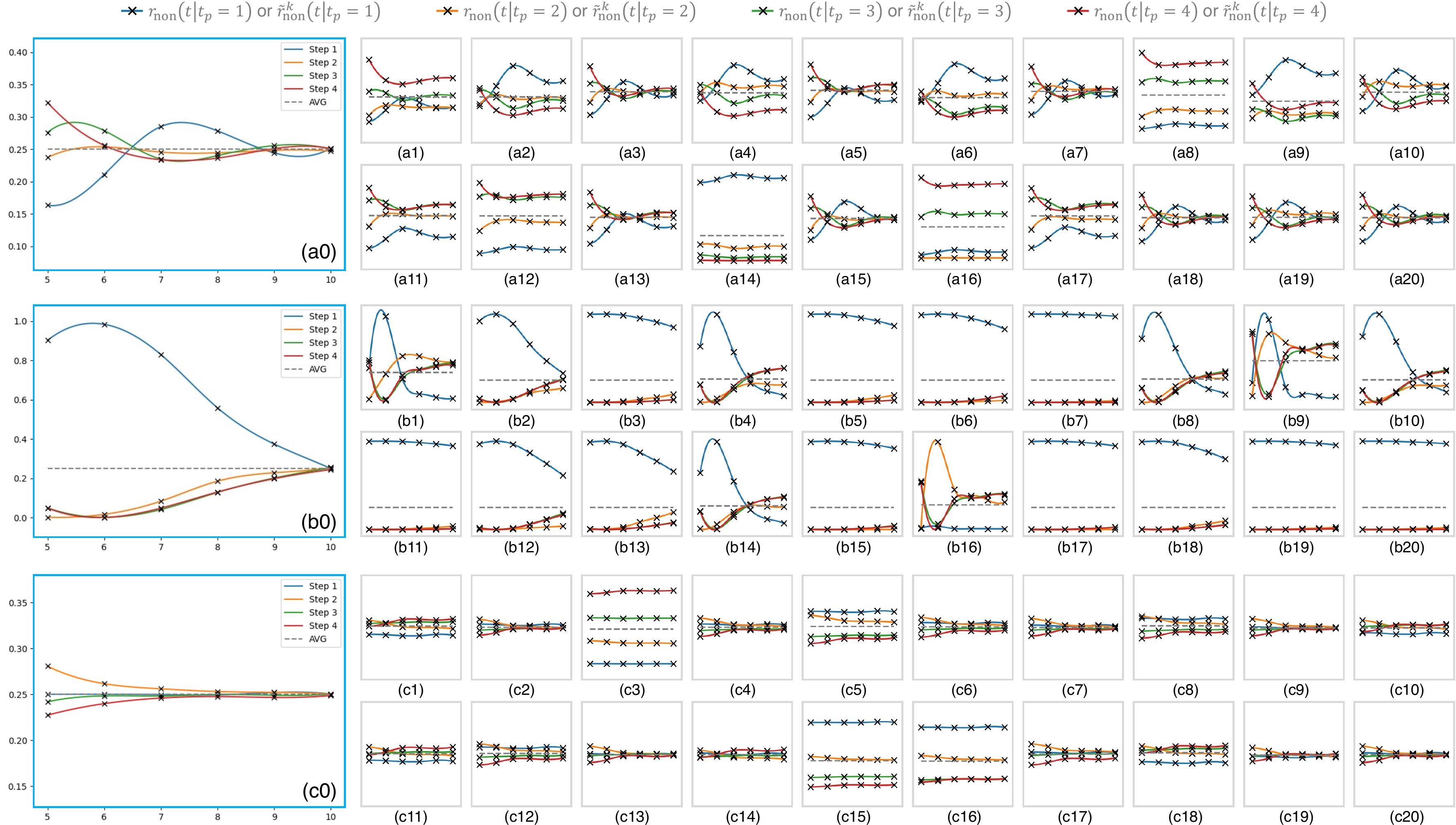}
    \caption{
        Examples of the original non-interactive reverberation strength $r_\NON (t|t_p)$ ($t_p \in \{1, 2, 3, 4\}$, in blue boxes) and its $K_g = 20$ altered $\tilde{r}_\NON^k (t|t_p)$ ($t_p \in \{1, 2, 3, 4\}$, $k \in \{1, 2, ..., K_g\}$) curves after applying the generating kernel $\mathbf{G}_\NON$.
        $k$ refers to indices of subfigures in grey boxes.
    }
    \label{fig_curve_G_non}
\end{figure*}

Latencies within trajectories, \IE, how long each agent will take to notice, to start handling, or finally to mark a trajectory-changing event as completed when planning trajectories, are our main focus.
We now provide intuitive descriptions and interpretations of how the proposed reverberation transform and the \MODELSHORT~model capture and forecast such latencies.

As defined in \EQUA{eq_rev_transform}, a reverberation transform $\mathcal{R}_{(\mathbf{R}, \mathbf{G})}$ is determined by one reverberation kernel $\mathbf{R}$ and its corresponding generating kernel $\mathbf{G}$.
In the proposed \MODELSHORT~model, we use two sets of such kernels, $\{\mathbf{R}_\NON, \mathbf{G}_\NON\}$ and $\{\mathbf{R}_\SOC, \mathbf{G}_\SOC\}$\footnote{
    Superscript $^i$ will be omitted for clarity when indicating agents' indices.
}, to learn and to forecast latencies of non-interactive and social trajectory-changing events correspondingly.
We will first discuss non-interactive kernels, then social kernels.

According to the structure of \MODELSHORT, for non-interactive latencies and predictions, the minimum unit for considering trajectory-changing events is an observation step, while it is a social partition at any observation step for social predictions.
For clarity, we refer to ``the feature of a \textbf{minimum unit} of trajectory-changing events'' as ``a trajectory-changing event'' or simply ``an event'' in the following discussions.
Also, due to the use of Haar transform in \MODELSHORT~models, in the following discussions, our mentioned ``one step'' actually refers to 2 adjacent observation/prediction frames, \IE, 0.8 seconds on ETH-UCY or SDD.
Therefore, the following discussions are conducted under $\{T_h, T_f\} = \{4, 6\}$, corresponding to the common $\{t_h, t_f\} = \{8, 12\}$ setting.

\NEWHALFLINE
\subsubsection{Non-Interactive Reverberation $r_\NON$}
\label{sec_discussion_r_non}
Inspired by the reverberation curve in acoustics, we now define how latencies are represented from the reverberation kernels $\mathbf{R}$ and $\mathbf{G}$.
The non-interactive reverberation curves aim to describe how a non-interactive (\emph{self-sourced}~\RECITE) trajectory-changing event, which happens at any past (observed) step $t_p \in \{1, 2, ..., T_h\}$, produces influences or contributes to the future trajectory at any future (prediction) step $t_q \in \{T_h + 1, T_h + 2, ..., T_h + T_f\}$.
The non-interactive reverberation kernel $\mathbf{R}_\NON$ is a 2D matrix with $T_h$ rows and $T_f$ columns, corresponding to $T_h$ historical and $T_f$ future events (minimum units of trajectory-changing events).
For the ease of presentation, we denote the element (a real number) at $t_p$th row and $(t_q - T_h)$th column of matrix $\mathbf{R}_\NON$ as $\mathbf{R}_\NON[t_p, t_q]$.
Then, for an observed step $t_p$, given variable $t \in \{T_h + 1, ..., T_h + T_f\}$, we define the relative non-interactive reverberation strength $r_\NON (t|t_p)$ (we also call it the non-interactive \emph{Reverberation Curve} in our discussions) as
\begin{equation}
    \label{eq_non_interactive_R}
    r_\NON (t|t_p) = \frac{
        (\mathbf{R}_\NON[t_p, t])^2
    }{
        \sum_{t_x=1}^{T_h} (\mathbf{R}_\NON[t_x, t])^2
    }.
\end{equation}
It describes how a unit non-interactive event (at step $t_p$, corresponding to the minimum unit of trajectories-changing events but with unit amplitudes) relatively provides additional energy during the prediction period, which is a measurement of the relative \emph{power gain} (See \APPENDIX{B}) for modifying future trajectories by an ego agent.
For a unit event happening at the $h_0$th historical step ($h_0 \in \{1, 2, ..., T_h\}$), the \textbf{peaking latency} $L_p(h_0)$ is then defined as the temporal offset between $h_0$ and the step where the reverberation strength $r(t|t_p = h_0)$ (including both the above $r_\NON$ and $\tilde{r}_\NON^k$, $r_\SOC$, $\tilde{r}_\SOC^k$ defined below) hits its maximum value:
\begin{gather}
    \label{eq_peaking_latency}
    L_p(h_0) = t_{\mathrm{peak}}(h_0) - h_0, \\
    \quad\mbox{where}~
    t_{\mathrm{peak}} (h_0) = \mathop{\arg\max}_t r(t|h_0).
\end{gather}
When the reverberation strength $r(t|t_p = h_0)$ is not monotonically increasing and exhibits significant decreasing during the prediction period (which declines to a level near the certain threshold $\epsilon = 0.03$ of the mean strength of all events $1/T_h \pm \epsilon$), the corresponding \textbf{vanishing latency} is defined as
\begin{gather}
    \label{eq_vanishing_latency}
    L_v (h_0) = t_{\mathrm{van}}(h_0) - h_0, \\
    \quad\mbox{where}~
    t_{\mathrm{van}} (h_0) = \mathop{\arg\min}_{t > t_{\mathrm{peak}}(h_0)} \left|
        r(t|h_0) - \left(\frac{1}{T_h} + \epsilon\right)
    \right|.
\end{gather}

\begin{figure*}[tbp]
    \centering
    \includegraphics[width=1.0\linewidth]{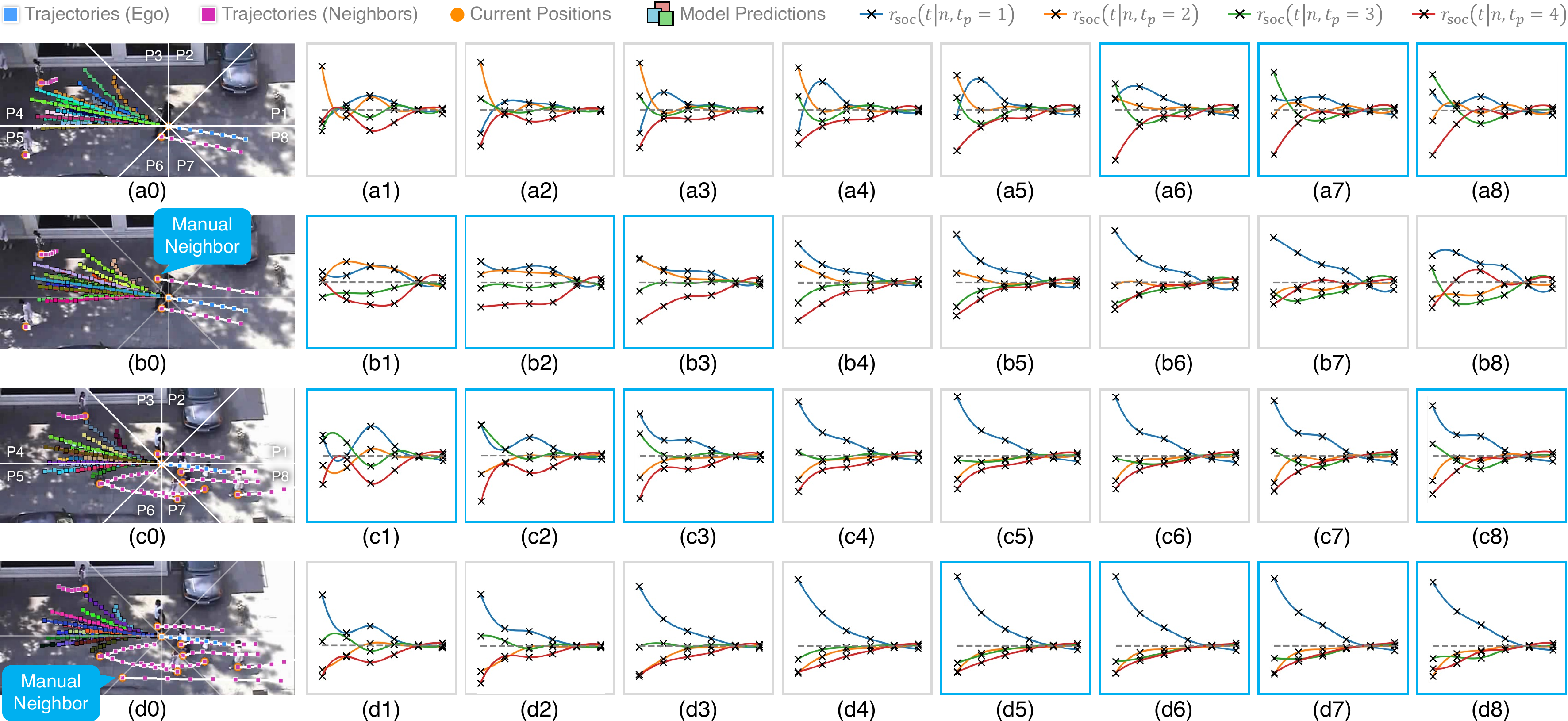}
    \caption{
        Examples of social reverberation strength $r_\SOC (t|n, t_p)$ for historical steps $t_p \in \{1, 2, 3, 4\}$ and social partition $n \in \{1, 2, ..., 8\}$.
        Index of a subfigure in each row indicates the index of social partitions, like subfigure (a1) displays social reverberation strength in the partition 1.
        Cases (b0) and (d0) have been added manual neighbors \SCCITE\SCPCITE~upon the original (a0) and (c0) scenes to verify the sensitivity of reverberation curves to handle social interactions.
    }
    \label{fig_curve_R_social}
\end{figure*}

\FIG{fig_curve_R_non} visualizes several $r_\NON$ curves.
In case (a), it considers that event occurring at the historical step 1 will be postponed\footnote{
    Grey dashlines in \FIG{fig_curve_R_non,fig_curve_G_non,fig_curve_R_social,fig_curve_G_social} represent the average strength of all $T_h = 4$ steps.
    It serves as a baseline when estimating latencies, like a past step can be considered delayed to take effect if its curve lies below this line initially.
} for about 6 steps (4.8s) to allow the ego agent to handle it, \IE, with a peaking latency $L_p(1)$ of about 6 steps.
After that, such event will receive the highest concentration for about 2 steps (1.6s) to decide future trajectories, while other events will be toned down as a compromise.
We can see from the left figure that the ego agent starts turning at step 1 (the first two observation frames).
Such forecasted latencies could be explained as the prediction model regards that the agent may focus more on finishing such turning before step 7, and then reconsiders its movements by taking into account where the turn is started and finally located.

Interestingly, all observed events may eventually contribute similarly to trajectories in the further future (like after 5 steps or 4.0s in (a)), aligning with the intuition that all trajectory-changing events may only take effect within limited periods, especially when we are walking or behaving as pedestrians, though such effect may dramatically change the future path we selected.
Such a conclusion also applies to \FIG{fig_curve_R_non} (c), where every historical event contributes almost the same to all future steps when the ego pedestrian keeps moving at a static speed, with no noteworthy events being captured.
We can conclude that the reverberation transform has the ability to learn and to judge different latency preferences for different agents in handling those non-interactive trajectory-changing events.

\NEWHALFLINE
\subsubsection{Altered Non-Interactive Reverberation $\tilde{r}_\NON$}
\label{sec_discussion_g_non}
Generating kernel $\mathbf{G}_\NON$ alters the original reverberation kernel $\mathbf{R}_\NON$ in $K_g$ ways to simulate randomness or variations of latencies for handling these non-interactive events.
To better visualize such modifications, we define the non-interactive reverberation curves that are obtained after applying a generating kernel $\mathbf{G}$ upon the corresponding reverberation kernel $\mathbf{R}$ as the set of \emph{Altered Reverberation Curves}.
Each set contains $K_g$ altered but separated reverberation curves, symbolized with $\tilde{r}_\NON = \{\tilde{r}_\NON^k\}_{k=1}^{K_g}$, corresponding to $K_g$ columns $\mathbf{G}_\NON$.
Like above discussions, we use symbol $\mathbf{G}_\NON [t_p, k]$ to index matrix elements.
Then, the $k$th ($k \in \{1, 2, ..., K_g\}$) altered non-interactive reverberation strength (curve) is defined as
\begin{equation}
    \label{eq_g_non}
    \tilde{r}_\NON^k (t|t_p) = \frac{
        \left(\mathbf{R}_\NON [t_p, t]~\mathbf{G}_\NON [t_p, k]\right)^2
    }{
        \sum_{t_x = 1}^{T_h}
        \left(\mathbf{R}_\NON [t_x, t]~\mathbf{G}_\NON [t_x, k]\right)^2
    }.
\end{equation}

\FIG{fig_curve_G_non} visualizes several altered reverberation curves along with their original ones, corresponding to three prediction scenes in \FIG{fig_curve_R_non}.
Here, we observe that generating kernels may not significantly alter the trend of how a single historical event is delayed or begins contributing to future trajectories, but provide alternative combinations that reflect how ego agents collectively consider latencies of all historical steps.
In \FIG{fig_curve_G_non} (a0), it can be seen that event at step 1 is delayed to take effect until steps 7 and 8.
Interestingly, increases or decreases within $r_\NON(t|t_p = 1)$ have been maintained in most of its altered ones (any $\tilde{r}_\NON^k (t|t_p = 1)$), like generations (a2) (a3) (a5) and (a11) that all assign the most concentration for the event at step 1 regarding the future step 7.
Compared to the original curves, the main difference is that latencies of different events are balanced in obviously different ways after applying this generating kernel $\mathbf{G}_\NON$.
For example, considering all historical events, event at step 1 has been paid the most concentration in generation (a2), while the minimum in (a12).
Such results could verify the effectiveness of generating kernels on providing multiple latency preferences.

We can conclude that the role of a reverberation kernel $\mathbf{R}_\NON$ is to learn latencies for different ego agents to handle (minimum units of) trajectory-changing events at each single historical step finely, while a generating kernel $\mathbf{G}_\NON$ is to macroscopically integrate latencies of all these historical events.
Both kernels work together to achieve our goal of learning and forecasting unique latency preferences for different ego agents for forecasting trajectories.

\NEWHALFLINE
\subsubsection{Social Reverberation $r_\SOC$}
\label{sec_discussion_r_soc}
Similar to the above non-interactive latencies, the proposed \MODELSHORT~model employs a social reverberation kernel $\mathbf{R}_\SOC$ and a social generating kernel $\mathbf{G}_\SOC$ to cooperatively handle latencies within social events.
Following previous works \cite{wong2023socialcircle}, an angle-partition-based social-interaction-modeling strategy has been used to spatially and temporally represent social interactions.
Compared to the above non-interactive cases, kernel $\mathbf{R}_\SOC$ has been expanded to learn latencies of events on all $N_\theta$ partitions simultaneously.
This means that the minimum unit for handling social events has become one partition at each temporal-spectral step.
In our experiments, $\mathbf{R}_\SOC$ is a 2D matrix with $T_h N_\theta$ rows and $T_f$ columns.
Here, each social event could be uniquely located both spatially ($N_\theta$ angle partitions) and temporally ($T_h$ steps), and then be bridged to each future step.
For the ease of presentation, we treat it as a 3D tensor of $N_\theta \times T_h \times T_f$, and use symbol $\mathbf{R}_\SOC [n, p, q]$ to index each element.
Like the previous definition in \EQUA{eq_non_interactive_R}, for the unit social event occurring in any social partition $n \in \{1, 2, ..., N_\theta\}$ at any past step $t_p \in \{1, 2, ..., T_h\}$, given variable $t \in \{T_h + 1, ..., T_h + T_f\}$, we have the \emph{Social Reverberation Strength}
\begin{equation}
    r_\SOC (t | n, t_p) = \frac{
        \left(\mathbf{R}_\SOC[n, t_p, t]\right)^2
    }{
        \sum_{t_x = 1}^{T_h} \left(\mathbf{R}_\SOC[n, t_x, t]\right)^2
    }.
\end{equation}

\FIG{fig_curve_R_social} visualizes several social reverberation curves.
For example, in case (a0), the ego pedestrian is walking along with its companion, which is located in the 6th social partition relative to the ego one, with other neighbors far away from them spatially.
In this case, we can regard that the main social trajectory-changing event for the ego pedestrian is to pay more attention to this companion and coordinate their future path together.
Interestingly, predicted social reverberation curves in \FIG{fig_curve_R_social} (a6) - (a8) (bounded with blue boxes) present noticeably different latency preferences for handling these social events.
Subfigure (a0) shows that the neighbor pedestrian starts to increase its speed (relative to the ego one) at step 3 (5th and 6th observed frames), while before that step this pair of agents share almost the same speed for walking together.
Correspondingly, event at such step 3 (the green curve) has been paid the most concentration at the beginning of the prediction period in partitions 6 - 8 where this neighbor's trajectory and social behaviors were located.
In addition, in these partitions, social events occurring at step 2 have been delayed till the end of the prediction period (indicating a large peaking latency $L_p(2)$), while event at the same step 2 has been consistently concerned with a vanishing latency $L_v(2) \approx 3$ steps (2.4s, from step 2 to 5) in other social partitions in cases (a1) to (a5).

To further verify the sensitivity of reverberation kernels in distinguishing latencies corresponding to different social events, as shown in \FIG{fig_curve_R_social} (b0) and (d0), we introduce the manual neighbor approach \cite{wong2024socialcircle+} to test how the kernel $\mathbf{R}_\SOC$ modifies latency properties for scenes where the social context has been manually \emph{intervened}.
For example, in \FIG{fig_curve_R_social} (a0), we have added a new companion to walk together with this ego pedestrian on its right-hand side, with a slightly higher velocity.
Compared to the original ones, social reverberation curves in the intervened social partitions (partitions 1 to 3, curves in (b1) to (b3)) present significant differences.
In detail, in partition 2, social event at step 4 has been suddenly delayed until step 9 to actually take effect, showing an overall social peaking latency $L_p(4)$ of about 5 steps (4.0s).
Similar modifications are also shown in \FIG{fig_curve_R_social} (d5) - (d8) after adding a new neighbor case (d0), where social event at step 1 has been forecasted to be more important to be handled, with other events less concerned in these intervened partitions.
These results could verify the sensitivity of the kernel $\mathbf{R}_\SOC$ for handling dynamics within social latencies under different interactive contexts when forecasting.

\begin{figure*}[tbp]
    \centering
    \includegraphics[width=1.0\linewidth]{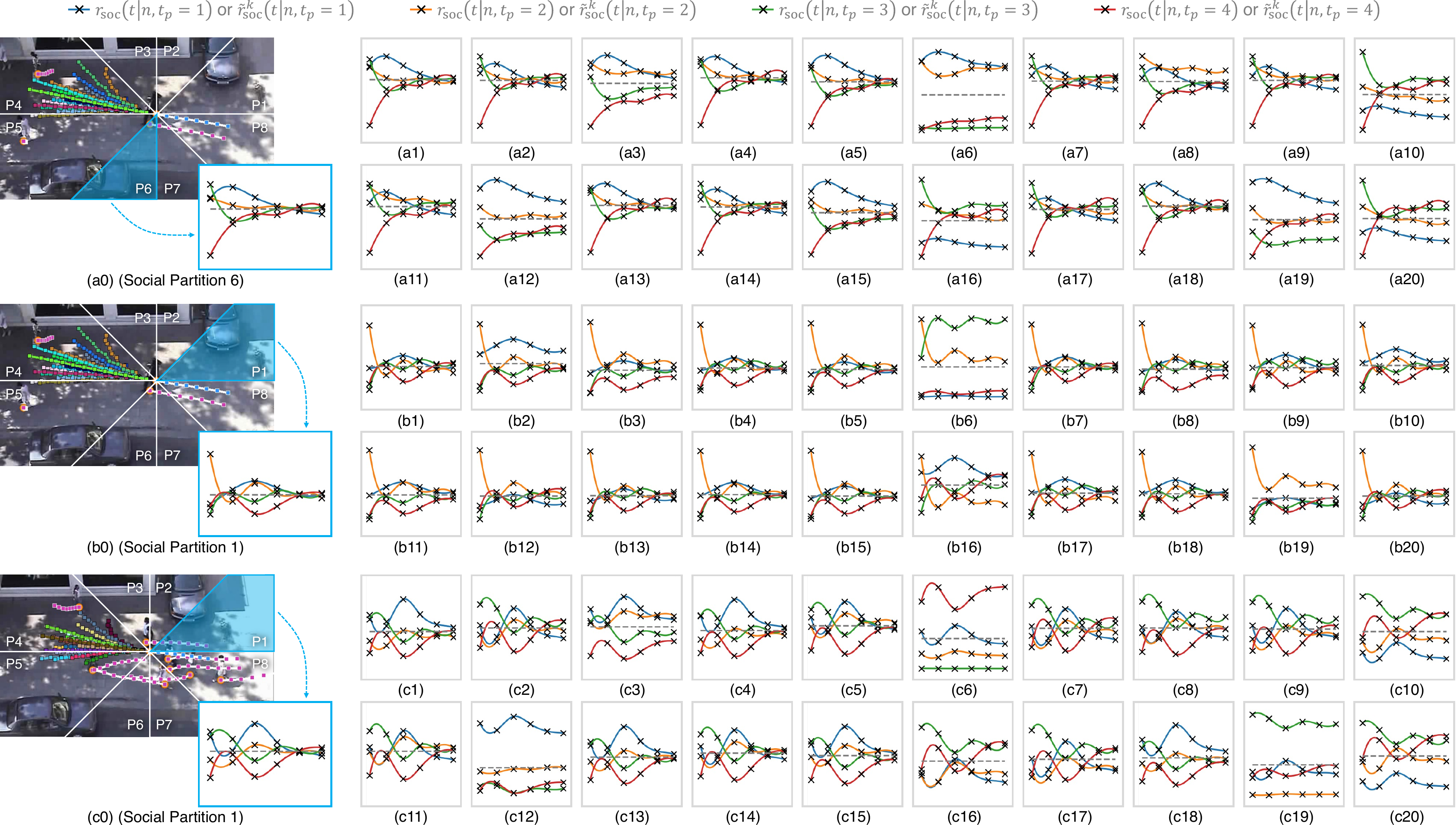}
    \caption{
        Examples of the original social reverberation strength $r_\SOC (t|n, t_p)$ ($t_p \in \{1, 2, 3, 4\}$, given specific social partition $n$ in blue boxes) and its $K_g = 20$ altered $\tilde{r}_\SOC^k (t|n, t_p)$ curves after applying the generating kernel $\mathbf{G}_\SOC$.
        $k$ refers to indices of subfigures in grey boxes.
    }
    \label{fig_curve_G_social}
\end{figure*}

\NEWHALFLINE
\subsubsection{Altered Social Reverberation $\tilde{r}_\SOC$}
Similar to the above non-interactive discussions, the social generating kernel $\mathbf{G}_\SOC$ alters the social reverberation kernel $\mathbf{R}_\SOC$ in $K_g$ ways for handling social events.
Given partition $n$ and step $t_p$, the altered social reverberation strength (curve) is defined as
\begin{equation}
    \tilde{r}_\SOC^k (t|n, t_p) = \frac{
        \left(\mathbf{R}_\SOC [n, t_p, t]~\mathbf{G}_\SOC [n, t_p, k]\right)^2
    }{
        \sum_{t_x = 1}^{T_h}
        \left(\mathbf{R}_\SOC [n, t_x, t]~\mathbf{G}_\SOC [n, t_x, k]\right)^2
    }.
\end{equation}

For the ease of presentation, we select several specific social partitions in \FIG{fig_curve_R_social} (a0) and (c0) to visualize how kernel $\mathbf{G}_\SOC$ alters the original reverberation curves in \FIG{fig_curve_G_social}.
Similar to discussions in \SECTION{sec_discussion_g_non}, for event occurring at step $t_p$ in social partition $n$, it shows that $\mathbf{G}_\SOC$ still may not significantly change a single reverberation strength $r_\SOC (t|n, t_p)$ for any future step $t$, but provides alternatives for considering and accounting for social events from all steps and positions (social partitions) simultaneously.
Notably, such alternatives are further associated with social events in their social partitions.

For example, in \FIG{fig_curve_R_social} (a6), events in partition 6 present considerably unique latencies than other social partitions.
Correspondingly, generating kernel $\mathbf{G}_\SOC$ recombines events at these four historical steps and predicts latency alternatives with a more ``adventurous'' attempt in \FIG{fig_curve_G_social} (a0) - (a20), where the focus of all events has been thoroughly rearranged compared to the original curve in case (a0).
As a comparison, for the social partition 1 in case (b0), where almost no significant social event occurred, more ``conservative'' alterations have been provided by the same generating kernel $\mathbf{G}_\SOC$.
Among cases in \FIG{fig_curve_G_social} (b1) - (b20), only a few ones (like (b2) and (b6)) present noticeable rearrangements.
Compared to the non-interactive $\mathbf{G}_\NON$, this difference verifies that social interactions themselves have become a main consideration even for agents with different latency preferences, while the $\mathbf{G}_\SOC$ still has the ability to notice and distinguish such latencies.

\begin{figure}[tbp]
    \centering
    \includegraphics[width=1.0\linewidth]{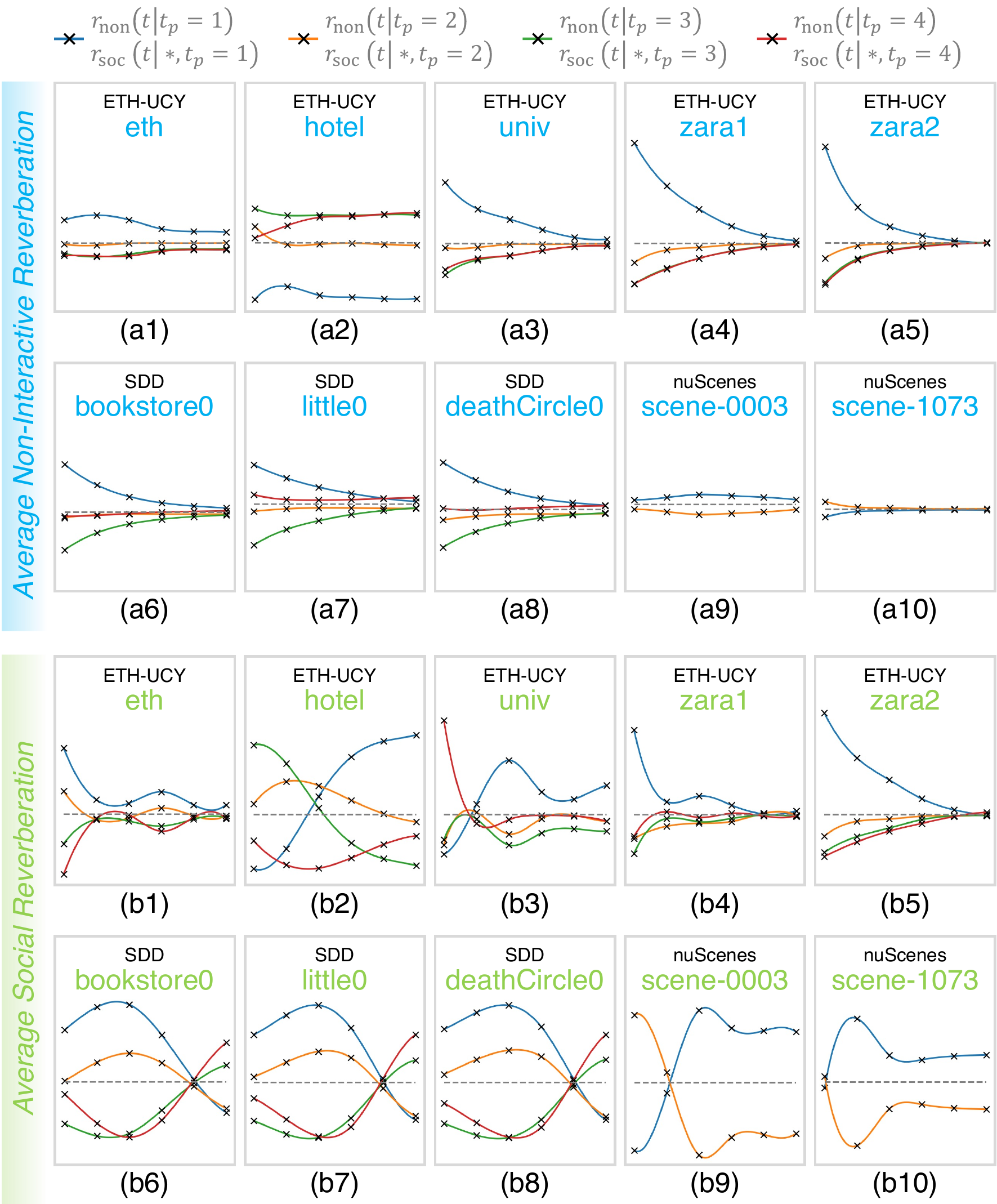}
    \caption{
        Examples of average reverberation curves of all agents and all social partitions (only for social reverberation) on several datasets.
        $\{T_h, T_f\}$ is set to $\{4, 6\}$ on ETH-UCY and SDD sets, while set to $\{2, 5\}$ on nuScenes.
    }
    \label{fig_curve_average}
\end{figure}

\NEWHALFLINE
\subsubsection{Latencies across Agents and Datasets}

It can be seen from the above figures that reverberation curves vary significantly across agents.
\FIG{fig_curve_average} visualizes the average of reverberation curves on several datasets, representing agents' average latency preferences\footnote{
    See further discussions on latencies of motion prediction in \APPENDIX{F}.
} correspondingly.

For non-interactive cases, we observe that event at the first historical step might be specifically concentrated on most sets.
Like \FIG{fig_curve_average} (a3) - (a8), the average curve shows that agents are forecasted to take at least a peaking latency $L_p(1)$ of 4 more steps (3.2s) to finish handling self-sourced trajectory-changing events that happened on the first observation step.
This can be interpreted as most agents may keep their motion status as those they presented from the beginning, meaning that such status may remain for about 8 steps (6.4s) since step 1 before vanishing ($L_v(1) \approx 8$ steps).
During this period, events at the following steps may gradually be considered when planning trajectories.
Curves of several other datasets like \FIG{fig_curve_average} (a2) (a9) (a10) show a different trend that events at almost all historical steps are considered equally contributed.
We can infer that agents in these scenes may prefer to maintain their motion status all the time, like walking or driving alongside or around straight lines.
Therefore, their peaking latencies $L_p$ for handling non-interactive events could be considered as 0, \IE, taking effect immediately.

Latencies for handling social trajectory-changing events present more diverse changes across datasets.
For example, in \FIG{fig_curve_average}, it predicts most agents may first handle events on step 3 in the near future (last for about 2 steps / 1.6s), during which event at historical step 1 has been delayed.
This event will finally start to be handled by agents at step 8, long after 7 steps (5.6s) since occurrence.
Differently, in SDD cases in \FIG{fig_curve_average} (b6) - (b8), it predicts agents to firstly handle social events on steps 1 and 2 until step 7, while events on steps 3 and 4 are delayed for about 4 - 5 steps (3.2 - 4.0s) before being handled since their happening.
In short, these curves indicate that ETH-UCY exhibits relatively tighter latency concentration than SDD, while nuScenes shows the largest dispersion.
Compared to the average non-interactive curves that present similar latency trends across datasets, such varying social curves could verify the dynamic-modeling capability of the proposed \MODELSHORT~for modeling agents' varying latency preferences for handling social events.

Also, it should be noted that nuScenes curves in \FIG{fig_curve_average} (a9) (a10) (b9) (b10) are obtained by only considering 2 historical steps for the future 5 steps.
These curves indicate that distinguishing latency preferences could still be learned, even though we have extremely limited observations for the vehicle dataset nuScenes, highlighting the robustness of the reverberation transform and its possible applications in intelligent transportation systems.

\NEWHALFLINE
\subsubsection{Summary of Latencies}
Reverberation transform includes two complementary and trainable kernels.
The reverberation kernel $\mathbf{R}$ captures event-level latencies, typically exhibiting \emph{single-peaked} curves that intuitively reflect the temporal offset to perceive, respond to, and complete trajectory-changing events since their occurrences, enhancing the interpretability of how agents respond to these events when planning trajectories.
The generating kernel $\mathbf{G}$ encodes agents' global latency preferences and randomness, offering alternative arrangements of multiple events.
By combining these kernels, the reverberation transform enables the \MODELSHORT~model to adaptively characterize and forecast heterogeneous and dynamic latency preferences when forecasting trajectories.

\subsection{Further Discussions}

\begin{figure}[tbp]
    \centering
    \includegraphics[width=1.0\linewidth]{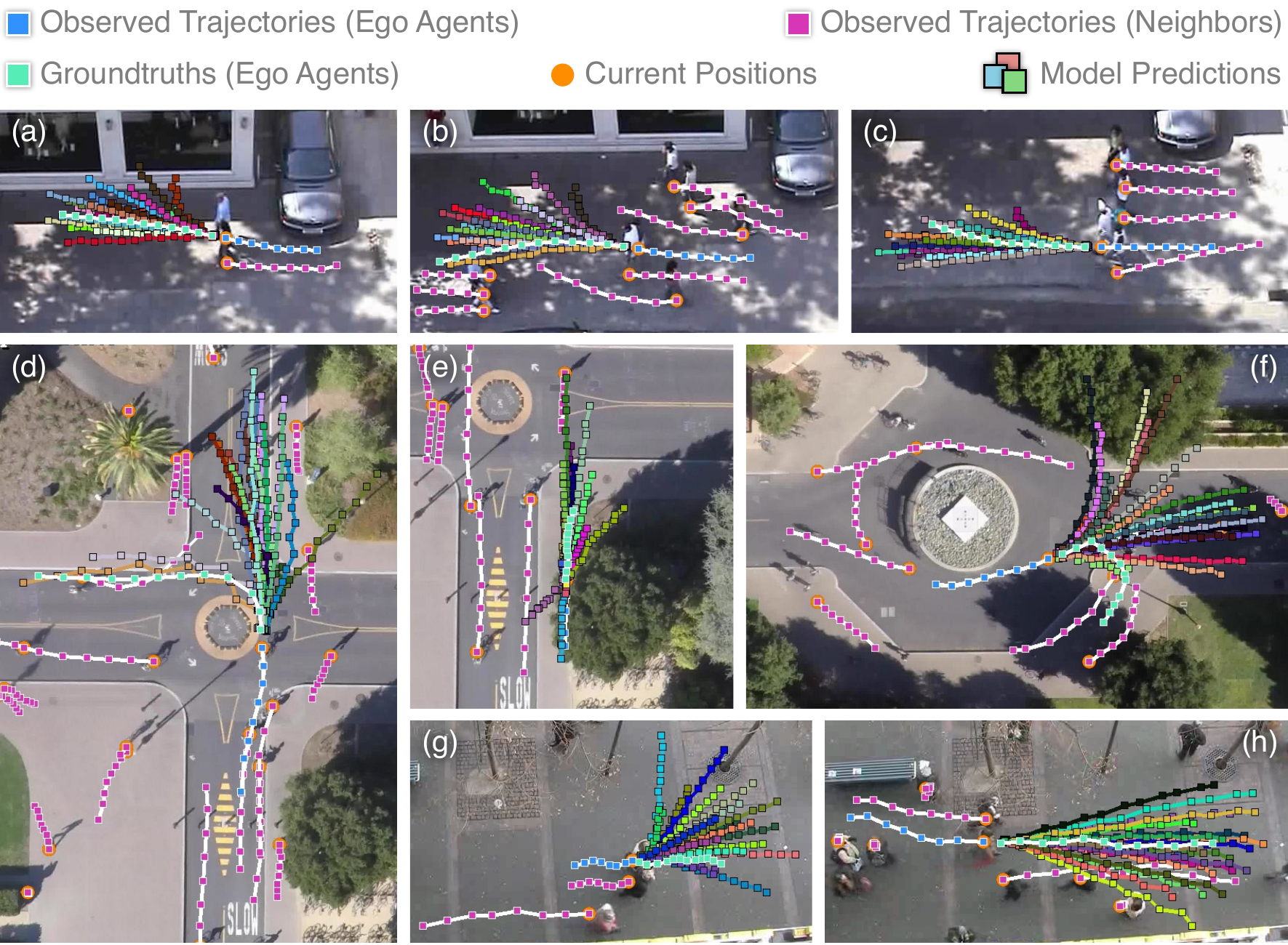}
    \caption{
        Examples of visualized \MODELSHORT~predictions (20 generations).
    }
    \label{fig_vis}
\end{figure}

\NEWHALFLINE
\subsubsection{Social Interactions}
\FIG{fig_vis} visualizes several \MODELSHORT~predictions.
It can be seen that social interactions have been considered in the forecasted trajectories.
For example, in \FIG{fig_vis} (a), the randomness of forecasted trajectories mainly lies in the right-hand side of the ego pedestrian, since there is a walking-together companion on the other side.
In \FIG{fig_vis} (b), most forecasted trajectories have been bent to avoid the upcoming pedestrian group.
Also, distinct turn choices have been forecasted in roundabout scenes in \FIG{fig_vis} (d) and (f), verifying its generalizability in complex scenes.
Though some predictions may be considered not suitable for the context, like crossing the border of roads in (d) and (f), such predictions could still prove the potential of the proposed \MODELSHORT~model in forecasting trajectories with only egos' and neighbors' observed trajectories, without the dependency on additional information like segmentation images or language models.

\begin{figure}[tbp]
    \centering
    \includegraphics[width=1.0\linewidth]{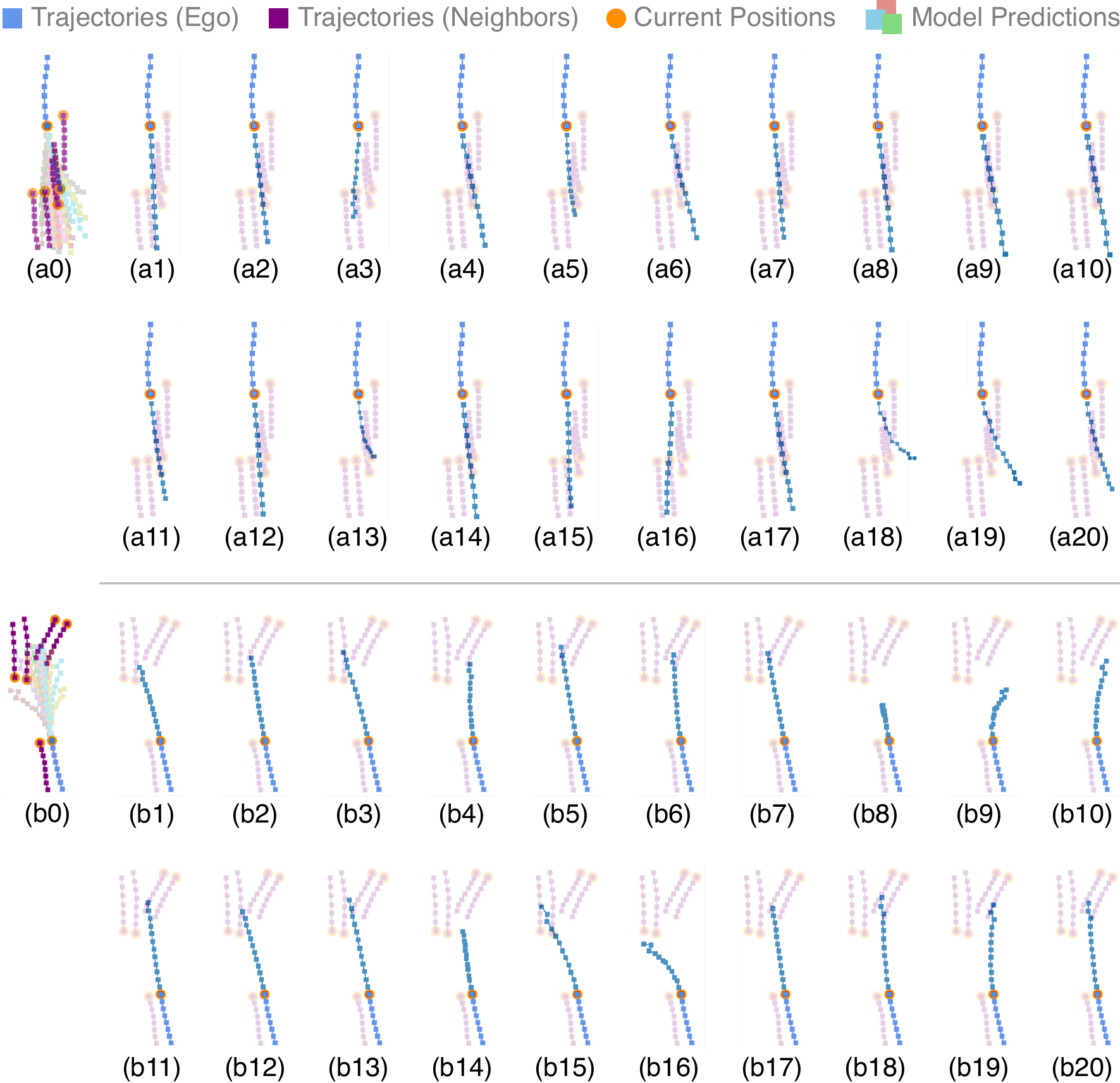}
    \caption{
        \MODELSHORT~predictions under different latency combinations ($K_g = 20$).
        (a0) and (b0) include all predictions, while other subfigures are obtained through their corresponding $k$th column of generating kernels $\mathbf{G} \in \mathbb{R}^{T_h \times K_g}$.
    }
    \label{fig_vis_channel}
\end{figure}

\NEWHALFLINE
\subsubsection{Generalizability}
\label{sec_gen}

\FIG{fig_vis_channel} visualizes predictions forecasted through different columns of generating kernels ($K_g = 20$), presenting unique preferences for handling interactive behaviors.
For example, in case (b0), the ego pedestrian is walking with a companion, with another duo approaching face-to-face.
Though we have mentioned that kernel $\mathbf{G}$ may not significantly modify latency preferences of a single historical frame, subfigures (b1) - (b20) have presented quite different prediction \emph{styles} \MSNCITE, like keeping walking and ignoring the upcoming duo in (b1), slightly avoiding them in (b4), noticing and avoiding them as quickly as possible in (b8) - (b10), and even heading for them directly in (b16) - (b17).
This implies that the way agents jointly consider the importance of historical events and selectively combine and handle their latencies could lead to totally different future trajectory-choices, while the generating kernels $\mathbf{G}$ have the ability to capture and learn such alternatives\footnote{
    This is actually associated with the low-rank properties of the proposed reverberation transform.
    See further discussions in \APPENDIX{A}.
} and finally generate interpretable while acceptable trajectories.
We have provided further discussions in \APPENDIX{G} regarding the model's generalizability, particularly in generating diverse or deterministic trajectories and its adaptability to trajectories with different sampling rates.

\begin{figure}[tbp]
    \centering
    \includegraphics[width=1.0\linewidth]{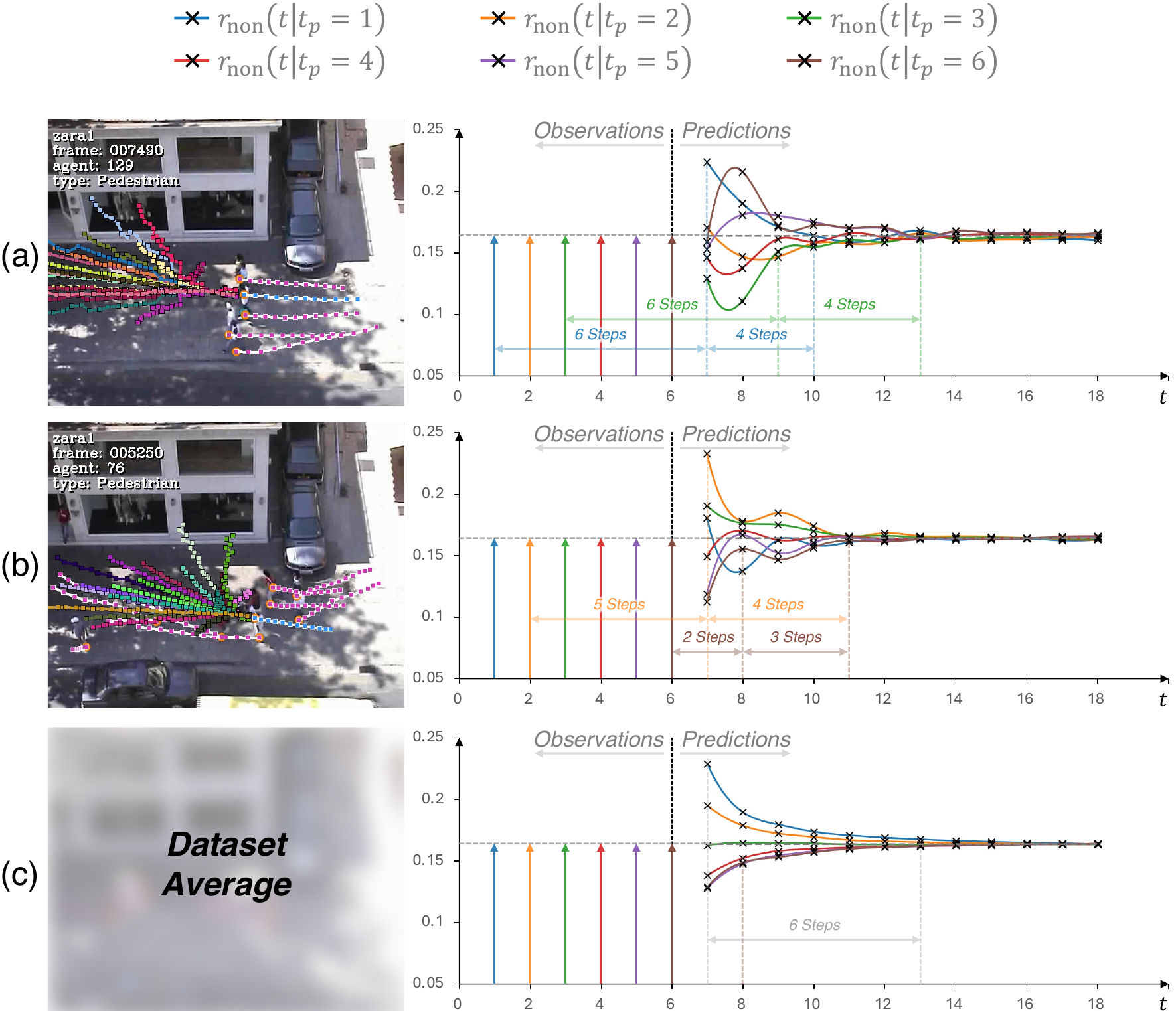}
    \caption{
        Examples of non-interactive reverberation strength $r_\NON (t|t_p)$ under the long-term prediction setting $\{T_h, T_f\} = \{6, 12\}$ ($t_p \in \{1, 2, 3, 4, 5, 6\}$).
    }
    \label{fig_curve_longterm}
\end{figure}

\NEWHALFLINE
\subsubsection{Latencies under Long-term Prediction Settings}
\label{sec_longterm}
Discussions in \SECTION{sec_discussion_r_non,sec_discussion_r_soc} indicate that pedestrians are forecasted to have a latency for up to 6 - 8 steps (4.8 - 6.4s) to start handling a trajectory-changing event, during the whole $T_h + T_f = 10$ steps (20 frames or 8.0s in total) of the prediction period.
We now discuss whether such latency varies with the prediction period or remains a constant for specific prediction scenarios, taking the $\{T_h, T_f\} = \{6, 12\}$ (36 frames or 14.4s in total) long-term setting as an example.
In \FIG{fig_curve_longterm} (a) and (b), it can be seen that agents' latency preferences are still similar to the former curves we have discussed.
For example, step 3 in case (a) has been delayed for about 6 steps (4.8s) to take effect at future step 9, and takes another 4 steps (3.2s) to keep concentrating on it, while similarly, step 2 in case (b) takes about 4 steps (3.2s) for the pedestrian to handle it since the first prediction step (step 7).
We also visualize the average curve of this dataset in \FIG{fig_curve_longterm} (c), under this long-term setting.
It shows that all historical events are predicted to be solved only after about 6 steps (4.8s) after the first prediction step, and then they will be paid almost the same attention till the end of the prediction period, presenting similar trends as former short-term results shown in \FIG{fig_curve_average}.
We can conclude that such captured and forecasted latencies are associated with pedestrians' walking or social preferences, rather than the simple data-fitting, further providing insightful interpretations of pedestrian behaviors beyond trajectory prediction.

\NEWHALFLINE
\subsubsection{Failure Cases}
Despite the above discussions, we still found some examples that the proposed \MODELSHORT~could not handle.
These examples contain two kinds, including sudden, unexpected latencies, and the low-frequency latency confusion problem due to the use of the Haar transform.
Specifically, the learning of the reverberation kernel $\mathbf{R}$ is based on historical statistics, and \MODELSHORT~cannot make such predictions for such instantaneous responses without any historical signs, such as a sudden change of movement direction during the prediction period.
Although the majority of current methods cannot make reasonable predictions in such a situation, this is actually part of agents' latency modes, which remains an open problem for future research.
On the other hand, when the trajectories are extremely smooth (more low-frequency portions), the Haar transform is then unable to extract enough high-frequency features to localize the temporal onset of trajectory-changing events.
This is due to the properties of the Haar transform, which makes it difficult for reverberation kernels to distinguish when an event occurs, as we have shown with nuScenes ($K=5$) that does not perform as well as \REMODEL.
We have discussed in depth in the \APPENDIX{D} the relationships between the choice of trajectory spectrums and the reverberation transform, attempting to provide alternatives in these specific prediction scenarios.
Also, as elaborated in \SECTION{sec_introduction}, although the reverberation kernels provide interpretable reverberation curves and latencies, real-world human reaction times or behavior delays are still unobservable or hard to label.
In other words, the learned latencies are data-induced temporal responses rather than exact ground-truth reaction times, which presents us with additional challenges in modeling realistic biological reaction times or behavioral delays in the future.

\section{Conclusion}

In this manuscript, we propose the Reverberation Transform and the corresponding \MODEL~(\MODELSHORT~for short) trajectory prediction model to explicitly model and forecast agents' latencies for handling trajectory-changing events when forecasting trajectories.
The proposed reverberation transform includes two trainable reverberation kernels.
The reverberation kernel $\mathbf{R}$ characterizes event-level latency responses while the generating kernel $\mathbf{G}$ simulates stochastic latency variations or randomness, providing a principled way to bridge agents' observations to the future while explicitly considering latencies for handling all trajectory-changing events.
The proposed \MODELSHORT~model not only achieves competitive prediction accuracy across pedestrian and vehicle datasets (ETH-UCY, SDD, and nuScenes), but also demonstrates strong latency-interpretability through the visualization of non-interactive and social reverberation curves.
Ablation studies and discussions further verified the necessity of each component and the complementary effects of forecasting non-interactive and social latencies, highlighting the importance of explicitly modeling latency dynamics in trajectory prediction, offering new insights into how agents anticipate, react, and adapt to diverse trajectory-changing events.
Beyond the immediate gains in accuracy and interpretability, the proposed reverberation transform also provides a general theoretical perspective for quantifying latency dynamics in sequential decision-making.
In future work, we plan to extend this framework to multi-modal perception inputs and more diverse spatio-temporal domains, such as human-robot interaction and autonomous driving.
Furthermore, to address the failure cases caused by the properties of the used Haar transform when processing extremely smooth or low-frequency trajectories, our future work will explore adaptive or learnable wavelets \cite{tang2003skeletonization} and multi-resolution spectral inputs, thereby enabling the reverberation transform to seamlessly maintain both sharp representation and responses to high-frequency events as well as smooth low-frequency adaptabilities of its input sequences.
We believe that latency-aware forecasting will play a crucial role in improving both the safety and adaptability of intelligent systems.

\bibliographystyle{IEEEtran}
\bibliography{ref.bib}

\begin{IEEEbiography}[{
    \includegraphics[width=1in,
                     height=1.25in,
                     clip,
                     keepaspectratio]{
        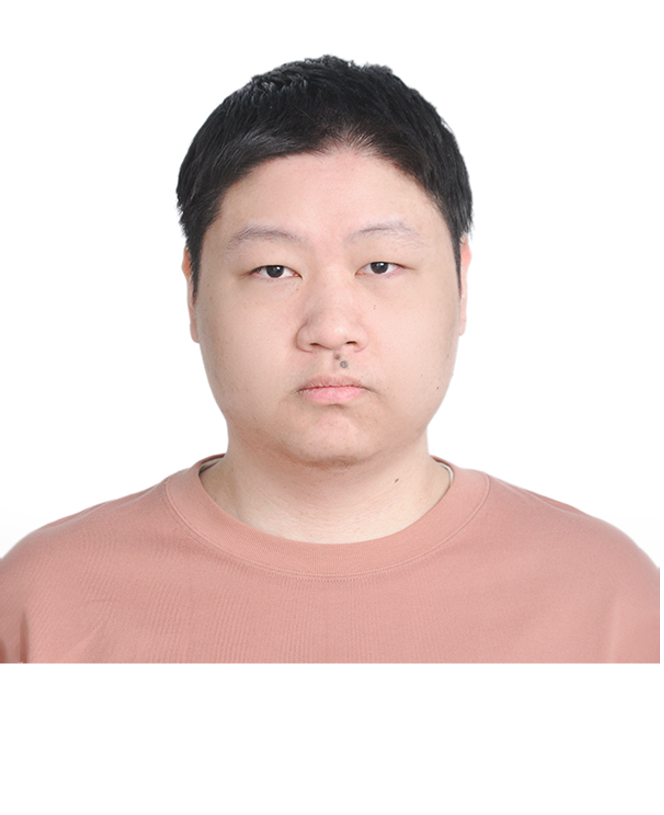
}}]{Conghao Wong}
    received the master's degree from Huazhong University of Science and Technology, Wuhan, in 2022, where he is currently pursuing the Ph.D. degree.
    His research interests include computer vision and pattern recognition.
\end{IEEEbiography}

\begin{IEEEbiography}[{
    \includegraphics[width=1in,
                     height=1.25in,
                     clip,
                     keepaspectratio]{
        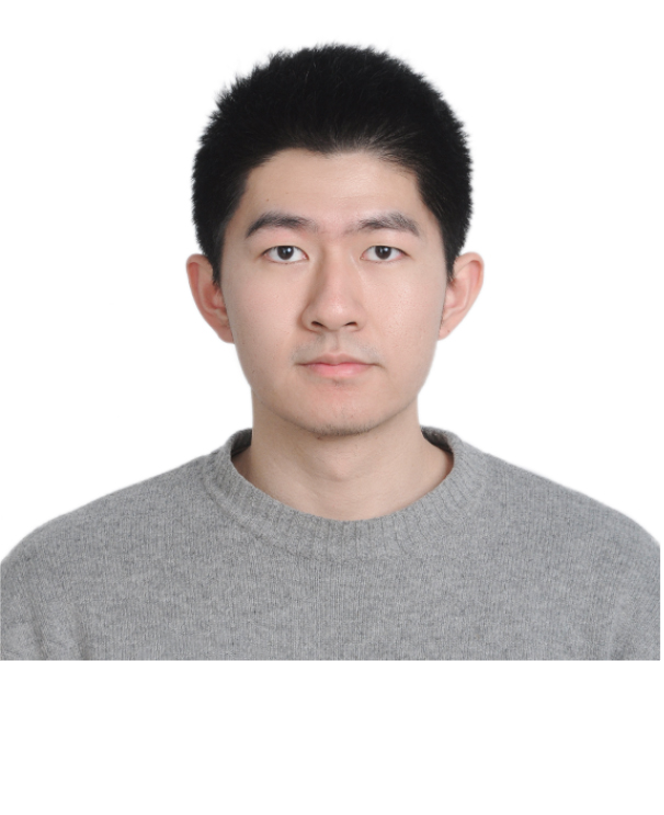
}}]{Ziqian Zou}
    received the master's degree from Huazhong University of Science and Technology, Wuhan, in 2025, where he is currently pursuing the Ph.D. degree.
    His research interests include pattern recognition and video understanding.
\end{IEEEbiography}

\begin{IEEEbiography}[{
    \includegraphics[width=1in,
                     height=1.25in,
                     clip,
                     keepaspectratio]{
        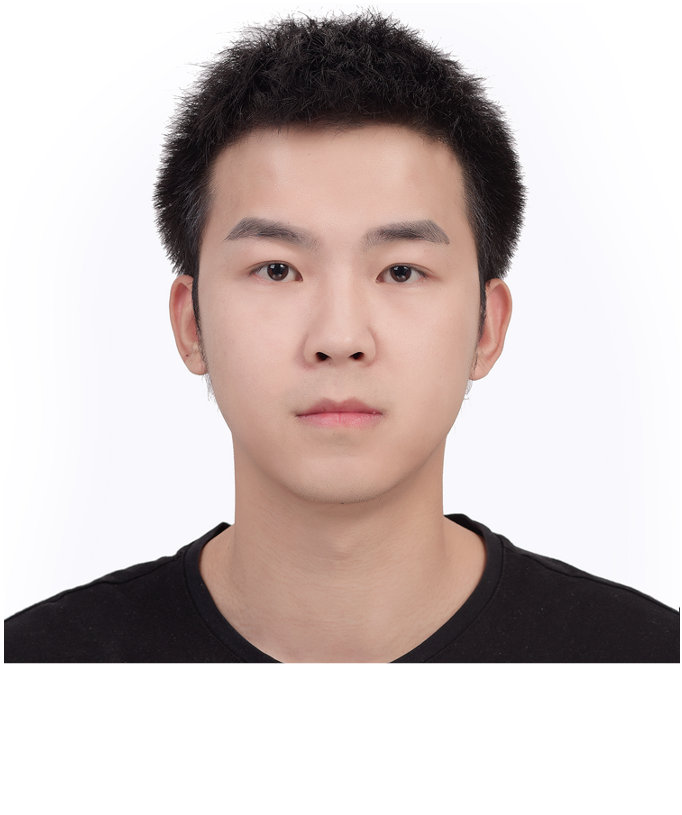
}}]{Beihao Xia}
    received his Ph.D. degree in Huazhong University of Science and Technology, Wuhan, China, in 2023.
    His research interests include trajectory prediction, behavior analysis, and understanding.
\end{IEEEbiography}

\begin{IEEEbiography}[{
    \includegraphics[width=1in,
                     height=1.25in,
                     clip,
                     keepaspectratio]{
        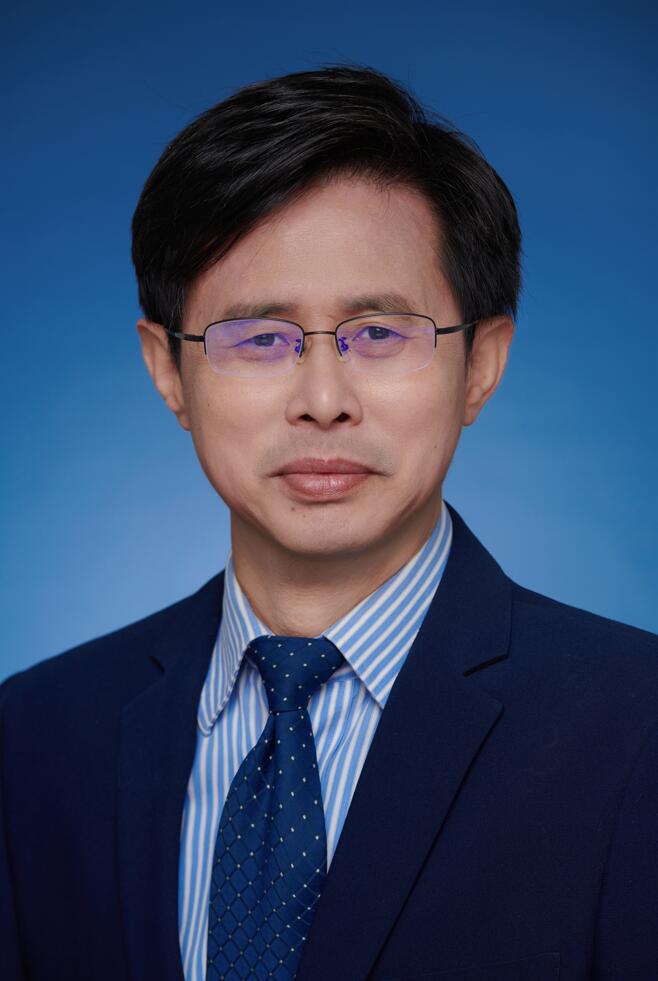
}}]{Xinge You}
    (Senior Member, IEEE) is currently a Professor in Huazhong University of Science and Technology. 
    He received his Ph.D. degree from the Department of Computer Science, Hong Kong Baptist University in 2004. 
    His work has appeared in 200+ publications, such as IEEE TPAMI, TIP, TNNLS, CVPR, ECCV, ICCV. 
    He served/serves as an Associate Editor of the \textit{IEEE TCyb}, \textit{TSMCA}. 
    His research interests include wavelet analysis, pattern recognition, machine learning, and computer vision. 
\end{IEEEbiography}

\vfill

\end{document}



\documentclass[../../paper.tex]{subfiles}

\begin{document}

\appendices

\documentclass[../../paper.tex]{subfiles}

\begin{document}

\section{Properties of Reverberation Transform}
\label{sec_appendix_rev}

We now discuss potential theoretical properties of the proposed reverberation transform, providing insights into why it could model and forecast latencies.

\NEWHALFLINE
\subsubsection{Formula Definition of Reverberation Transform}

Based on our definitions in the Method section, we now formalize the reverberation transform as an operator.
The reverberation transform acts on the observed sequential similarity tensor $\mathbf{F} \in \mathbb{R}^{T_h \times T_h \times D}$, computed from feature embeddings $\mathbf{f} \in \mathbb{R}^{T_h \times D}$.
Its output is a structured rehearsal field of shape $K_g \times T_f \times D$.
For each feature dimension $d \in \{1, 2, ..., D\}$, the reverberation transform is defined as
\begin{equation}
    \label{eq_appendix_rev}
    \overline{\mathbf{F}}_{::d}
        = \mathcal{R}_{(\mathbf{R}, \mathbf{G})} [\mathbf{F}_{::d}]
        := \mathbf{G}^\top \mathbf{F}_{::d} \mathbf{R} \in \mathbb{R}^{K_g \times T_f}.
\end{equation}

This mapping is linear in $\mathbf{F}_{::d}$ and bilinear in $(\mathbf{R}, \mathbf{G})$.
Therefore, $\mathcal{R}_{(\mathbf{R}, \mathbf{G})}$ is a bounded linear operator from the space of sequential similarities to a low-rank subspace determined by $\mathbf{R}$ and $\mathbf{G}$.
This formalization enables the analysis of its linearity, expressive subspace, and rank-constrained structure in the following sections.

\NEWHALFLINE
\subsubsection{Linear Property}
The reverberation transform is a bilinear operator for the sequential similarity, and it shares a similar linearity property to the Fourier transform.
\EQUA{eq_appendix_rev} is linear-additive for any input $\mathbf{F}$, meaning that for any inputs $\mathbf{F}_1$ and $\mathbf{F}_2$, given real numbers $\alpha$ and $\beta$, we have
\begin{equation}
    \begin{aligned}
        \left(\overline{
            \alpha \mathbf{F}_1 + \beta \mathbf{F}_2
        }\right)_{::d}
        &= \mathbf{G}^\top \left(
            \alpha \mathbf{F}_1 + \beta \mathbf{F}_2
        \right)_{::d} \mathbf{R} \\
        &= \alpha \mathbf{G}^\top \left(\mathbf{F}_1\right)_{::d} \mathbf{R} + 
        \beta \mathbf{G}^\top \left(\mathbf{F}_2\right)_{::d} \mathbf{R} \\
        &= \alpha \left(\overline{\mathbf{F}_1}\right)_{::d} + \beta \left(\overline{\mathbf{F}_2}\right)_{::d}.
\end{aligned}
\end{equation}

This linear property actually provides us support for linearly decomposing latencies into non-interactive and social portions in the proposed \MODELSHORT~model, and also ensures the event-level decoupled latency prediction when forecasting trajectories.
It makes such a transform in line with human intuitions for separately handling multiple trajectory-changing events, \IE, people tend to process independent events separately and then superpose their effects when planning trajectories or future behaviors.

\NEWHALFLINE
\subsubsection{Approximate Completeness}
According to \EQUA{eq_appendix_rev}, kernels $\mathbf{R}$ and $\mathbf{G}$ span a subspace of agents' possible latency mappings.
In detail, $\mathbf{R}$ establishes \emph{biases} of agents to schedule historical trajectory determinants in the future, while $\mathbf{G}$ recombines such biases and provides $K_g$ alternatives of latencies.

For any $\mathbf{F}$ (actually $\mathbf{F}_{::d}$, omit subscripts for clarity), we have
\begin{align}
    \mathrm{Row}\left(
        \mathbf{G}^\top \mathbf{F} \mathbf{R}
    \right) &\subseteq
    \mathrm{Row}\left(
        \mathbf{G}^\top
    \right), \\
    \mathrm{Col}\left(
        \mathbf{G}^\top \mathbf{F} \mathbf{R}
    \right) &\subseteq
    \mathrm{Col}\left(
        \mathbf{R}
    \right).
\end{align}
Thus, the transformed $\overline{\mathbf{F}}$ lies in a bi-factor subspace limited by the row space of $\mathbf{G}^\top$ and the column space of $\mathbf{R}$.
Since $\mathrm{rank}(\mathbf{R}) \leq \min \{T_h, T_f\}$ and $\mathrm{rank}(\mathbf{G}) \leq \min \{T_h, K_g\}$, this subspace cannot exceed
\begin{equation}
    \mathrm{rank}(\mathbf{R}) \mathrm{rank}(\mathbf{G}) \leq \min \{T_h, T_f\} \min \{T_h, K_g\},
\end{equation}
unless $\mathrm{rank}(\mathbf{R}) = T_f$ and $\mathrm{rank}(\mathbf{G}) = K_g$.
Considering that $T_f$ might be larger than $T_h$ under current common trajectory prediction settings, and $K_g$ could also be larger than $T_h$ when forecasting, the reverberation transform cannot be considered strictly complete.
Also, we use $\tanh$ to bound both reverberation kernels.
Such bounds do not change rank in the algebraic sense but act as effective regularizers, making it an approximate rather than exact representation of extreme mappings.

This rank limitation explains why the transform spans a rich but structured subspace of latency mappings: rich enough (when $\mathbf{R}$ and $\mathbf{G}$ have higher rank) to approximate the finite-complexity latency patterns in real-world scenes, and structured enough (row or column space constraints) to regularize learning while preserving interpretability.

In short, approximate completeness follows from the fact that the reverberation transform has the ability to represent any matrix in a large bi-factor subspace of dimension up to $\mathrm{rank}(\mathbf{R}) \mathrm{rank}(\mathbf{G})$, which can be made sufficiently expressive by learning kernels $\mathbf{R}$ and $\mathbf{G}$ to be \emph{almost} full-rank under the experimental $K_g, T_f, T_h$ settings.
Especially, this approximate completeness ensures that the proposed \MODELSHORT~model can capture heterogeneous latency patterns without resorting to excessively high-dimensional parameterizations.

\NEWHALFLINE
\subsubsection{Low-Rank \& Compression}
According to \EQUA{eq_appendix_rev}, for any feature dimension $d \in \{1, 2, ..., D\}$, the rank of each transformed slice $\overline{\mathbf{F}}_{::d}$ is bounded by
\begin{equation}
    \label{eq_rank_transform}
    \mathrm{rank}\left(
        \overline{\mathbf{F}}_{::d}
    \right) \leq \min \left\{
        \mathrm{rank} (\mathbf{G}),
        \mathrm{rank}\left(
            \mathbf{F}_{::d}
        \right),
        \mathrm{rank} (\mathbf{R})
    \right\}.
\end{equation}
Under current common trajectory prediction settings, variations $T_h, T_f, K_g$ mostly satisfy
\begin{equation}
    K_g > T_f \geq T_h.
\end{equation}
Specifically, sometimes $K_g$ may be greater than $T_f$.
Therefore, \EQUA{eq_rank_transform} has become
\begin{equation}
    \mathrm{rank}\left(
        \overline{\mathbf{F}}_{::d}
    \right) \leq \min \left\{
        T_h, T_f
    \right\}.
\end{equation}

This means that the rank of the transformed $\overline{\mathbf{F}}$ is independent of the number of latency generations $K_g$ in real-world applications of trajectory prediction.
In other words, the effective information dimension of the transformed $\overline{\mathbf{F}}$ remains bounded by both $T_h$ and $T_f$, and mostly only by $T_h$ in practice, indicating a low-rank and compression property of the reverberation transform.

In detail, it compresses the original similarity representation $\mathbf{F}$ into a subspace of dimension at most $T_h$, and the $K_g$ output reverberation alternatives are different recombinations within this same subspace.
This design produces apparent redundancy when $K_g \gg T_h$, but the redundancy is intentional-enabling diverse candidate predictions while ensuring that all outputs are constrained to a structured low-dimensional manifold.
The low-rank property also plays as an inherent regularizer, which prevents the network from overfitting, even when $K_g$ is very large.
It also makes the storage and computation more efficient, since the reverberation transform can be factorized through only two kernels $\mathbf{R}$ and $\mathbf{G}$ without explicitly materializing the full $K_g \times T_f \times D$ feature matrix.
Thus, this low-rank/compression property could ensure the \MODELSHORT~to generate diverse latencies and final predictions, while keeping their stabilizability especially in high-dimensional trajectory prediction scenes.

\end{document}

\documentclass[../../paper.tex]{subfiles}

\begin{document}

\section{Further Interpretations of the Reverberation Transform from the System-Theory's View}

\subsection{Role of the Reverberation Transform}

Unlike directly combining the $T_h$ temporal features based on attention scores or other fusion methods, the weights in the reverberation kernel $\mathbf{R}$ present explicit response characteristics, where the $T_f$ elements in the specific $h_0$th row of the kernel $\mathbf{R}$ (for $h_0 \in \{1, 2, ..., T_h\}$ and $f \in \{1, 2, ..., T_f\}$, denoted as $w(f|h_0)$) are computed only based on the state of the corresponding $h_0$th observation step.
Denote the representation of all inputs (whether intentions or interactions) occurring at the $h$th observation step as $e(h)$, the process to predict the final representation at the $f$th future step (denoted as $Y(f)$) can be formulated as
\begin{equation}
    \label{eq_response_r3_1}
    Y(f) = \sum_{h = 1}^{T_h} e(h) w(f|h).
\end{equation}

This actually takes a similar mathematical form as the impulse response of a linear system.
For a discrete input signal $x[t]$, given the system's impulse response $k[t]$ (the output in response to the unit impulse $\delta[t]$, suppose this response is time-invariant), then at any future step $n$, the system's output $y[n]$ can be formulated as
\begin{equation}
    \label{eq_response_r3_2}
    y[n] = \sum_{t = 0}^{n} x[t] k[n - t].
\end{equation}
Here, comparing \EQUA{eq_response_r3_1} and \EQUA{eq_response_r3_2}, it can be seen that each row $\mathbf{R}_{h:} \in \mathbb{R}^{T_f}$ in the reverberation kernel (the $w(f|h)$ in \EQUA{eq_response_r3_1}) plays a role similar to the ``impulse response'' to inputs occurring at the $h$th historical step, and the \EQUA{eq_response_r3_1} has become an interpretation of the trajectory prediction process as the linear time-varying system (we regard that each row of the reverberation kernel to not be exactly the same, \IE, agents are allowed to react differently to observed trajectory events at different historical steps).
Under this interpretation, the $\mathbf{R}$ matrix can be regarded as a set of row vectors, where each row vector directly corresponds to the ``unit impulse response'' of the event occurring solely at the $h$th historical time step (rather than depending on all other events).
Notably, the event here refers specifically to the discrete-time unit impulse $\delta[n-h]$, where
\begin{equation}
    \delta[n-h] = \left\{\begin{aligned}
        &1, \quad n = h, \\
        &0, \quad \mbox{otherwise}.
    \end{aligned}\right.
\end{equation}

Under these interpretations, the core of the reverberation transform is to learn a set of such ``unit impulse responses'' as the $\mathbf{R}$ matrix, thereby describing the time-varying response (\IE, how future trajectories would change) to different past events on the observed trajectories.
For these $T_f$ elements in each impulse response vector, the further response curve could better reflect the impact of the event that happens at the $h$th step ($\delta[n-h]$) on the final trajectories, which shows how the system's response changes or varies over time after this event.
Therefore, our defined latencies as well as reverberation curves do not strictly correspond to human reaction times or action delays, but rather indicate how long the system output (trajectories) will start to be altered or stabilized or finally vanish after the specific (interactive) event occurs.

\subsection{Ideal Events}

In practice, we do not use any event detector, change-point criterion, or further mechanism to separate discrete events from ordinary temporal samples.
As we defined in the reverberation transform, one trajectory-changing event is literally some events that may cause the changes of agents' own future trajectory once after that event has been settled.
This type of event is considered additive and may include joint products resulting from multiple different causes.
These events are described directly using feature vectors, typically with the form of a temporal series that varies across different temporal steps, where one event may last for either only one step or multiple observation steps.
The reverberation transform is built upon such clean and ideal trajectory-changing events, free from any extraneous semantic meanings or limitations, as represented by $\mathbf{f} \in \mathbb{R}^{T_h \times D}$ in the beginning of Sec. III-A1.
In this context, $\mathbf{f}$ is considered to be a direct representation of certain specific input event(s) occurring during the observation period.

Given the input (also known as the \emph{excitation}, denoted as $x[h]$) over some specific observed temporal period $h \in \mathcal{T}_h$, suppose the system's responses $w[n, h]$ to unit impulse $\delta[n - h], \forall h \in \mathcal{T}_h$ ($n$ denotes the temporal variable, and $h$ indexes where the unit impulse is temporally located) at these corresponding temporal steps are known, the system's output at any future step ($n \in \mathcal{T}_f$) can still be calculated by performing a temporal convolution (note that this is slightly different from convolution used in CNNs) of the excitation and the time-variant unit impulse response:
\begin{equation}
    \label{eq_response_r3_event_conv}
    y[n] = \sum_{h \in \mathcal{T}_h, h \leq n} x[h] w [n, h].
\end{equation}

Here, the sequential representation $x[h]~(h\in\mathcal{T}_h)$ (\IE, the $\mathbf{f} \in \mathbb{R}^{T_h \times D}$ in our definitions) is a uniform instance of the referred ``trajectory-changing events'' when constructing the \MODELSHORT~trajectory prediction model.
Specifically, these events are represented at the temporal-sequence-like feature-level, and they play a role similar to the excitation of the system, where the system's output (trajectories) could be changed after the occurrence of these events ($n \geq h$).
Correspondingly, each row (the $w[:, h]$) in the reverberation kernel $\mathbf{R}$ plays the role of a learnable time-variant response to the unit impulse $\delta[h]$ occurring at the corresponding historical step $h$.
Thus, our considered events here could be any input feature sequences, which play as the excitations of the system, without being restricted by any further numerical limits or behavioral classifications.
In other words, from the perspective of signals and systems, an excitation can take any numerical form.
As long as the unit impulse response is known, the system's response can then be calculated (predicted).

Note that \EQUA{eq_response_r3_event_conv} applies to cases where the excitation $x[h]~(h\in\mathcal{T}_h)$ and the impulse response $w[n, h]$ are both represented as temporal sequences.
For linear time-invariant systems, the convolution theorem actually provides a more convenient way to calculate the system's final response $y[n]$ in the frequency domain, by simply multiplying the excitation and the unit impulse response in the frequency domain to compute the spectral system response.
Given that computations in the frequency domain are greatly simplified, and considering the need for an accurate description of the time-varying impulse response $\mathcal{W}[k, h]$, we apply the Haar transform on the observed trajectories before further embeddings or encodings.
By leveraging the property of the minimum compact support in the time domain, Haar spectrum provides a better temporal resolution for describing time-varying features.
Furthermore, the Haar transform has a vanishing moment of 1, meaning that the high-frequency components will be mainly accounted for in the Haar spectrum, while those smooth or stationary parts will receive less focus.
In other words, the Haar transform could finely describe events in trajectory prediction scenarios that cause rapid changes on the trajectories, without the need for additional event detectors to identify changing points or ordinary temporal samples.
Therefore, the proposed reverberation transform presents a matrix multiplication form, under the usage of Haar transform.

\subsection{Implementation of Trajectory-Changing Events}

The above ideal events are employed to build the reverberation transform in a uniform and application-irrelevant way.
When constructing the trajectory prediction network \MODELSHORT, according to the linear property (\APPENDIX{A}) of the reverberation transform, we have divided the source of trajectory-changing events into two main parts, the non-interactive part and the interactive part.
When handling these events, considering the temporal-spatial precision of different events, we propose the concept of the ``minimum unit'' for representing and handling trajectory-changing events under different event-scales.
This means that when focusing on those non-interactive events, one temporal-spectral step of the trajectory representation has become the minimum unit for analyzing events.
Differently, following previous angle-based methods to represent social interactions in several angular partitions, interaction representations in one angular partition at one specific temporal-spectral step have become the minimum unit for analyzing interactive trajectory-changing events.

In our network implementations of trajectory-changing events, all computations are based on such defined minimum units.
Different from the above ideal events (where events could be any temporal sequences and last for any steps), there are three operations to make sure that real-world trajectory-changing events will be enhanced and captured from the raw observations when implementing, including the event separation, the Haar transform, and the outer product.

The event separation indicates that we have separated different sources of trajectory-changing events.
This ensures that those linear motion patterns could be filtered, leaving those changing trajectory points for further analyses.
After that, like mentioned above, the Haar transform could further concentrate on those high-frequency motion patterns upon such non-linear changing points, making sure that sudden changes in the trajectories can be captured with higher sensitivity in the Haar spectrums.
After that, the outer product (Eq. (1)) is applied on the sequential representation of those events, providing final enhancements for strengthening differences between different temporal-spectral representations (minimum units of events).
Therefore, even without using a temporal event detector, features that pass through the above three steps may to a large extent exhibit characteristics such as fast-changing and high-frequency concentration, which correspond to those trajectory-changing events occurred in real-world scenarios.

\subsection{Reverberation Strengths (Curves)}

Note that the ``reverberation curves'' (whose formal name is the ``relative non-interactive/interactive reverberation strength'') were never intended to be independent quantities outside of the reverberation transform.
Instead, they describe the core properties of the reverberation transform, or more specifically, properties of the two learnable reverberation kernels $\mathbf{R}$ and $\mathbf{G}$, since they are directly computed only from these kernels rather than manually labeled groundtruths.

Multiple trajectory-changing events could be simultaneously handled by the ego agent, even though these events might have happened at different temporal steps.
As a result, it might be insufficient to solely take into account and then track only one specific single event during the whole prediction period when forecasting agents' future behaviors.
The use of an event-wise normalization on the raw reverberation kernels to compute relative reverberation strength, \IE, the relative energy share for each specific observed event across the prediction horizon, is a direct measurement of how the different system's responses (corresponding to different events as excitations) are linearly superposed in the prediction period.
Specifically, the primary concern about these reverberation strengths (curves) is only where these values hint at the maximum or finally decrease alongside the temporal axis, not their actual values.
Our selection of using the square-sum normalization is not arbitrary.
For the reverberation kernels $\mathbf{R}$ and $\mathbf{G}$ themselves, the value of each element has been restricted between -1 and 1 ($\tanh$ activation).
Calculating the simple average would result in positive and negative terms canceling each other out, while the 1-norm (absolute value) would cause a discontinuity in the gradient around zero.
Note that although these reverberation curves do not involve the network training and optimizing, such discontinuity may lead to instability and artificial abnormalities when continuously measuring the peaks or vanishing points of these responses.
In addition, such normalization method should be parity-symmetric about zero to describe the absolute effect of the event on the response.
Therefore, the 2-norm has become the most straightforward choice for normalizing row vectors in these reverberation kernels.

From the above signals and systems' perspective, where kernel $\mathbf{R}$ could be treated as a set of impulse responses, the squares of the unit impulse response are actually the \emph{power gain} measurement of the system to respond to excitations at different temporal steps.
For the concerned ideal events or specific trajectory-changing events, these events typically manifest as power signals that persist over a period of time rather than as instantaneous energy pulses.
Given any signal $x[h]~(h \in \mathcal{T}_h)$, its average power is defined as
\begin{equation}
    P_{x} = \frac{1}{\left|\mathcal{T}_h\right|} \sum_{h \in \mathcal{T}_h} \left|x[h]\right|^2.
\end{equation}
Corresponding to \EQUA{eq_response_r3_event_conv}, for one event $x[h_0]$ happened at the $h_0$th step and its corresponding unit impulse response $w[n, h_0]$ ($n$ denotes the temporal variable), the power relation thus satisfies
\begin{equation}
    P_{y, h_0} = P_{x, h_0} \left(\sum_{n} \left|w[n, h_0]\right|^2\right).
\end{equation}
Here, the sum in the parentheses to the right side of the equation is called the total power gain.
More importantly, each squared element $\left|w[n, h_0]\right|^2$ (\IE, squares of each element in the $\mathbf{R}$ matrix) explicitly defines the \emph{instantaneous power gain}, denoted as
\begin{equation}
    g_P(n|h_0) = \left| w[n, h_0] \right|^2.
\end{equation}
This instantaneous power gain describes how much extra energy will be brought into the system specifically at the future step $n$ after the occurrence of a single event $h_0$.
Therefore, when multiple events (minimum units of trajectory-changing events) occur, the relative reverberation strength actually measures the relative power-gain share of one specific event $x[h_0]$ at a specific future step.
Take the non-interactive reverberation strength as an example, we have
\begin{equation}
    r (t|t_p) = \frac{
        (\mathbf{R}[t_p, t])^2
    }{
        \sum_{t_x=1}^{T_h} (\mathbf{R}[t_x, t])^2
    } = \frac{
        g_P(t|t_p)
    }{
        \sum_{t_x = 1}^{T_h} g_P(t|t_x)
    }.
\end{equation}
(Note that all $_\NON$ subscripts are omitted for better presentation, and such equation also applies to other altered/original non-interactive/social reverberation strengths.)
Therefore, the relative reverberation strength, or the reverberation curve in short, numerically describes the relative magnitude of the additional energy supplied to the system at each specific prediction step after the occurrence of one single event.
This is why we compute reverberation strengths in such ways.

\subsection{Summary of Interpretations}

From the above discussions, we interpret each row in the reverberation kernel $\mathbf{R}$ as the time-varying learnable unit impulse response corresponding to the input event at the same temporal step.
Especially, these ``unit impulse responses'' are learned directly from the representation of each minimum unit of input events themselves, and their unique temporal (spectral) properties can be represented from the corresponding relative reverberation strengths.
In other words, before forecasting trajectories, the \MODELSHORT~trajectory prediction model first learns to predict each row in the reverberation kernel $\mathbf{R}$ according to the representation of each input event in parallel, then forecasting trajectories based on the mutual relationships of these rows as well as their multimodal properties.
It should be noted that the neural networks only learn and understand matrices such as $\mathbf{R}$ and $\mathbf{G}$.
However, the physical meaning behind these reverberation kernels is directly reflected through the reverberation curves and the ultimate latencies defined based on these curves.
Therefore, neither the reverberation curves nor the latencies are single, manually annotated physical quantities, but rather quantitative descriptions of the properties of reverberation kernels themselves.

Also, the power-gain interpretation aligns with the definition of reverberation curves in acoustics, which describes how the loudness gain at each location changes over time under a single excitation source.
Therefore, these defined reverberation curves as well as the normalizations are not arbitrary, which is the inspiration behind this manuscript, including the reverberation transform and the \MODELSHORT~trajectory prediction model, as well as the origin of its title.

\end{document}

\documentclass[../../paper.tex]{subfiles}

\begin{document}

\begin{figure*}[tbp]
    \centering
    \includegraphics[width=1.0\linewidth]{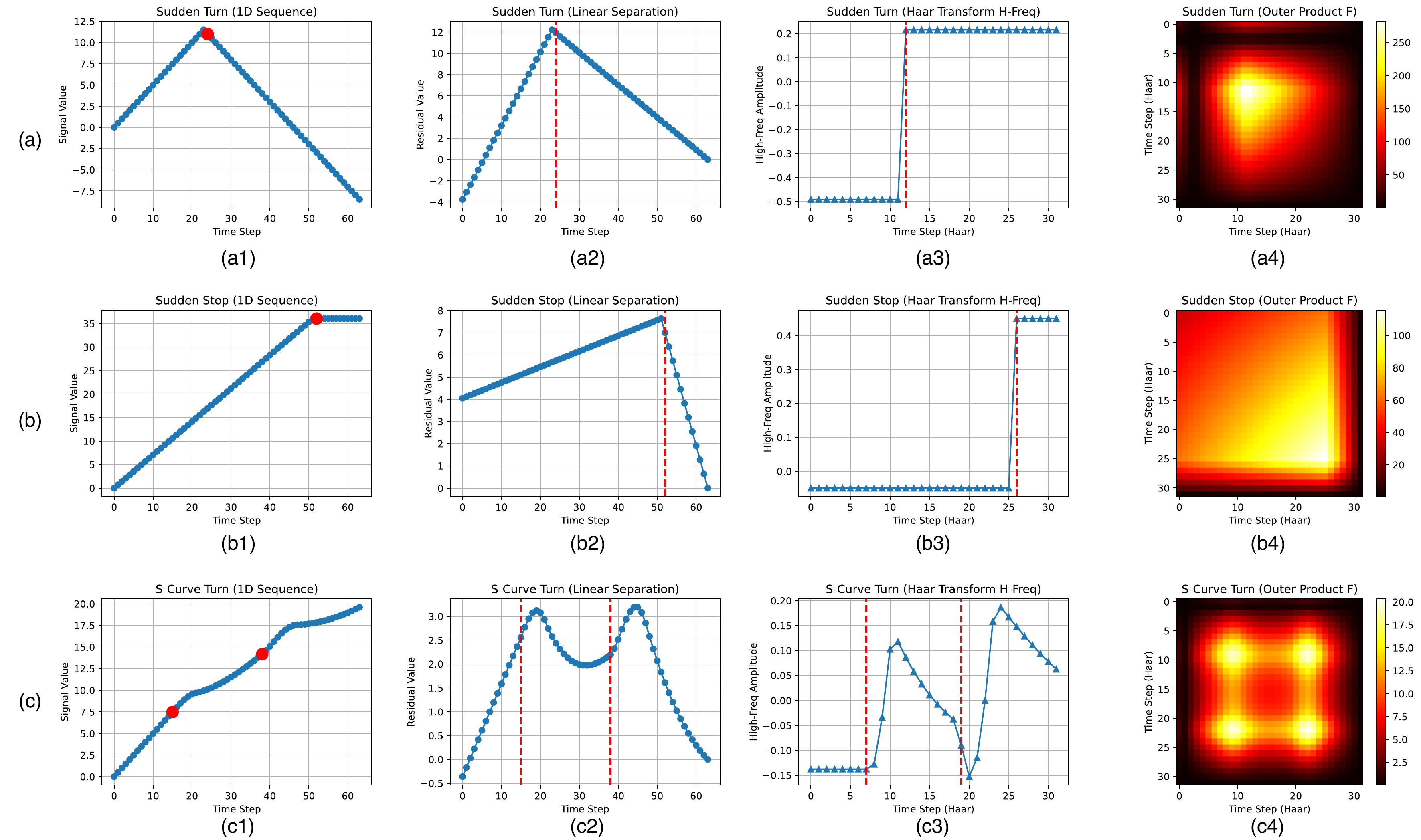}
    \caption{
        Toy example of how the proposed \MODELSHORT~model handles events without further event detectors.
        The first column (subfigures (a1), (b1), (c1)) shows several simple 1D temporal sequences with specific sequence-changing events (marked with red dots) occurring at different temporal steps.
        The second column ((a2), (b2), (c2)) shows sequences' values after applying the separation of linear components, and the third column ((a3), (b3), (c3)) shows their high-frequency components in the Haar spectrum.
        Correspondingly, the last column ((a4), (b4), (c4)) shows how these values are distributed after the outer product.
    }
    \label{fig_event_toy}
\end{figure*}

\section{Experimental Analyses of How Noticeable Events are Detected and Located}

There are no explicit event detectors used in the proposed \MODELSHORT~trajectory prediction model to ``detect'' and ``locate'' real-world noticeable trajectory-changing events from the representations of input sequences, since it applies three continuous steps to ensure the enhanced capture of those high-frequency portions or fast-changing points within those temporal representations, including the separation of linear components, the Haar transform, and the outer product.
Here, we will provide several simple toy examples of 1D sequences to directly show how these steps work to emphasize such noticeable trajectory-changing events visually.

As illustrated in \FIG{fig_event_toy}, we have created three kinds of 1D sequences to simulate changes in trajectories.
For example, case (a) describes a sequence that grows stably until the 24th temporal step, where a sudden direction change has been made.
After applying the separation of linear components and Haar transform, it can be seen in \FIG{fig_event_toy} (a3) that a distinct step has been captured in the 12th temporal-spectral step.
The application of the outer product makes this step-like change even more distinguishable, where the maximum value in the outer matrix $\mathbf{F}$ is precisely located at the (12, 12) position.
Outer matrix $\mathbf{F}$ in \FIG{fig_event_toy} (c4) presents a similar trend for enhancing two different events (starting at 15th and 38th observation step and then both last for about 5 more steps) as four noticeable regions, where two regions on the main diagonal (centered roughly at (9, 9) and (22, 22)) correspond to the steps of maximum impact of these events, while the two regions on the secondary diagonal describe the similarity between these two real-world trajectory-changing events.

In particular, please note that these toy computations are purely static, without any trainable MLPs or Transformers to additionally encode sequential features.
This means that such regions in the outer product matrices are directly computed from the raw sequences' values.
In the real \MODELSHORT~trajectory prediction model, these regions could be further enhanced by cooperating with additional trainable network components like Transformers.
As a result, even without further trainable structures, the network itself could still temporally locate and capture where these observable real-world noticeable trajectory-changing events occur during observation, even without further ``event detectors''.

\end{document}

\documentclass[../../paper.tex]{subfiles}

\begin{document}

\section{Discussions on Time-Frequency Representations and Latencies}

In the Method section, we mentioned that the Haar transform is used before encoding trajectory features for its better time-frequency joint representation capability, as well as its better characteristics for modeling high-frequency signals.
This section further verifies how these time-frequency representations and their properties help the proposed reverberation transform and the \MODELSHORT~model achieve better latency estimations and trajectory predictions.

\subsection{Haar Transform and Time-Frequency Representation}

We start by formally introducing the Haar wavelet (Haar transform).
Wavelets are proposed to solve the shortage of Fourier transform that the temporal variable $t$ is not directly or indirectly included in the Fourier spectrum $X(k)$ of a temporal sequence $x(t)$.
This limitation means that sequence $x(t)$ only has one specific Fourier spectrum $X(k)$, no matter how the time variable $t$ changes over time.
In other words, Fourier spectrum $X(k)$ has no \emph{time resolution}, so we cannot infer which frequency portions present a specific time.

This limitation corresponds directly to our discussion of latencies as well.
If we cannot directly correspond to different moments ($t$) in the Fourier spectrum, it means that if the subsequent reverberation transform is performed directly in such a spectrum, the transformed result will lose any physical meaning, turning into the massive additions of frequency portions, even though the Fourier spectrum does contain much richer spectral information about trajectories.
Many previous approaches \cite{wong2022view,wong2023another} have shown that the Fourier transform may be able to help neural networks learn and predict trajectories better, but this also directly limits the representation and study of the temporal latencies we focused on.
In other words, the core is that Fourier transform only captures global stationarity, but trajectory events are inherently non-stationary.
Therefore, finding a transform that can take into account both the temporal variable $t$ of the signal as well as analyze the spectrum of the trajectory is a direct requirement for our required representations for using the reverberation transform.

Haar transform is the simplest wavelet, making it possible to analyze the time-frequency joint representation of signals.
Formally, the continuous wavelet transform of $x(t)$ at scale $a > 0$ and translational value $b \in \mathbb{R}$ can be expressed as
\begin{equation}
    \label{eq_wavelet}
    X_w (a, b) = \frac{1}{\sqrt{a}}
        \int_{-\infty}^{\infty}
        x(t)
        \bar{\psi} \left(\frac{t-b}{a}\right) \mathrm{d}t,
\end{equation}
where $\psi (t)$ is the continuous mother wavelet, and $\bar{\psi}(t)$ represents its complex conjugate.
In the Haar wavelet, we have
\begin{equation}
    \label{eq_mother_haar}
    \psi_{\mathrm{Haar}} (t) = \left\{\begin{aligned}
        1, &\quad 0 \leq t < \frac{1}{2}, \\
        -1, &\quad \frac{1}{2} \leq t < 1, \\
        0, &\quad \mathrm{Otherwise}.
    \end{aligned}\right.
\end{equation}

With the formulas above, it can be seen that Haar spectrums are quite convenient to compute, with just simple additions or subtractions within the integrals, without needing to compute integrals over complex numbers or exponential powers, as is the case with Fourier transforms.
Most importantly, the translation coefficient $b$ is then able to directly represent the spectral response of the signal at a given moment in time, which means that we are able to consider the spectral features while focusing on how these features change over time, and are thus able to apply the reverberation transform to represent latencies for handling trajectory-changing events in the frequency domain, retaining the possibility of analyzing the spectra of the trajectories while guaranteeing the learning and prediction of our concerned latencies.

In short, the proposed reverberation transform requires that temporal information must be included in its input sequences, so that latencies could be correctly learned and forecasted as what we expected.
Further, previous works indicate that the frequency domain may help the prediction network further learn how to predict trajectories, like applying the Fourier transform on trajectories before the initial embedding steps.
However, in the Fourier transform, no time information is included in the spectrum.
As the simplest kind of wavelet, the Haar transform can construct time-frequency joint representations for trajectories by taking into account the temporal information of the sequence while including the trajectory spectrum, while providing the minimum complexity to satisfy these constraints.
Especially, the Haar wavelet possesses the minimum compact support in the time domain.
Consequently, its extremely short temporal support enables the finest time localization among wavelet bases, thereby reducing ambiguity when estimating event latencies, which constitutes the primary rationale for selecting the Haar transform over other wavelet families\footnote{
    This also explains why 1 spectral step equals 2 real-world steps/observed moments in our experimental settings, as each Haar coefficient is computed over two adjacent temporal observation steps.
}.
Accordingly, we simply adopt the Haar transform in the \MODELSHORT~model as a lightweight and latency-friendly choice.

\subsection{Further Discussion: High-Frequency Sensitivity and Trajectory-Changing Events}

Haar transform has its limitations due to the properties of its mother function $\psi (t)$, where it focuses more on the high-frequency portions of the signal.
In other words, a signal with more low-frequency portions may not be represented well by the Haar transform (In fact, this explains why \MODELSHORT~does not perform as well as other methods on nuScenes ($K=5$), see later analyses).
In wavelet transforms, the $n$th moment of a mother wavelet $\psi (t)$ is defined as
\begin{equation}
    \label{eq_vanishingmoment}
    M_n = \int t^n \psi (t) \mathrm{d} t.
\end{equation}
Simply, if there exists a positive integer $p$ such that $M_0 = M_1 = ... = M_{p-1} = 0$, then the mother wavelet $\psi (t)$ has $p$ \emph{Vanishing Moments}.
Correspondingly, for the Haar transform, it has $p = 1$ vanishing moment.
Wavelets with more vanishing moments are better at representing smooth (low-frequency) signals, because they annihilate higher-order polynomial trends.
In contrast, Haar wavelet, with only one vanishing moment, naturally emphasizes abrupt variations and high-frequency components.
It means the larger the number of vanishing moments, the better the wavelet can represent smooth signals\footnote{
    Detailed theoretical explanations can be found in the supplementary materials of the previous work \EVMODEL\EVCITE.
}.

Although this could be a strong limitation for the Haar transform to represent smoother signals in other signal processing tasks or applications, high-frequency portions in trajectories do perfectly match our focused trajectory-changing events, providing us options for specially characterizing these fast-changing parts in agents' trajectories.
In other words, those lower-frequency portions in trajectories mainly represent agents' global trends rather than specific local changings or details.
Considering that inertial motion may still be the fundamental mode that agents maintain in the absence of other external interactions or state changes, these low-frequency portions will actually be more filtered out of the Haar spectrum, at least to the extent that they do not occupy an objective share of the energy in the spectrum.
This means that the high-frequency part in trajectories, \IE, the part of the trajectory that changes rapidly alongside time, which is also the effect of self-sourced or social trajectory-changing events, will be taken into account and paid more attention to in the frequency domain.
Further, on the basis of the similarity calculation ($\sigma$) that we used before multiplying reverberation kernels $\mathbf{R}$ and $\mathbf{G}$, these high-frequency differences across different steps will be further amplified, thus allowing the reverberation transform and the whole \MODELSHORT~model to better learn the latencies that lead to these high-frequency differences.
Therefore, despite the fact that the Haar transform has a rather small vanishing moment $p = 1$, it is still a suitable trajectory transform for our needs under the demands of our reverberation transform.

Nevertheless, we have to admit that this does also lead to a decrease in the prediction accuracy of the model in some special cases.
For example, the prediction results on nuScenes when generating fewer samples ($K=5$), which can be explained by the fact that on the already smoother vehicle trajectory dataset (where obviously the low-frequency portion occupies more energy of the spectrum), the reverberation transform will likely be challenging to distinguish between different spectral steps from each other.
At the same time, the mathematical properties of the reverberation transform make the representation capability naturally limited when the $K_g$ is small (see our above analysis on the rank of $\mathbf{R}$ and $\mathbf{G}$).
As a result, in this specific case, \MODELSHORT~does not perform as well as the previous \REMODEL. 
Also, the improvement of the same results on the $K=10$ nuScenes case corroborates our thoughts.
Therefore, the proposed Haar-transform-based reverberation transform may be more suitable for applications with complex pedestrian trajectories and complex vehicle scenarios, making it possible to accurately learn and forecast the latencies of agents.

\subsection{Quantitative Verifications}


\documentclass[../../paper.tex]{subfiles}

\begin{document}

\begin{table*}[tbp]
    \caption{
        Ablation studies of trajectory transforms in the proposed \MODELSHORT~model.
        Results are reported as ``minADE$_{20}$/minFDE$_{20}$''.
    }
    \label{tab_haar_ablation}
    \centering
    \begin{tabular}{cccccccccccccc}
        \toprule
        ID & Transform & eth & hotel & univ & zara1 & zara2 & ETH-UCY \\

        \midrule
        t1 & None   & 0.230/0.348       & 0.109/0.165       & \X{0.250}/0.430   & \X{0.177/0.298}   & \X{0.133}/0.225   & \X{0.180}/0.293 \\
        t2 & DFT    & \X{0.234/0.362}   & \X{0.112/0.167}   & 0.245/0.431       & 0.174/0.291       & \X{0.133/0.228}   & \X{0.180/0.296} \\
        t3 & DB2    & \C{0.226}/0.350   & \C{0.105}/0.155   & 0.248/\X{0.433}   & 0.175/0.296       & \C{0.131}/0.223   & 0.177/0.291 \\
        
        \midrule
        full & Haar & 0.229/\C{0.347}   & \C{0.105/0.154}   & \C{0.237/0.409}   & \C{0.172/0.283}   & \C{0.131/0.220}   & \C{0.174/0.282} \\

        \bottomrule
    \end{tabular}
\end{table*}

\end{document}

We have conducted several ablation variations to verify how the Haar transform helps in the \MODELSHORT~model quantitatively.
We choose three additional kinds of transforms, including the None transform (perform reverberation transform on temporal features directly), discrete Fourier transform (DFT), and the Daubechies-2 transform (DB2, with the vanishing moment of $p = 2$), comparing to the Haar transform (vanishing moment $p = 1$) used in the full \MODELSHORT~model.
Experimental settings are the same as experiments on ETH-UCY in the main manuscript.

As reported in \TABLE{tab_haar_ablation}, we see that Haar helps the \MODELSHORT~model achieve the best performance on most ETH-UCY sets.
Overall, Haar variation outperforms other transforms' variations by at least 2.8\% in ADE and 3.1\% in FDE, which is quite noteworthy for the ETH-UCY.
However, it is unexpected that the DFT variation t2 performs the worst, even worse than the non-transform variation t1 on most sets.
This further verifies our thought that the DFT-based reverberation transform has lost its physical meaning of structurally representing latencies for handling trajectory-changing events, since the temporal variable $t$ is not directly or indirectly included in Fourier spectrums.
Under such a situation, spectral representations of all trajectory-changing events would be randomly superposed in the single frequency domain, making the \MODELSHORT~prediction model even more confused, further leading to the performance drops.
Note that our thoughts can only be roughly hypothesized from quantitative data alone.
We further confirm our thoughts about the Fourier transform later by visualizing the reverberation curves.
Please refer to the next section.

We also introduce the DB2 wavelet to verify our thoughts about the vanishing moment and trajectory-changing events.
Mathematically, DB2 enhances Haar by further considering signals' lower-frequency portions in the spectrums.
However, from the results reported in \TABLE{tab_haar_ablation}, it can be seen that the DB2 variation t3 performs not as well as the full Haar variation on most sets.
We infer that this is due to the trajectory-changing events mostly correspond to higher-frequency portions of the trajectory-spectrums, which means that DB2 could not capture such high-frequency differences as Haar.
Interestingly, DB2 variation does bring better prediction performance on several sets, like eth and hotel, further providing insights for selecting transforms for different prediction scenes initially.

\subsection{Qualitative Discussions}


\documentclass[../../paper.tex]{subfiles}

\begin{document}

\begin{figure*}[tbp]
    \centering
    \includegraphics[width=1.0\linewidth]{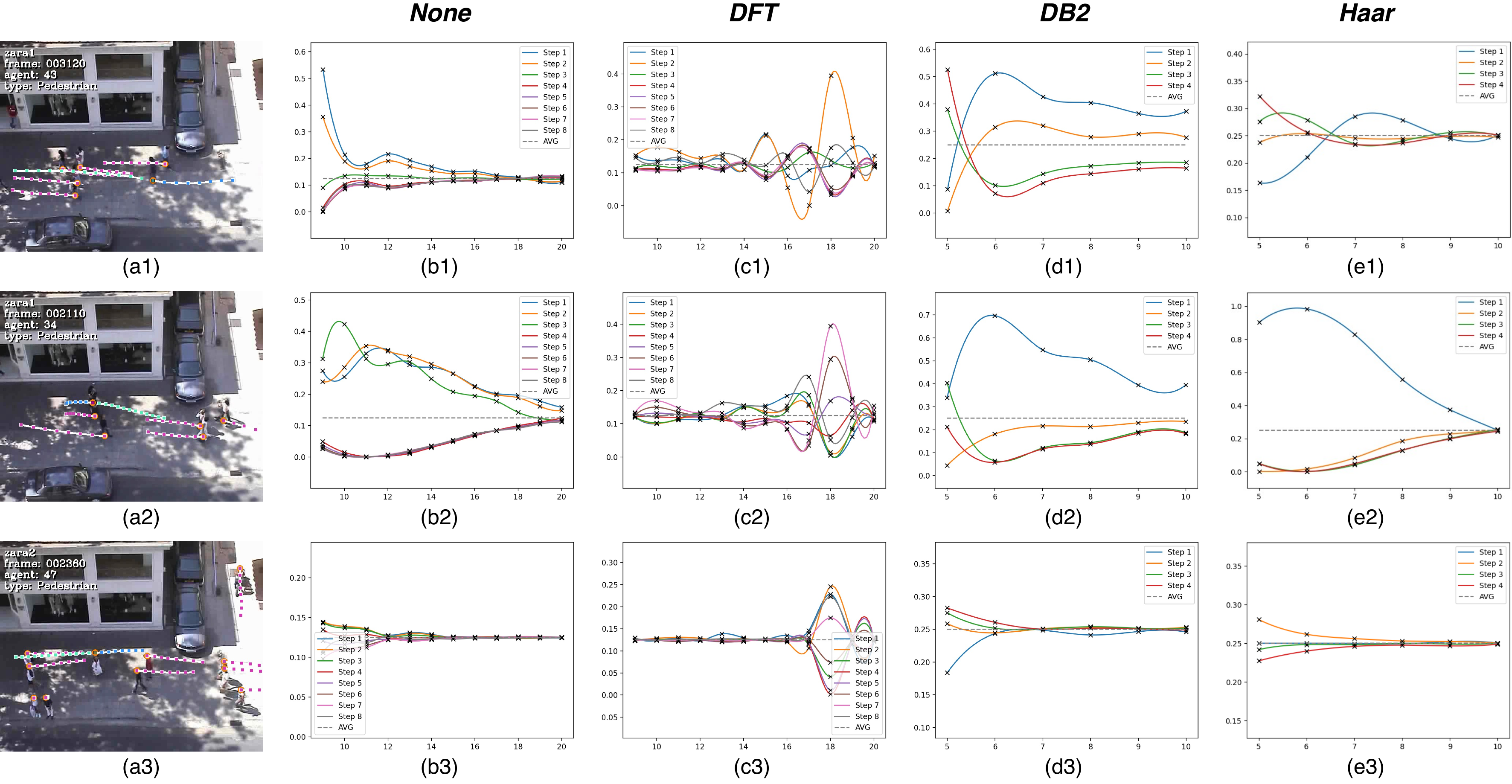}
    \caption{
        Examples of non-interactive reverberation strength $r_\NON$ of kernel $\mathbf{R}_\NON$ in the full \MODELSHORT~model, under the use of different kinds of trajectory transforms: None (no trajectory transforms, b1 to b3), DFT (c1 to c3), DB2 (d1 to d3), and Haar (e1 to e3).
    }
    \label{fig_curve_R_non_transform}
\end{figure*}

\end{document}

We now visualize reverberation curves to further discuss how different trajectory transforms affect the learning of reverberation kernels $\mathbf{R}$ and $\mathbf{G}$.
To make it easy to compare, we only visualize the original non-interactive reverberation strengths $r_\NON (t|t_p)$ computed from the non-interactive reverberation kernel $\mathbf{R}_\NON$ from the full \MODELSHORT~model, with the use of None transform, DFT, DB2, and Haar transform on trajectories when forecasting under the $\{t_h, t_f\} = \{8, 12\}$ setting.

\NEWHALFLINE
\subsubsection{Wavelets with Different Vanishing Moments}
By comparing different columns in \FIG{fig_curve_R_non_transform}, it can be seen that wavelet variations (DB2 and Haar) predict similar reverberation curves for these three scenes (a frequency step of these wavelets both equal to 2 adjacent temporal steps).
For example, comparing \FIG{fig_curve_R_non_transform} (d1) and (e1), we observe that events on historical steps 1 and 2 have been both predicted delayed until the future step 6 (DB2) or step 7 (Haar), though their reverberation strengths are differently estimated.
On the contrary, events on steps 3 and 4 have been both considered finished after the step 6, where both these steps share similar reverberation strength.
Interestingly, this slight latency difference can be explained by the different properties of Haar and DB2.
Haar's vanishing moment is 1 and DB2's is 2. 
This means that DB2 is able to focus more on the low frequency part of the signal compared to Haar.
In other words, DB2's description of the high-frequency part, which mostly corresponds to trajectory-changing events, occupies less energy in the spectrum.
For trajectories, this means that spectral differences on different historical steps will be more difficult to capture, which may lead to the model not being able to distinguish well between trajectory-changing events on consecutive steps from the feature level. 
Thus, in \FIG{fig_curve_R_non_transform} (d1), steps 1 and 2 have been predicted almost the same tendency of reverberation curves, while in (e1) it is well distinguished as the actual scene where the agent accomplishes the turn in step 2 (frames 3-4 in (a1)).

Nevertheless, it is worth noting that such differences tend to be relatively weaker as well.
Other curves of the models using Haar and DB2 have almost identical latency trends in the scenes in (a2) and (a3).
Further, combined with our quantitative analyses, this difference also suggests that transforms with smaller vanishing moments (Haar) are suitable for trajectory-changing events that occur and have an impact more quickly, whereas larger ones (like DB2) are suitable for those that are slow to have an impact, which provides theoretical support for choosing different trajectory transforms for different scenarios.

\NEWHALFLINE
\subsubsection{Fourier Transform}
Although there are minor differences between the reverberation curves of non-transform variations and these wavelet variations, these curves still present similar latency trends.
However, curves learned and computed from the Fourier spectrums (\FIG{fig_curve_R_non_transform} (c1) to (c3)) are totally non-interpretable, especially from the latencies' point of view.
Since the temporal variable $t$ is not included in the Fourier spectrum (Fourier spectrum represents all temporal events globally within the only spectrum), all observed trajectory-changing events would be overlappingly distributed throughout the entire spectrum without any distinguishable flags or features.
Under this circumstance, the proposed reverberation transform no longer learns and predicts how future steps are combined with historical events with specific latencies, but how the future spectrums would be superposed spectrally.
In other words, curves in \FIG{fig_curve_R_non_transform} (c1) to (c3) are actually frequency mapping curves that illustrate how the spectrum of a trajectory will be \emph{globally} shifted over the observation and prediction periods in the frequency domain.

While the Fourier-transform-based reverberation transform could still be accounted for through the above spectral mapping perspective, it does not address our core concern, \IE, the latencies of agents when responding to trajectory-changing events.
This is precisely because different events are not effectively localized in such spectrums.
Although the Fourier spectrum is more spectrally expressive than the Haar transform, this problem prevents it from paying additional attention to the time-varying properties of the spectrums. 
As analyzed in our quantitative results in the \TABLE{tab_haar_ablation}, especially the worse FDEs on all ETH-UCY sets, this spectral mapping in the Fourier spectrum buries all the trajectory-changing events, which leads to faster performance degradation in more complex or prediction scenarios with more agents involved.
Since in these scenarios, the spectral representations of all interactive or non-interactive events are superposed in the same frequency domain, thus completely obfuscating these events, not to mention the underlying latencies. 
This further proves why the Fourier transform should not be used to help the proposed reverberation transform learn latencies, though it has better spectral representation capability.

\end{document}

\documentclass[../../paper.tex]{subfiles}

\begin{document}

\section{Discussions of Latencies, Dependencies, Temporal Attention, and Temporal Convolution}

Learning and forecasting the temporal latencies of agents for handling trajectory-changing events is the main concern of this manuscript.
Note that latency itself is not an artificial enhancement mechanism introduced by the proposed reverberation transform.
Instead, it describes the temporal properties of how each agent responds to different trajectory-changing events and finally modifies their trajectories, or how the system responds to the input excitations and finally modifies its outputs from the system-theory's view.
On the contrary, in certain network structures where the input event is assumed to propagate instantly without further temporal offsets, such properties may tend to be overshadowed or even neglected.
This section provides clear formula descriptions of several relevant conceptions, explaining why the (temporal) latencies are significant to this task, and why current temporal dependencies, temporal attention, and temporal convolution approaches do not explicitly model latencies.

\subsection{Temporal Dependencies and Latencies}

Temporal dependency characterizes how strongly one moment (either an observation or a prediction moment) contributes to another.
Formally, most dependency-based models aim to estimate a mapping of the form
\begin{equation}
    \mathcal{D}(t_1, t_2) = g \left(
        \mathbf{f}(t_1), \mathbf{f}(t_2)
    \right),
\end{equation}
where $g$ measures temporal correlation between representations (or embeddings) $\mathbf{f}(t_1)$ and $\mathbf{f}(t_2)$ of two considered temporal steps $t_1$ and $t_2$.

In recurrent-networks-based methods, these dependencies are mostly considered as the relationship of two adjacent temporal steps, like a $t_1 = t$ and the other $t_2 = t + 1$, thus learning how to build recurrent cells within RNNs or LSTMs to infer features of the next step based on the current status (or saved memories if available).
For models using the Transformer, such dependencies have become less recurrent and more relative to the global.
These methods extract features over the observation time period (temporal attention is not discussed here, see the next section) and holistically evaluate how these features contribute to the prediction time period.
Thus, temporal dependencies focus more on how different moments affect or contribute to each other.

Differently, temporal latency concerns more with when an event at time $t_0$ becomes behaviorally effective, or starts to change the outcome of the system.
It describes a temporal shift
\begin{equation}
    \label{eq_latency_shift}
    \mathcal{D}: t_0 \to t_0 + \Delta t_0
\end{equation}
indicating the onset of its impact, and further the offset, \IE, the whole life-cycle of this event.
Thus, latencies represent the concept of a time latency rather than a simple intensity contribution, with a focus on considering how the impact of an event changes over the timeline.
In short, dependencies operate in the magnitude domain, whereas latencies operate in the time-index domain.
In human-agent reasoning, this corresponds to the distinction between what matters (dependencies) and when it starts to matter (latencies).
This is the main difference between temporal dependencies and latencies.

\subsection{Why Temporal Attention Does Not Explicitly Encode Latencies}

The computation as well as the physical meanings could be different if the attention-like feature mixing or weighting were applied.
Under the normal temporal self-attention\footnote{
    We do not discuss cross-attention here since there are no real future trajectories available during the model inference.
    This maintains conceptual consistency with our reverberation transform, which directly predicts trajectories rather than acting as a post-refinement process after an initial prediction.
} setting \cite{vaswani2017attention}, a set of attention weights $\{w(i, j)\}$ (where $i, j \in \{1, 2, ..., T_h\}$, and $\sum_{j} w(i, j) = 1$) will be computed after querying for weighted-summing features on all observed temporal steps, therefore paying more focus (the ``attention'') on those attentive steps for feature fusion or further decisions.
Formally, given a set of temporal representations $\{e(1), e(2), ..., e(T_h)\}$, attention computation aims at constructing a mixed feature for the target $i$th historical step:
\begin{equation}
    \label{eq_appendix_attention}
    Y(i) = \sum_{j = 1}^{T_h} e(j) w(i, j).
\end{equation}

For the reverberation transform, denote the $h$th row vector in the reverberation kernel $\mathbf{R}$ as $\mathbf{w}(f|h)$ and the representation of all inputs occurring at the $h$th observation step as $\mathbf{x}(h)$, the process to predict the final representation at the $f$th future step (denoted as $\mathbf{y}(f)$) after the multiplication of the reverberation kernel $\mathbf{R}$ can be formulated as
\begin{equation}
    \label{eq_appendix_R_matrix}
    \mathbf{y}(f) = \sum_{h \in \mathcal{T}_h} \mathbf{x}(h) \mathbf{w}(f|h), \quad\forall f \in \mathcal{T}_f.
\end{equation}

\subsubsection{Differences between the Reverberation Transform and Attention Mechanism}

Comparing \EQUA{eq_appendix_R_matrix} and \EQUA{eq_appendix_attention}, two main differences can be seen between the reverberation kernel and the attention weights.
We summarize these differences using terms \emph{reversibility} and \emph{temporality}.

\NEWHALFLINE
(i)~\textbf{Reversibility.}
As attention weights are normalized ($\sum_{j} w(i, j) = 1$), attention focuses more on relative relations or relative contributions of different temporal steps (like between $e(1)$ and $e(2)$) for fusing and constructing some specific new global temporal representations.
During this process, the original information across different temporal steps might be compressed because of the $\mathbf{W}$ matrix (composed of the above $w(i, j)$) that applies softmax and row-wise normalization.
Thus, the attention matrix itself is more likely to be (or encouraged to be under specific loss functions) sparse, and its rows tend to be correlated with one another, which means that the rank of this weight matrix will be probably low, and such attention decomposition might be therefore non-unique.
Also, due to the information compression during the normalization when computing attention weights, it could be hard to find an inverse form that could reconstruct the original temporal features from the mixed one, since the inverse of matrix $\mathbf{W}$ might not exist because its rows could be correlated.

In contrast, the reverberation transform does not employ row or column normalizations.
Also, since each row of the $\mathbf{R}$ matrix is computed from each $e(h)$ (or $\mathbf{x}(h)$) in parallel, and due to the use of Haar transform and outer product, each row in the $\mathbf{R}$ matrix is encouraged to exhibit unique high-frequency feature patterns, \IE, each row is more likely to be independent of others than the above attention-like multiplication.
Especially, the reverberation transform can have an inverse form, only when matrices $\mathbf{R}$ and $\mathbf{G}$ are both row full-rank:
\begin{gather}
    \mathbf{F} = 
        \mathcal{R}^{-1}_{(\mathbf{R}, \mathbf{G})} \left[~
            \overline{\mathbf{F}}
        ~\right] =
        \overline{\mathbf{G}}^\top
        \overline{\mathbf{F}}~
        \overline{\mathbf{R}}, \\
    \mbox{where}\quad
        \overline{\mathbf{R}} = \mathbf{R}^\top (\mathbf{R}\mathbf{R}^\top)^{-1},
        \quad
        \overline{\mathbf{G}} = \mathbf{G}^\top (\mathbf{G}\mathbf{G}^\top)^{-1}.
\end{gather}
This means that the reverberation transform as well as its inverse form describe how energy will be transferred between two domains (observation and prediction), rather than how information is unidirectionally compressed and mixed.

\NEWHALFLINE
(ii)~\textbf{Temporality.}
There are no explicit or structural temporal dependencies, temporal orders, or temporal causal relations between the attention output $Y(i)$ and the original $e(i)$ or the attention weights $w(i, j)$.
When computing the attention matrix, the entire time series $\{e(1), e(2), ..., e(T_h)\}$ is all known, and the computation of each $W_{i:} \in \mathbb{R}^{T_h}$ row in the attention matrix depends strictly on querying the relative relationships between the concerned $e(i)$ and all other temporal features $e(k),~k \in \{1, 2, ..., T_h\}$.
In other words, all temporal steps are treated equally and simultaneously when being assigned attention scores from the dataflow level.
When computing attention weights, the query is designed to compute the similarity or further measurements between the features at all time steps and the currently focused one.
This means that the temporal relations have been limited to the observation period.
Although we could employ cross-attention approaches, they may require prior knowledge of future trajectories, which means that the use of cross-attention of future trajectories becomes a post-processing refinement rather than the direct prediction process.
Consequently, it cannot be reliably compared to the reverberation transform, where future representations are computed using a single forward pass.
In other words, attention queries and mixes features of one domain according to one another, while the reverberation transform learns how to map features from one to another.

For the reverberation transform, each row in the reverberation kernel $\mathbf{R}$, \IE, the symbol $\mathbf{w}(f|h)$ in \EQUA{eq_appendix_R_matrix}, corresponds directly to a specific and unique temporal feature mapping $w: e(h) \mapsto Y(f)$.
Unlike the positional encoding in Transformer attention, this temporal mapping relationship is constrained by the structure of the reverberation transform.
After feature embedding and encoding of the observed information through backbone networks, the reverberation transform uses a bilinear multiplication to directly map features from the observed time steps to the predicted ones.
Specifically, for each feature dimension of the output matrix ($\overline{\mathbf{F}}_{::d}\in\mathbb{R}^{K_g \times T_f}$), $T_f$ elements in each row directly correspond in order to the feature representations of the $f$th ($f \in \{1, 2, ..., T_f\}$) prediction step, which is directly operated by the kernel $\mathbf{R} \in \mathbb{R}^{T_h \times T_f}$.
This also means that kernel $\mathbf{R}$ is asymmetric across both two temporal dimensions ($T_h$ and $T_f$).
Exchanging any rows or columns will directly result in the swapping of impact responses to different events, or the disruption of predicted trajectories at different future steps.
From this point of view, attention computations are not that sensitive to temporal orders structurally.
Except for the additional ``soft'' positional embeddings, the attention weights are calculated with the permutation invariance property so that features at every temporal step are treated and measured equally.

\NEWHALFLINE
\subsubsection{Experimental Validations}

In the above discussions, we have summarized two major differences between the proposed reverberation transform and current attention approaches as \emph{reversibility} and \emph{temporality}.
We first validate the reversibility property, verifying whether the reverberation transform is reversible in the experiments.

\NEWHALFLINE
(i)~\textbf{Validations of the Rank Properties of Reverberation Kernels \{$\mathbf{R}, \mathbf{G}$\} and Attention Matrix $\mathbf{W}$.}
We have randomly selected 10,000 trajectory prediction samples in the ETH-UCY dataset for testing, under the standard $(t_h, t_f, \Delta t) = (8, 12, 0.4)$ setting.
For the \MODELSHORT~trajectory prediction model, there will be two different reverberation transforms used in two different network branches, including the non-interactive reverberation transform and the social reverberation transform.
When computing self-attention matrix $\mathbf{W}^i$ for each agent, we use the standard self-attention pipeline:
\begin{equation}
    \mathbf{W}^i = \text{softmax} \left( \frac{\mathbf{f}^i (\mathbf{f}^i)^\top}{\sqrt{d}} \right),
\end{equation}
where $\mathbf{f}^i$ (can be $\mathbf{f}^i_\NON$ or $\mathbf{f}^i_\SOC$) represents the sequential features of the ego agent itself or further interaction representations, and $d$ is the feature dimension.
Under the same prediction setting, the shapes of the corresponding reverberation kernels and attention weights are listed in \TABLE{tab_response_r3_overall_tensor_shapes}.

\begin{table*}[tbph]
    \caption{
        Tensor Shapes of Used Reverberation Kernels and Self-Attention Weights
    }
    \label{tab_response_r3_overall_tensor_shapes}
    \centering
    \begin{tabular}{cccccccc}
        \toprule
        Symbol & Name & Shape & Tensor Shape \\

        \midrule

        $\mathbf{R}^i_\NON$ & Non-interactive reverberation kernel & $T_h \times T_f$ & (4, 6) \\
        $\mathbf{G}^i_\NON$ & Non-interactive generating kernel & $T_h \times K_g$ & (4, 20) \\
        $\mathbf{R}^i_\SOC$ & Social reverberation kernel & $T_hN_\theta \times T_f$ & (32, 6) \\
        $\mathbf{G}^i_\SOC$ & Social generating kernel & $T_hN_\theta \times K_g$ & (32, 20) \\
        
        \midrule
        
        $\mathbf{W}^i_\NON$ & Non-interactive self-attention weights & $T_h \times T_h$ & (4, 4) \\
        $\mathbf{W}^i_\SOC$ & Social self-attention weights & $T_hN_\theta \times T_hN_\theta$ & (32, 32) \\
        
        \bottomrule
    \end{tabular}
\end{table*}

We have calculated the rank of all these above matrices on 10,000 randomly selected samples.
Results are reported in the following \TABLE{tab_response_r3_overall_rank}.
As we analyzed before, the inverse form of the reverberation transform is strictly valid only under the condition that matrices $\mathbf{R}$ and $\mathbf{G}$ are of full row rank.
Hence, it is clear that the reversibility of the reverberation transform only holds when the number of its input steps $T_h$ is no greater than its output steps $T_f$ and the number of multiple generations $K_g$.
Otherwise, the inverses of both matrices $(\mathbf{R}\mathbf{R}^\top)^{-1}$ and $(\mathbf{G}\mathbf{G}^\top)^{-1}$ would not exist.
This means that only kernels $\mathbf{R}_\NON$ and $\mathbf{G}_\NON$ meet the above requirements in the proposed \MODELSHORT~model.
When reverberation kernels are both \textbf{row} full-rank (as the reported ranks), the inverse form of the reverberation transform can hold, making it possible to map features from one domain to another without further information loss.

Regarding the two kernels in the social branch, \IE, the reverse form of the proposed transform is not defined in the social branch of the proposed \MODELSHORT~model, here we provide further discussions and interpretations.
For the non-interactive branch, \MODELSHORT's use of the reverberation transform is as expected.
It maps features from the observation period $T_h$ to the prediction period $T_f$.
However, for the social branch, the mapping we used, \IE, the ``$T_h N_\theta \to T_f$'', does not actually conform to the intended usage of the reverberation transform itself.
Instead, such ``$T_h N_\theta \to T_f$'' mapping is designed to simultaneously gather social context (spatially on all $N_\theta$ social partitions) as well as map features when making social-aware predictions.
Therefore, when discussing the reversibility of the reverberation transform itself, we should not directly use \MODELSHORT~here, particularly in the social branch for validation due to its bundled pipeline with deterministic information fusion and compression on those social contexts.

Here, we introduce a supplementary experiment to further verify this reversibility property of the reverberation transform on handling such $T_h N_\theta$-step social sequences by conducting an additional ablation experiment.
Such validation does not affect our structural design and conclusions of the proposed \MODELSHORT~trajectory prediction model.
Instead, it is an experimental attempt to further explore the expression boundary of the proposed reverberation transform.
To achieve this goal, the new ablation variation (named $v_\SOC$) modifies the reverberation transform part of the original social branch, making it available to map features simultaneously and directly in all social branches across all input-output temporal steps.
In other words, the former ``$T_h N_\theta \to T_f$'' mapping has become ``$T_h N_\theta \to T_f N_\theta$'' to better validate its social representation capability.
Correspondingly, we have
\begin{equation}
    \overline{\mathbf{F}^i_\SOC} \in \mathbb{R}^{K_g \times (T_f N_\theta) \times d}
\end{equation}
in this new $v_\SOC$ variation.
For better presentation, we use symbol $\mathbf{V}^i_\SOC$ to denote its ``unflattened'' form:
\begin{equation}
    \mathbf{V}^i_\SOC = \mathrm{Reshape}\left(
        \overline{\mathbf{F}^i_\SOC}
    \right)
    \in \mathbb{R}^{
        K_g \times T_f \times N_\theta \times d
    }
\end{equation}
Such representation will then be simply max-pooled to gather information from all social branches (since such pooling methods have been widely validated in previous methods).
Formally, its final predictions are obtained by
\begin{equation}
    \Delta \hatbf{Y}^i_\SOC = \mathcal{T}^{-1} \left[
        D_\SOC \left({\mathbf{V}^i_\SOC}'\right)
    \right] \in \mathbb{R}^{K_g \times t_f \times m},
\end{equation}
where
\begin{equation}
    {\mathbf{V}^i_\SOC}' = \max_{\theta = 1}^{N_\theta} \left(\mathbf{V}^i_\SOC\right)_{::\theta:} \in \mathbb{R}^{K_g \times T_f \times d}.
\end{equation}

We have conducted ablation trainings\footnote{
    The $K_g = 20$ $v_\SOC$ variation performs similar to the original \MODELSHORT~model in the main manuscript, with about $\pm 1\%$ similar ADE and FDE on ETH-UCY datasets, therefore we omit these quantitative results and comparisons here.
} under the same settings as our rank validations to report ranks of these ``step-equivalent'' social branches.
In addition, the current $K_g = 20$ setting comes from the \emph{best-of-20} validation of the common trajectory prediction setting.
To make generating kernel $\mathbf{G}_\SOC$ also available to be full row rank, we have also prepared an additional $K_g = 40$ $v_\SOC$ variation to further test the representation capability of the reverberation transform.
These results are reported in \TABLE{tab_response_r3_overall_rank}.

It can be seen that the social reverberation kernel $\mathbf{R}_\SOC$ in the new $v_\SOC~(K_g = 20)$ variation could still almost keep the full row rank ($T_h N_\theta = 4 \times 8 = 32$) properties, with only 7 or 8 samples having the rank of 31 over all the 10,000 sampled prediction cases.
Notably, when increasing $K_g$ to 40, both kernels $\mathbf{R}_\SOC$ and $\mathbf{G}_\SOC$ achieve full row rank on all of these samples.
This provides us an interesting interpretation that both kernels could collaborate during the training process to shape the final representation of the reverberation transform, where the limited $K_g = 20$ generating kernel may lead to the limited representation capability of its corresponding reverberation kernel (which finally leads to several 31-rank $\mathbf{R}_\SOC$ kernels).

Nevertheless, the condition where the reversibility of the reverberation transform can hold is still that matrices $\mathbf{R}$ and $\mathbf{G}$ are both full row rank.
As a conclusion, in the proposed \MODELSHORT~model, all kernels in the non-interactive branch could meet such requirements, and therefore the corresponding non-interactive reverberation transform is reversible.
On the contrary, due to the different matrix shapes of kernels in the social branch, its corresponding social reverberation transform is not reversible.
Additional experimental discussions of the ``step-equivalent'' model variation further show that the reverberation transform itself could still handle such mixed and relatively longer social representation sequences (``$T_h N_\theta \to T_f N_\theta$''), indicating its noticeable representation capability.
Also, such reversibility is not the core design goal or strict mathematical condition of building the reverberation transform.
Instead, this is an additional property that can be experimentally verified, which is different from the information compression and the irreversibility of attention weight matrices, indicating the unique mathematical properties as well as applications of current attention mechanisms and the proposed reverberation transform.
 
Differently, although the non-interactive attention weights $\mathbf{W}^i_\NON$ are full-rank on most samples (with only about 0.56\% ones has the rank of 3), the social attention weights $\mathbf{W}^i_\SOC$ have never reached full-rank.
This means that attention can maintain a certain degree of reversibility for relatively short sequences.
However, as the sequence length increases, it becomes difficult to preserve the rank of the attention weights, meaning that the decomposition is no longer unique in this case, and the original input cannot be reconstructed from the attention results.

\begin{table*}[tbph]
    \caption{
        Ranks of Different Matrices over 10,000 Trajectory Prediction Samples
    }
    \label{tab_response_r3_overall_rank}
    \centering
    \begin{threeparttable}
        \begin{tabular}{cccccccc}
            \toprule
            Symbol & Tensor Shape & Average Rank & Min Rank & Max Rank & Full-Rank Rate \\

            \midrule
            $\mathbf{R}^i_\NON$ & (4, 6)    & 4.00  & 4     & 4     & 100.00\% (full row rank)\\
            $\mathbf{G}^i_\NON$ & (4, 20)   & 4.00  & 4     & 4     & 100.00\% (full row rank)\\
            $\mathbf{R}^i_\SOC$ & (32, 6)   & 6.00  & 6     & 6     & 100.00\% (full column rank)\\
            $\mathbf{G}^i_\SOC$ & (32, 20)  & 20.00 & 20    & 20    & 100.00\% (full column rank)\\

            \midrule
            $\mathbf{R}^i_\SOC~(v_\SOC, K_g = 20)$ & (32, 48)   & 31.99  & 31     & 32     & 99.92\%\\
            $\mathbf{G}^i_\SOC~(v_\SOC, K_g = 20)$ & (32, 20)  & 20.00 & 20    & 20    & 100.00\% (full column rank)\\
            $\mathbf{R}^i_\SOC~(v_\SOC, K_g = 40)$ & (32, 48) & 32.00 & 32 & 32 & 100.00\% (full row rank)\\
            $\mathbf{G}^i_\SOC~(v_\SOC, K_g = 40)$ & (32, 40) & 32.00 & 32 & 32 & 100.00\% (full row rank)\\

            \midrule
            $\mathbf{W}^i_\NON$ & (4, 4)    & 3.99  & 3     & 4     & 99.44\%\\
            $\mathbf{W}^i_\SOC$ & (32, 32)  & 17.61 & 13    & 26    & 0.00\%\\
            
            \bottomrule
        \end{tabular}
        \begin{tablenotes}[flushleft]
            \footnotesize
            \item[]
                Results are obtained by testing on 10,000 randomly selected ETH-UCY real-world trajectory data under the longer prediction $(t_h, t_f, \Delta t) = (8, 12, 0.4)$ setting.
                For the $v_\SOC$ variation, the reverberation transform used in the social branch has been modified to map features from $T_h N_\theta$ input steps into $T_f N_\theta$ output steps.
        \end{tablenotes}
    \end{threeparttable}
\end{table*}

\begin{table*}[tbph]
    \caption{
        Ranks of Different Matrices over 2,000 Longer Trajectory Prediction Samples
    }
    \label{tab_response_r3_overall_rank_longer}
    \centering
    \begin{threeparttable}
        \begin{tabular}{cccccc}
            \toprule
            Symbol & Tensor Shape & Average Rank & Min Rank & Max Rank & Full-Rank Rate \\
            \midrule
            $\mathbf{R}^i_\NON$ & (8, 12)   & 8.00  & 8     & 8     & 100.00\% (full row rank)\\
            $\mathbf{G}^i_\NON$ & (8, 20)   & 8.00  & 8     & 8     & 100.00\% (full row rank)\\
            $\mathbf{R}^i_\SOC$ & (64, 12)  & 12.00 & 12    & 12    & 100.00\% (full column rank)\\
            $\mathbf{G}^i_\SOC$ & (64, 20)  & 20.00 & 20    & 20    & 100.00\% (full column rank)\\
            \midrule
            $\mathbf{W}^i_\NON$ & (8, 8)    & 7.74  & 5     & 8     & 80.05\%\\
            $\mathbf{W}^i_\SOC$ & (64, 64)  & 25.47 & 22    & 33    & 0.00\%\\
            
            \bottomrule
        \end{tabular}
        \begin{tablenotes}[flushleft]
            \footnotesize
            \item[]
                Results are obtained by testing on 2,000 randomly selected ETH-UCY real-world trajectory data under the longer prediction $(t_h, t_f, \Delta t) = (16, 24, 0.4)$ setting.
        \end{tablenotes}
    \end{threeparttable}
\end{table*}

We have also verified the rank properties under the longer prediction setting ($(t_h, t_f, \Delta t) = (16, 24, 0.4)$) in \TABLE{tab_response_r3_overall_rank_longer}.
It can be observed that as the sequence length increases further, it becomes more difficult for the attention matrix to maintain full rank.
Nevertheless, even under such a long prediction period, the non-interactive reverberation kernels are still of full row rank.
This also implies that even in such cases, the original observed temporal features in this branch can still be recovered by applying the inverse reverberation transform.

In other words, when the full row rank condition is satisfied, such decomposition of the bilinear matrix multiplication of the proposed reverberation transform is unique for these real-world trajectory samples, while attention may not be for most of them.
Therefore, this serves as important evidence distinguishing the proposed reverberation transform from temporal attention, where the reverberation transform focuses on how energy will be transferred, while attention focuses on how information should be redistributed.

\begin{figure*}[tbp]
    \centering
    \includegraphics[width=1.0\linewidth]{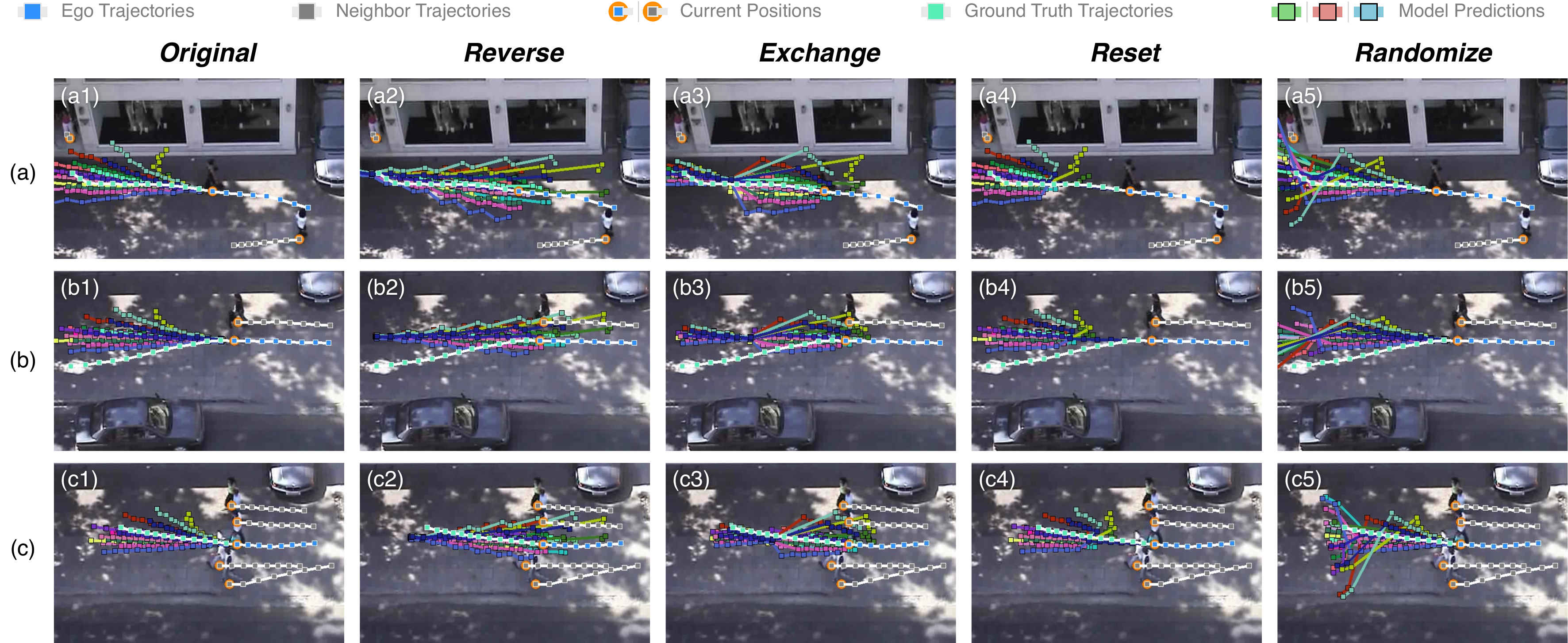}
    \caption{
        Counterfactual validations of temporal structures in the reverberation kernel $\mathbf{R}$, under four kinds of interventions: reverse rows, exchange rows, reset to zero, and randomize specific rows.
    }
    \label{fig_intervene_R}
\end{figure*}

\NEWHALFLINE
(ii)~\textbf{Counterfactual Validations of Temporal Structures in $\mathbf{R}$.}
The other major difference between the proposed reverberation transform and temporal attention is that the temporal correspondences are structurally embedded in kernels $\mathbf{R}$ and $\mathbf{G}$, \IE, the \emph{temporality}.
We now conduct several simple interventions on kernel $\mathbf{R}$ (including both the non-interactive $\mathbf{R}^i_\NON \in \mathbb{R}^{T_h \times T_f}$ and the social $\mathbf{R}^i_\SOC \in \mathbb{R}^{T_hN_\theta \times T_f}$) to verify such temporal structures.
We use four different ways to intervene kernel $\mathbf{R}$, including \textbf{reverse}, \textbf{exchange}, \textbf{reset}, and \textbf{randomize}, and their corresponding counterfactual predictions are visualized in \FIG{fig_intervene_R}.

(ii-a) \textbf{Reverse.}
The reverse intervention is to reverse all columns in the $\mathbf{R}$ matrix.
For the uniform $\mathbf{R} = \left(\mathbf{R}_{:1}, \mathbf{R}_{:2}, ..., \mathbf{R}_{:T_f}\right) \in \mathbb{R}^{T_h \times T_f}$, this intervention can be denoted using the do-calculus as
\begin{equation}
    \begin{split}
        &do(\mathbf{R} = \mathbf{R}_{\mathrm{reverse}}), \quad\mbox{where}~ \\
        &\mathbf{R}_{\mathrm{reverse}} = \Big(\mathbf{R}_{:T_f}, \mathbf{R}_{:(T_f - 1)}, ..., \mathbf{R}_{:1}\Big).
    \end{split}
\end{equation}
As visualized in \FIG{fig_intervene_R} (a2), (b2), (c2), we see that all predicted trajectories are temporally reversed after applying such intervention.
Compared to those original predictions in \FIG{fig_intervene_R} (a1), (b1), (c1), we see that those counterfactual predictions converge to a single point at the final temporal step of the prediction period, while exhibiting the greatest uncertainty (multimodal properties) at the first prediction step, which demonstrates completely reversed temporal characteristics.
Furthermore, given that the Haar transform treats every pair of adjacent temporal steps as a single time-frequency step, each two adjacent points still appear in pairs in those counterfactual predictions.
This method of applying interventions on a specific input to observe changes in the outcome while keeping all other conditions constant is a post-hoc verification of the causal relationship between that input and the outcome.
This implies that applying a reverse intervention to the $\mathbf{R}$ matrix will directly affect the corresponding prediction results, providing preliminary validation of the causal relationship between the order of each column in the $\mathbf{R}$ matrix and the corresponding prediction results.

(ii-b) \textbf{Exchange.}
The exchange intervention aims at exchanging specific columns within the $\mathbf{R}$ matrix.
Here, we exchange the first $T_f / 2$ columns with the last $T_f / 2$ columns to perform this intervention (note that this is a uniform representation, which also applies to $\mathbf{R}^i_\SOC$ with $T_hN_\theta$ rows):
\begin{equation}
    \begin{split}
        &do(\mathbf{R} = \mathbf{R}_{\mathrm{exchange}}), \quad\mbox{where}~ \\
        &\mathbf{R}_{\mathrm{exchange}} = \Big(
            \mathbf{R}_{:\left(\frac{T_f}{2} + 1\right)},
            \mathbf{R}_{:\left(\frac{T_f}{2} + 2\right)},
            ...,
            \mathbf{R}_{:\left(\frac{T_f}{2} + \frac{T_f}{2}\right)}, \\
        &\qquad\qquad\qquad\quad \mathbf{R}_{:1},
            \mathbf{R}_{:2},
            ...,
            \mathbf{R}_{:\frac{T_f}{2}}
        \Big).
    \end{split}
\end{equation}
As visualized in \FIG{fig_intervene_R} (a3), (b3), (c3), we see that the original continuous predictions have become two main ``segments'' in each forecasted trajectory.
Specifically, it can be seen that the first half of the prediction indicates an observable spatial jump compared to current observation point while presenting higher multimodal properties.
Differently, between the $(T_f/2)$th and $(T_f/2 + 1)$th temporal prediction step, all predictions suddenly converge to the same point again, and then grow with smaller spatial uncertainty.
In other words, after applying the exchange intervention, the counterfactual predictions present a clear temporal misalignment, where predictions that should have appeared in the latter half of the forecast have been forcibly inserted into the earlier part, while those that should have appeared in the earlier part have been incorrectly placed in the latter half.
It is worth noting that it is these spatial discontinuities that provide stronger evidence of the unique temporal structure of each element in the $\mathbf{R}$ matrix, which ensures that these elements are not averaged out or masked by their neighboring elements.

(ii-c) \textbf{Reset \& Randomize.}
Unlike the above reverse and exchange interventions, the reset and randomize interventions directly set some columns in the reverberation kernel $\mathbf{R}$ with specific values.
The reset intervention manually sets the first $T_f/2$ columns with zero vectors:
\begin{equation}
    \begin{split}
        &do(\mathbf{R} = \mathbf{R}_{\mathrm{reset}}), \quad\mbox{where}~ \\
        &\mathbf{R}_{\mathrm{reset}} = \Big(
            \mathbf{0},
            ...,
            \mathbf{0}, \\
        &\qquad\qquad\quad~ \mathbf{R}_{:\left(\frac{T_f}{2} + 1\right)},
            \mathbf{R}_{:\left(\frac{T_f}{2} + 2\right)},
            ...,
            \mathbf{R}_{:\left(\frac{T_f}{2} + \frac{T_f}{2}\right)}
        \Big).
    \end{split}
\end{equation}
Similarly, the randomize intervention sets the last one column (corresponding to 2 temporal steps since the use of Haar transform) in the $\mathbf{R}$ matrix with the randomly sampled noise vector $\mathbf{z}\sim \mathcal{U}(-1, 1)~(\mathbf{z} \in \mathbb{R}^{T_h})$:
\begin{equation}
    \begin{split}
        &do(\mathbf{R} = \mathbf{R}_{\mathrm{random}}), \quad\mbox{where}~ \\
        &\mathbf{R}_{\mathrm{random}} = \Big(
            \mathbf{R}_{:1},
            \mathbf{R}_{:2},
            ...,
            \mathbf{R}_{:(T_f - 1)},
            \mathbf{z}
        \Big).
    \end{split}
\end{equation}
As visualized in the last two columns in \FIG{fig_intervene_R}, direct prediction changes have been presented on these specific intervened temporal steps.
For example, in \FIG{fig_intervene_R} (a4), (b4), (c4), the first half of each prediction has become the pure linear prediction after the reset intervention on these corresponding columns, with both non-interactive and social predictions strictly remaining zero during the first-half of the prediction period.
Also, after applying the randomize intervention on the last column of the $\mathbf{R}$ matrix, the last two prediction points suddenly change and become randomly distributed around the space.

According to the above interventions on the $\mathbf{R}$ matrix, it can be verified that temporal structures have been structurally embedded in this reverberation kernel.
Exchanging or modifying any columns in such kernel directly leads to corresponding trajectory changes during the prediction period, specifically at the selected temporal steps where those interventions are applied.
This indicates that there are specific temporal semantics structurally embedded in the reverberation kernels, rather than relying on additional soft positional embeddings.
Therefore, this represents another structural difference between the reverberation transform and the operations in attention mechanisms, \IE, the core of the reverberation transform is to learn such directional temporal mappings, rather than potentially irreversible feature weightings or mixings.

\subsection{Why Temporal Convolution Does Not Explicitly Encode Latencies}

The temporal convolution shares a similar form with the proposed reverberation transform.
However, the reverberation transform is not a normal temporal convolution or one of its variations.
Temporal convolutions commonly used in neural networks usually use a trainable (mostly fixed after training) kernel $\mathbf{h}(t)$ to process the input signal.
Denote the observation and prediction periods as $\mathcal{T}_h$ and $\mathcal{T}_f$ (we do not discuss how spectral transforms with different temporal resolutions change these periods here).
Denoting the input sequence as $\mathbf{x}(t)$ and the temporal range where the kernel $\mathbf{h}(t)$ is available as $\mathcal{T}_\delta$, the convolution output $\mathbf{y}(t)$ can be formulated as
\begin{equation}
    \label{eq_appendix_conv}
    \mathbf{y}(t) = \sum_{\delta \in \mathcal{T}_\delta} \mathbf{h}(\delta)\mathbf{x}(t - \delta), \quad\forall t \in \mathcal{T}_h.
\end{equation}

Here, it can be seen that the reverberation transform has a similar mathematical form to such temporal convolution.
However, comparing \EQUA{eq_appendix_conv} and \EQUA{eq_appendix_R_matrix}, two major differences can be seen between the implementation of the reverberation transform and temporal convolutions.
Similar to the above discussions on the attention mechanism, we summarize these differences as \emph{kernel coverages} and \emph{temporality}:

\NEWHALFLINE
(i)~\textbf{Kernel Coverages.}
Comparing \EQUA{eq_appendix_conv} and \EQUA{eq_appendix_R_matrix}, it can be seen that the temporal range for summing input features is different, where the temporal convolution sums features strictly within the range $\mathcal{T}_\delta$ at which the convolution kernel $\mathbf{h}(\delta)$ is defined, while the reverberation transform multiplies and sums features over the entire period $\mathcal{T}_h$ where the input feature $\mathbf{x}(h)$ has been represented.
Specifically, in most computer vision or pattern recognition tasks, the lengths of convolution kernels are usually much shorter than the input signals themselves.
Taking 2D convolutions as an example, the size of some classic convolution kernels (like Sobel filters and VGG filters) is as small as 3x3, yet they have been used in a wide range of image processing tasks where the spatial dimensions of these images are far greater than 100.
This means that the goal of these convolutions, whether 1D or 2D or more, is not the global gathering of all observed features, but how local features are distributed or have been changed around the local period $\{t - \delta\}$ of the currently focused step $t \in \mathcal{T}_h$, according to the size of the convolution kernel $\delta \in \mathcal{T}_\delta$.
However, the goal of the reverberation transform is to map features from the observation period into the prediction period.
From this point of view, all historical steps $h \in \mathcal{T}_h$ should be considered to finish such mappings.
Therefore, for the coverages of kernels, convolution kernels are usually smaller than the input sequences, and they focus more on capturing how the local features are distributed or changed.
Differently, the reverberation kernel $\mathbf{R}$ has the same length as the input sequences, aiming at mapping all the observed features to the future prediction period.

In addition, convolution kernels are mostly static after training.
However, each row in the reverberation kernel $\mathbf{R}$ (and the generating kernel $\mathbf{G}$) corresponds to a specific observation step.
For example, the $h_0$th row of these reverberation kernels is learned from the representation at the corresponding $h_0$th observation step, and it specifically plays the role similar to the system's unit impulse responses at these corresponding observation steps.
In other words, reverberation kernels can be treated as the sets of time-varying responses, which specifically correspond to their unique observation steps.
Therefore, such time-varying properties are another key difference from the fixed convolution kernels.

\NEWHALFLINE
(ii)~\textbf{Temporality.}
Similar to our summarized temporality properties in attention mechanisms, temporality is also the other major difference between the reverberation transform and temporal convolutions.
As we previously discussed, the reverberation transform aims at mapping features from one period (observation) to another (prediction), while attention aims at enhancing features from one period (observation) by querying features and computing their similarities within such period.
For the temporal convolution, this is also straightforward, according to the range of variable $t \in \mathcal{T}_h$ and $f \in \mathcal{T}_f$ in the above \EQUA{eq_appendix_conv,eq_appendix_R_matrix} directly, where temporal convolutions used in most neural networks do not change the period where such input representations reside (especially under the ``same'' padding strategy).
Differently, the output of the reverberation transform lies in a different period from its input sequential features $\mathbf{x}(h)$.
Therefore, this is another major difference between those approaches.

\end{document}

\documentclass[../../paper.tex]{subfiles}

\begin{document}

\section{Latencies for Motion Prediction}
\label{sec_appendix_motion}

Both the reverberation transform and the \MODELSHORT~model are designed for the trajectory prediction task.
In this section, we further discuss their generalizability for other sequence forecasting tasks, taking the motion prediction dataset \textbf{Human3.6M}~\cite{ionescu2013human3,ionescu2011latent} (3D skeleton prediction) as an example.

\subsection{Datasets, Settings, and Metrics}
Human3.6M is a large-scale dataset with 3.6 million 3D human poses by 11 professional actors in 15 scenarios (such as discussion, smoking, taking photos, and talking on the phone).
Each scene only contains one subject.
Following some previous works \cite{wong2023another,xu2023eqmotion}, we regard this task as a 3D point forecasting task, where each frame contains a human skeleton made up of 17 3D points, by observing $t_h = 10$ frames of positions and forecasting the next $t_f = 10$ frames of positions under the sample rate of 25 Hz ($\Delta t = 0.04$s).
Following their settings, subjects \{1, 6, 7, 8, 9\} will be used for training, subject 11 for validating, and subject 5 for testing.
Compared to our discussed trajectory prediction scenes, this setting brings far more data dimensions and complexities, making it more challenging to forecast such human motions in the form of skeletons.

When training on Human3.6M, feature dimension $d$ has been extended to 512 to suit the larger trajectory dimensionality.
All noise-samplings in the \MODELSHORT~model are disabled to ensure the deterministic skeleton prediction.
Also, social-interaction-modeling networks are removed from the original model, meaning that only non-interactive latencies will be considered when forecasting motions under this setting.
When training, the learning rate (Adam) is set to 2e-4, and the maximum training epochs are set to 200.
Other settings are the same as mentioned in the main manuscript.

Following previous works, we use the Mean Per Joint Position Error (MPJPE) to measure prediction performance (in millimeters).
It is the average $\ell_2$ distance between each predicted joint and its groundtruth, at some specific prediction step.
For our considered $\{t_h, t_f, \Delta t\} = \{10, 10, 0.04\}$ settings, MPJPEs at the 2nd/4th/8th/10th prediction steps (80/160/320/400 milliseconds) will be reported and compared.

\subsection{Quantitative Analyses}


\documentclass[../../paper.tex]{subfiles}

\begin{document}

\begin{table}[tbp]
    \caption{
        Comparisons to other recent baseline methods on Human3.6M dataset.
        Metrics are reported as the MPJPE at different prediction horizons, including 80/160/320/400 milliseconds.
    }
    \label{tab_sota_motion}
    \centering
    \begin{tabular}{lcccc}
        \toprule
        Models & 80ms & 160ms & 320ms & 400ms \\
        
        \midrule
        Traj-GCN~\cite{mao2019learning}     & 26.1      & 52.3      & 63.5      & 63.5      \\
        MSR-GCN~\cite{dang2021msr}          & 12.1      & 25.6      & 51.6      & 62.9      \\
        PGBIG~\cite{ma2022progressively}    & 10.3      & 22.7      & 47.4      & 58.5      \\
        SPGSN~\cite{li2022skeleton}         & 10.4      & 22.3      & 47.1      & 58.3      \\
        EqMotion~\cite{xu2023eqmotion}      & \C{9.1}   & 20.1      & \C{43.7}  & \C{55.0}  \\
        \VMODEL~(Haar)~\VCITE               & \C{8.7}   & \C{19.7}  & 45.5      & 59.5      \\
        \EVMODEL~(Haar)~\EVCITE             & \C{9.0}   & \C{19.6}  & \C{44.4}  & \C{57.8}  \\

        \midrule
        \MODELSHORT~(Ours)                  & 9.5       & \C{18.9}  & \C{41.5}  & \C{53.7}  \\
    
        \bottomrule
    \end{tabular}

    \centering
\end{table}

\end{document}

\TABLE{tab_sota_motion} reports the motion prediction performance of the proposed \MODELSHORT~model and several recent baseline methods.
It can be seen that notable performance gains have been achieved by the proposed \MODELSHORT.
Compared to the newly published \EVMODEL, \MODELSHORT~obtains up to 7.1\% lower MPJPE across all prediction horizons.
Notably, both \VMODEL~and \EVMODEL~report their results using the Haar transform, which directly shows the performance improvements brought by the two proposed reverberation kernels, since the same usage of Haar transform in \MODELSHORT.
Compared to another sota \EQMOTION, the proposed \MODELSHORT~outperforms it for about 2.4\% lower MPJPE at 400ms.
Note that such validation is only conducted to roughly verify the overall usefulness of the proposed \MODELSHORT~model for handling different sequence forecasting tasks.
Therefore, only this well-applied dataset and limited number of baselines have been used for comparison.
Thus, we have omitted further quantitative analyses like detailed ablations and statistical verifications.

However, \MODELSHORT~performs not as well as other baselines like \EQMOTION~and \EVMODEL~at the relatively short prediction horizon (MPJPE at 80ms, for about 5\% worse).
Interestingly, such performance drop aligns with the way how humans react to move their bodies physically, \IE, there will always be latencies for changing their postures, meaning that those short-period motions may be challenging for the reverberation transform to model and forecast.
Also, please note that only non-interactive latencies will be considered, meanwhile generating kernels $\mathbf{G}$ will degenerate into a set of constant values, since there are no human-human interactions and multi-generation needs under this motion-prediction-setting, leading to the limitations of model capabilities.
Nevertheless, \MODELSHORT~still shows competitive performance for longer motion prediction periods, indicating the potential usefulness of the proposed reverberation transform and the \MODELSHORT~model beyond the only trajectory prediction task that it is specifically designed for.

\subsection{Discussions of Latencies within Motions}


\documentclass[../../paper.tex]{subfiles}

\begin{document}

\begin{figure*}[tbp]
    \centering
    \includegraphics[width=1.0\linewidth]{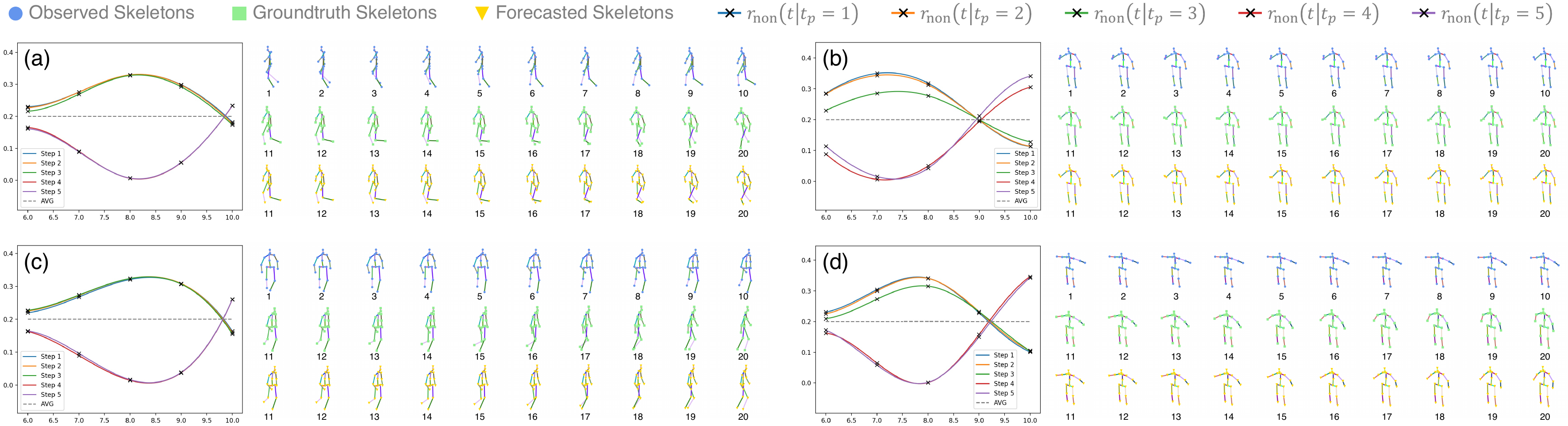}
    \caption{
        Examples of non-interactive reverberation strength $r_\NON (t|t_p)~(t_p \in {1, 2, 3, 4, 5})$ when forecasting human motions (3D skeletons, $\{T_h, T_f\} = \{5, 5\}$).
    }
    \label{fig_latency_motion}
\end{figure*}

\end{document}

Similar to our discussions in the main manuscript, we now discuss how agents handle trajectory-changing events when planning their future motions in the form of 3D skeletons.
For clarity, we still refer to words like motions or skeletons as \emph{trajectories}.
Since the removal of social-interaction-modeling networks and multiple generations in kernel $\mathbf{G}$, this section only discusses non-interactive latencies within human motions.
\FIG{fig_latency_motion} visualizes several non-interactive reverberation strengths computed from the non-interactive kernel $\mathbf{R}_\NON$.
Similar to our discussions in the main manuscript, 1 step in reverberation curves refers to 2 adjacent real-world frames (80ms per step) due to the use of Haar transform.

Focusing on \FIG{fig_latency_motion} (a), it can be seen that this subject is observed running during the past 10 frames (5 steps), with its right leg moving forward (since frame 1 / step 1) while its left leg moving backward (since about frame 6 / step 3 or 4).
Its groundtruth is in line with our intuition of running experiences that this subject may first keep its right leg still, then move its left leg forward after certain latencies of balancing its body (at about frame 15 / step 8 in the groundtruth).
Such latency trends have been well captured and predicted by the kernel $\mathbf{R}_\NON$, as illustrated by the reverberation curve in \FIG{fig_latency_motion} (a).
It predicts that the first 3 observation steps (frames 1 - 6) will be first concentrated and handled during steps 6 to 8, corresponding to frames 11 - 16, which aligns with our ``body-balancing'' assumption.
After that, the following 2 observation steps (frames 7 - 10) will be gradually taken into account in the further future, starting from step 8 - 9 (frames 15 - 18), thus presenting the trend of moving its left leg forward in the predictions.
During this process, the latency of about 8 frames of body-balancing has been well captured and predicted.

Other examples in \FIG{fig_latency_motion} also show unique latencies within human motions.
In \FIG{fig_latency_motion} (b), where the subject is finishing smoking and pushing the cigarette butt away, a latency of about 3 steps (6 frames) to shake off the ashes has been predicted, and the motion of putting its hand down has been well forecasted after such latency, starting from about step 8 (frame 16).
Latencies of about 3 - 4 steps (6 - 8 frames) have also been captured in \FIG{fig_latency_motion} (c) and (d) for forecasting different types of leg movements when running and lower arms from the T-pose then entering a relaxed state.
Therefore, latencies within motions could still be learned and forecasted when forecasting, even though the proposed \MODELSHORT~is not specifically designed for this task while some of the model components have been disabled.
Specifically, these latencies typically fall between 6 - 8 frames (240ms - 320ms) across different actions, suggesting a consistent temporal delay in human motion responses, which these motion prediction baselines could hardly learn and explicitly forecast.
Such results could verify the usefulness of reverberation transform and the \MODELSHORT~model for learning and handling latencies beyond the trajectory prediction task.

\end{document}

\documentclass[../../paper.tex]{subfiles}

\begin{document}

\section{Further Discussions on Model Generalizability}
\label{sec_gen}

\subsection{Trajectory Generating Capabilities}

In \MODELSHORT, generating kernel $\mathbf{G}$ provides multiple latency preferences for the same ego agent to make decisions upon different trajectory-changing events, simulating stochastic properties in future trajectories.
We now further discuss its generalizability, taking \MSN\MSNCITE~as the baseline.

We first discuss the kernel shape ($K_g$).
It determines the number of trajectories generated during training.
Like symbols used in \SOCIALGAN\SOCIALGANCITE, we use ``$n$V-$m$'' to denote training-test configuration that generates $n$ trajectories during training and test models under the \emph{best-of-$m$} validation.
In \TABLE{tab_generalizability}, we see that the number of generated trajectories during training ($K_g$ in \MODELSHORT~models) heavily affects how they perform.
Especially, results indicate that, with symbol $m$V-$n$, \MODELSHORT~models could achieve better performance when $m$ is slightly larger than $n$, applicable to both $K_g$V-10 and $K_g$V-20 cases.
Such results could prove the learning capability of the generating kernel $\mathbf{G}$ during the model training while also revealing limitations in generating trajectories far beyond its capacity (like using a $\mathbf{G}$ with $K_g = 1$ to generate 40 stochastic trajectories).


\documentclass[../../paper.tex]{subfiles}

\begin{document}

\begin{table}[tbp]
    \caption{
        Generalizability Tests
    }
    \label{tab_generalizability}
    \centering
    \begin{threeparttable}
        \begin{tabular}{lcccccccccccccc}
            \toprule
            ID & $K_g$ & $K_g$V-10 & $K_g$V-20 & $K_g$V-40 \\
            
            \midrule
            b1 & 1      & 0.401/0.851       & 0.397/0.839       & 0.393/0.828 \\
            b2 & 2      & 0.331/0.677       & 0.327/0.667       & 0.324/0.658 \\
            b3 & 5      & 0.253/0.491       & 0.246/0.474       & 0.240/0.458 \\
            b4 & 10     & \CC{0.211/0.390}  & 0.206/0.377       & 0.202/0.364 \\
            b5 & 20     & \C{0.242/0.466}   & \CC{0.172/0.283}  & \C{0.167/0.272} \\
            b6 & 30     & \C{0.247/0.473}   & \C{0.191/0.334}   & \C{0.163/0.263} \\
            b7 & 40     & 0.252/0.493       & \C{0.192/0.339}   & \CC{0.146/0.219} \\

            \bottomrule
        \end{tabular}
        \begin{tablenotes}[flushleft]
            \footnotesize
            \item[]
                Results are reported in meters on zara1, using ``$n$V-$m$'' to denote results under \emph{best-of-$m$} for models trained with $n$ generations during training.
        \end{tablenotes}
    \end{threeparttable}
\end{table}

\end{document}

Now we further discuss how $K_g$ affects \MODELSHORT~models' performance by fixing the number of generated trajectories during testing (let $m = K$ in $n$V-$m$).
Several \MODELSHORT-\MSN~variations have been conducted to serve as baselines by replacing generating kernels $\mathbf{G}$ with the \MSN-like generating strategy \MSNCITE.
Results in \FIG{fig_gen_msn} indicate that the generalizability of kernel $\mathbf{G}$ is worse than MSN when only predicting one trajectory for each agent.
However, \MODELSHORT~models perform much better when generating more trajectories.
For example, \MODELSHORT~($K_g = 5$) outperforms the corresponding MSN variation MSN* (5) by about 4.28\% in ADE when generating $K = 1000$ trajectories, while \MODELSHORT~($K_g = 10$) even outperforms its MSN variation for about 5.12\% when generating $2000$ trajectories.
Such results demonstrate the effectiveness of the kernel $\mathbf{G}$ in generating multiple latency preferences for handling trajectory-changing events, especially when more trajectories are required.


\documentclass[../../paper.tex]{subfiles}

\begin{document}

\begin{figure}[tbp]
    \centering
    \includegraphics[width=1.0\linewidth]{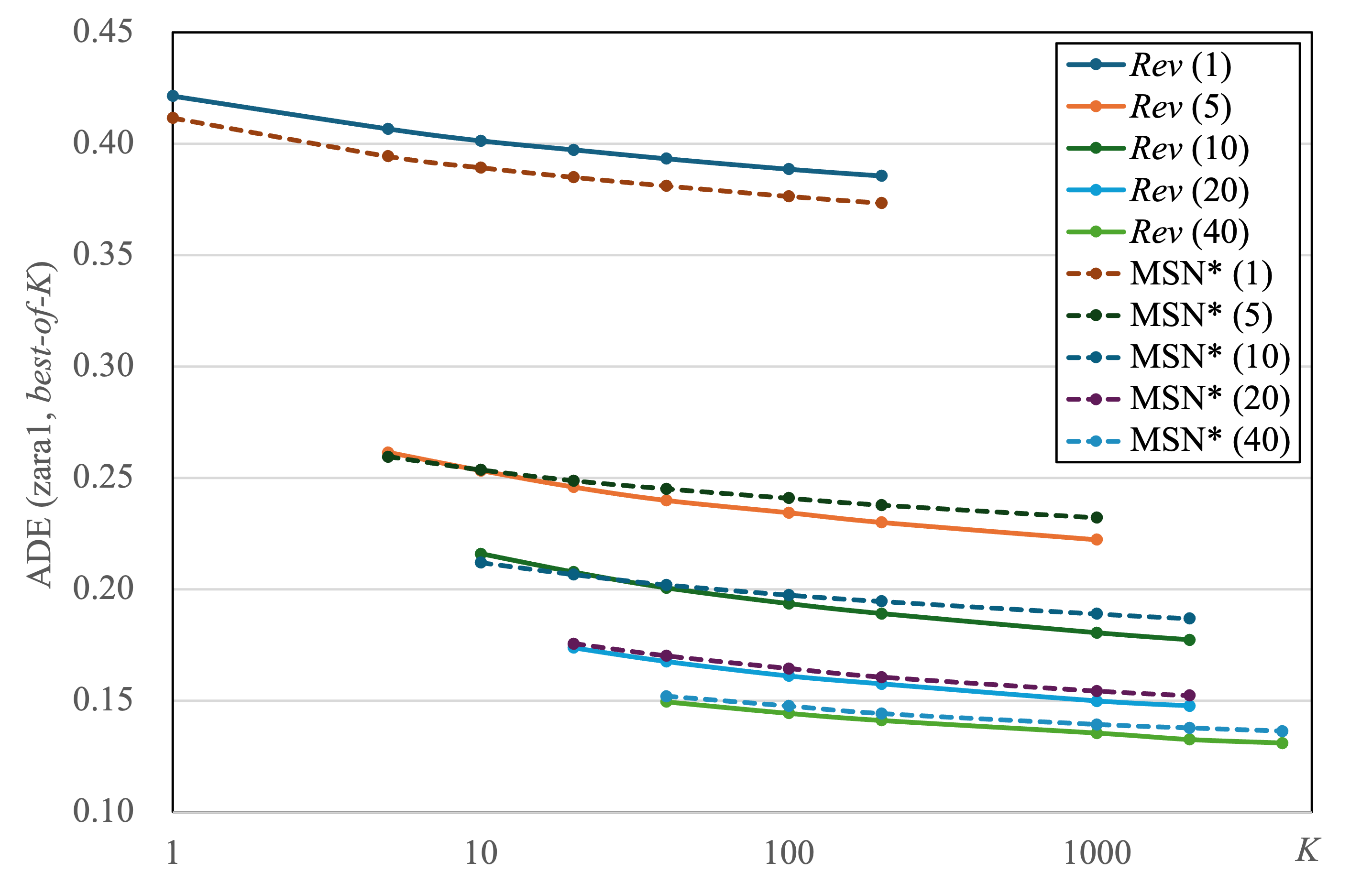}
    \caption{
        Comparisons of generalizability ($K_g$V-$K$ ADE on zara1) with \MSN\MSNCITE.
        Numbers in brackets indicate $K_g$ for \MODELSHORT~variations, and indicate the number of style channels for \MSN~variations.
    }
    \label{fig_gen_msn}
\end{figure}

\end{document}

\NEWHALFLINE
We have also conducted experiments to verify the proposed \MODELSHORT~model under the deterministic trajectory prediction setting, where only one trajectory will be forecasted to each ego agent when training and evaluating.
Accordingly, we have selected the following methods as baselines:
\SOCIALLSTM\SOCIALLSTMCITE
, \SOCIALGAN\SOCIALGANCITE
, \SRLSTM\SRLSTMCITE
, \SRPAMI\SRPAMICITE
, \STAR\STARCITE
, \STGCNN\STGCNNCITE
, \TRAJECTRONPP\TRAJECTRONPPCITE
, \YNET\YNETCITE
, SGCN \cite{shi2021sgcn}
, \AGENTFORMER\AGENTFORMERCITE
, \CSCNET\CSCNETCITE
, Social-Implicit \cite{mohamed2022social}
, \SSL\SSLCITE
, \MSN\MSNCITE
, DSTIGCN \cite{chen2025dstigcn}
, PCHGCN \cite{chen2025pchgcn}
, and MSDHGNN \cite{sang2025msdhgnn}.

To evaluate the deterministic prediction performance of the \MODELSHORT~model, we entirely disable the generating kernel $\mathbf{G}$ as well as other noise-sampling-related network components.
Note that such disable is different from the symbol ``\xm'' used to conduct ablation variations in the Tabs. V and VI of the main text, where such symbols under the generating kernel $\mathbf{G}$ indicate to disable this kernel and use a basic \MSN-like network with similar number of parameters to make stochastic predictions.
Differently, both the generating kernel and other noise-samplings have been removed when making deterministic predictions.
Results are reported in \TABLE{tab_ethucy_det}.

\begin{table*}
    \caption{
        Comparisons of Deterministic Models on ETH-UCY
    }
    \label{tab_ethucy_det}
    \centering
    \begin{threeparttable}
        \begin{tabular}{lcccccc}
            \toprule
            Method (Deterministic) & eth $\downarrow$ & hotel $\downarrow$ & univ $\downarrow$ & zara1 $\downarrow$ & zara2 $\downarrow$ & Average $\downarrow$ \\
            \midrule

            \SOCIALLSTM\SOCIALLSTMCITE~(2016)         & 1.09/2.35 & 0.79/1.76 & 0.67/1.40 & 0.47/1.00 & 0.56/1.17 & 0.72/1.54 \\
            \SOCIALGAN\SOCIALGANCITE~(2018)           & 1.14/2.25 & 0.78/1.68 & 0.77/1.68 & 0.64/1.40 & 0.55/1.21 & 0.78/1.64 \\
            \SRLSTM\SRLSTMCITE~(2019)                 & 0.63/1.25 & 0.37/0.74 & 0.51/1.10 & 0.41/0.90 & 0.32/0.70 & 0.45/0.94 \\
            \SRPAMI\SRPAMICITE~(2020)                 & 0.58/1.13 & 0.31/0.62 & 0.50/1.10 & 0.41/0.90 & 0.33/0.73 & 0.43/0.89 \\
            \STAR\STARCITE~(2020)                     & 0.56/1.11 & 0.26/0.50 & 0.52/1.13 & 0.40/0.89 & 0.31/0.71 & 0.41/0.87 \\
            \STGCNN\STGCNNCITE~(2020)                 & 1.05/1.88 & 0.80/1.50 & 0.74/1.42 & 0.65/1.22 & 0.57/1.06 & 0.76/1.42 \\
            \TRAJECTRONPP\TRAJECTRONPPCITE~(2020)     & 1.04/2.12 & 0.37/0.67 & 0.56/1.24 & 0.47/1.03 & 0.34/0.75 & 0.56/1.16 \\
            \YNET\YNETCITE~(2021)                     & 1.60/3.23 & 0.61/1.21 & 1.46/3.09 & 1.13/2.33 & 0.79/1.65 & 1.12/2.30 \\
            SGCN~\cite{shi2021sgcn}~(2021)            & 0.90/1.82 & 0.57/1.15 & 0.61/1.23 & 0.50/1.03 & 0.37/0.76 & 0.59/1.20 \\
            \AGENTFORMER\AGENTFORMERCITE~(2021)       & 1.42/2.96 & 0.73/1.61 & 0.84/1.85 & 0.74/1.64 & 0.72/1.65 & 0.89/1.94 \\
            \CSCNET\CSCNETCITE~(2022)                 & 0.51/1.05 & 0.22/0.42 & 0.36/0.81 & 0.31/0.68 & 0.47/1.02 & 0.37/0.79 \\
            Social-Implicit~\cite{mohamed2022social}~(2022) & 1.06/2.34 & 0.77/1.48 & 0.73/1.50 & 0.57/1.15 & 0.54/1.12 & 0.73/1.52 \\
            \SSL\SSLCITE~(2022)                       & 0.69/1.37 & 0.24/0.44 & 0.51/0.93 & 0.42/0.84 & 0.34/0.67 & 0.44/0.85 \\
            \MSN\MSNCITE~(2023)                       & 0.65/1.27 & 0.23/0.42 & 0.62/1.23 & 0.46/0.91 & 0.41/0.81 & 0.47/0.93 \\
            DSTIGCN~\cite{chen2025dstigcn}~(2025)     & 0.96/1.99 & 0.59/1.19 & 0.61/1.24 & 0.48/1.03 & 0.40/0.85 & 0.61/1.26 \\
            PCHGCN~\cite{chen2025pchgcn}~(2025)       & 0.92/2.00 & 0.38/0.80 & 0.51/1.12 & 0.40/0.87 & 0.31/0.68 & 0.50/1.09 \\

            \midrule
            \MODELSHORT~(Ours)                        & 0.54/1.09 & 0.26/0.48 & 0.56/1.20 & 0.42/0.90 & 0.33/0.69 & 0.42/0.87 \\
            \bottomrule
        \end{tabular}
        \begin{tablenotes}[flushleft]
            \footnotesize
            \item[]
                Metrics are reported as ``ADE/FDE'' under the deterministic trajectory prediction setting.
                Lower metrics indicate better prediction performance.
        \end{tablenotes}
    \end{threeparttable}
\end{table*}

As reported in \TABLE{tab_ethucy_det}, it can be seen that even under the strict deterministic prediction setting, where the generating kernel $\mathbf{G}$ is entirely disabled, our proposed \MODELSHORT~still achieves competitive performance against various deterministic baselines.
Specifically, our method performs similarly or even slighly outperforms several recent GNN-based approaches (such as PCHGCN and DSTIGCN).
This further demonstrates the representation capability and generalizability of the proposed reverberation transform in capturing temporal responses and social interactions, even without relying on its full stochastic generating abilities.

\subsection{Latencies and Sampling Rates of Trajectories}

Followed by the common settings used by other trajectory prediction approaches, we mostly conduct trainings and tests with a fixed sampling rate on each dataset.
Here, we conduct additional trainings and tests with different sampling rates to further discuss how they affect the learned reverberation kernels.
For ETH-UCY, the original sampling interval is 0.4 seconds (with the sampling rate of 2.5Hz).
We have conducted additional trainings of the proposed \MODELSHORT~model using intervals of 0.8 seconds and 1.2 seconds on these ETH-UCY datasets, respectively, and visualized the corresponding non-interactive reverberation strengths (curves) computed from these new $\mathbf{R}$ kernels.
To ensure fair comparisons, we use the same prediction length settings ($\{t_h, t_f\} = \{8, 12\}$ and $\{T_h, T_f\} = \{4, 6\}$ correspondingly after the application of Haar transform), keep all model parameters identical, and compute the average curve across all test samples in the dataset to avoid random variations when training.

\begin{figure}[tbp]
    \centering
    \includegraphics[width=\linewidth]{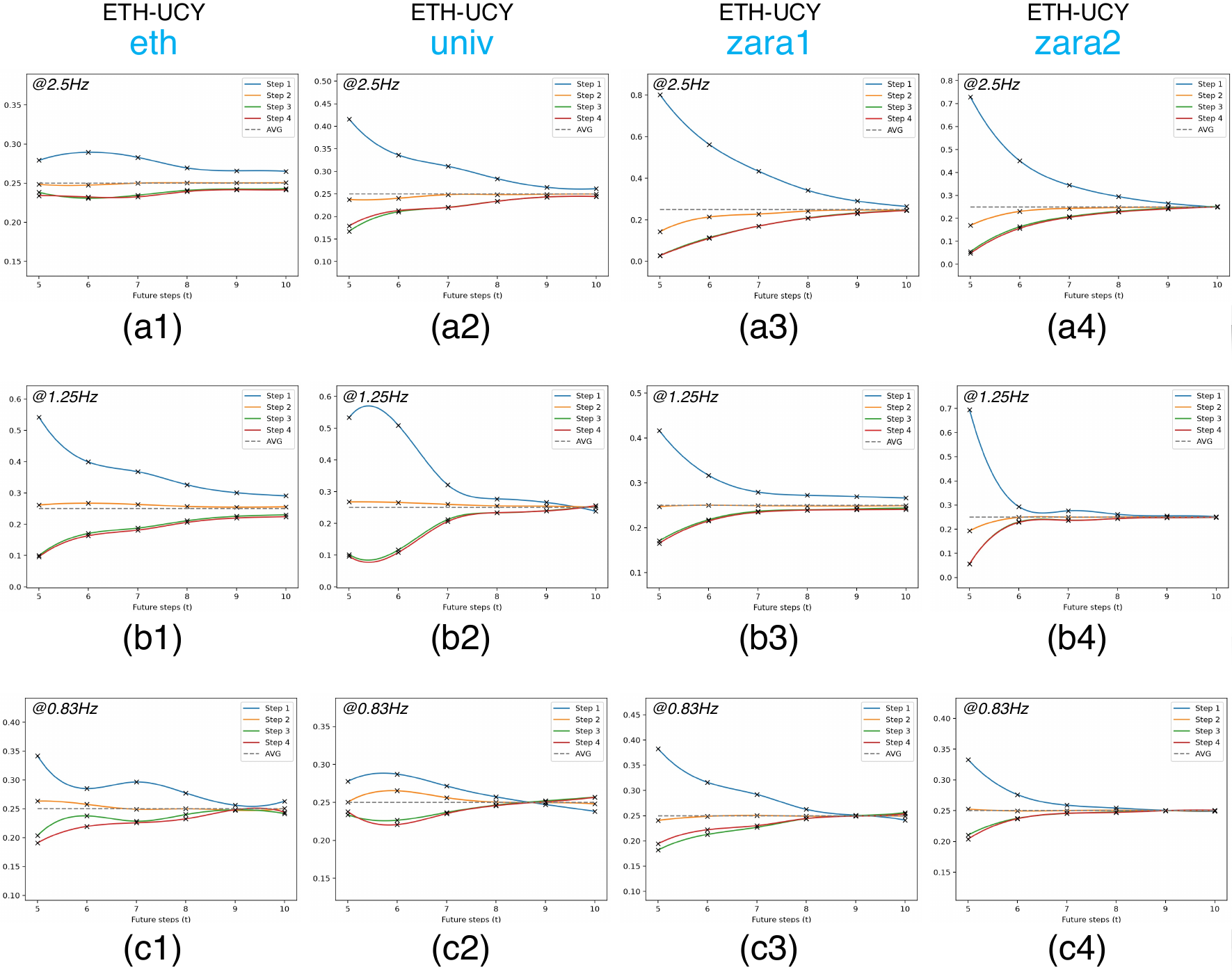}
    \caption{
        Non-interactive reverberation strengths computed from \MODELSHORT~models trained with different sampling rates of trajectories.
        Cases in (a1) to (a4) are trained with the sample rate of 2.5Hz (0.4s of sampling interval), 1.25Hz in the second row ((b1) to (b4)), and 0.83Hz in the third row ((c1) to (c4)).
    }
    \label{fig_sample_rate}
\end{figure}

As illustrated in \FIG{fig_sample_rate}, it can be seen that reverberation curves (kernels) learned under different sampling rates of trajectories could be adaptively adjusted to present different temporal properties.
Comparing the first two rows in \FIG{fig_sample_rate}, we observe that reverberation curves learned from trajectories at 1.25Hz have been ``compressed horizontally'' across the same number of future steps than those trained at 2.5Hz.
Specifically, we notice that this horizontal scaling is approximately 0.5, corresponding to the change of sampling rate from 2.5Hz to 1.25Hz.
For example, in case (a4), the top blue curve takes 5 steps to drop from its maximum value to the average level (which is defined as the ``vanishing latency'' $L_v$).
Differently, the same curve presents about 2 to 3 steps of such latency in case (b4).
Cases (b1) to (b4) all show the same trend, where almost all curves stop changing within 3 more future steps.
In other words, the linear scaling property of the reverberation kernel can be considered valid when the sampling interval becomes 0.8s.

However, it should be noted that this horizontal linear scaling of the curves does not become significantly apparent when the sampling interval is increased to 1.2 seconds (at a sampling rate of 0.83 Hz).
Instead, the amplitudes of these curves become much closer to each other.
In cases (a4) and (b4), the maximum values of the curves are both greater than 0.7.
However, in case (c4), this maximum value drops sharply to around 0.34.
This means that at slower sampling rates, the differences in the relative contributions of all past events to the future trajectory become similar to one another.
Specifically, at this sampling rate, 8 frames of observed trajectories correspond to 9.6 seconds in the real world, while 12 frames of predictions correspond to 14.4 seconds.
In the ETH-UCY pedestrian walking scene, this sampling frequency may already be lower than twice of the actual frequency of trajectory change events.
According to Nyquist's theorem, a signal can only be fully reconstructed when the sampling frequency is at least twice the signal's frequency.
Interestingly, most of the latencies we discussed in the earlier sections also fall within the range of about 1.6 to 6.4 seconds.
Based on these results, sampling the trajectory at a 1.2 second interval just fails to satisfy the Nyquist theorem, where the interval of 0.8 seconds has become the rough threshold to maintain these latencies on ETH-UCY.
Therefore, such sampling may have resulted in information loss, making it fail to reconstruct the true data.
This also explains why the reverberation curves no longer follow the linear scaling properties at this sampling rate.

Given the changes in these curves as the sampling rate varies from 2.5 Hz to 1.25 Hz, we conclude that the linear latency-scaling properties remain valid within a specific sampling frequency range.
This implies that the reverberation kernels could learn the temporal latency properties embedded in the trajectory data, rather than the fixed data weightings assigned for each temporal step.

\end{document}

\documentclass[../../paper.tex]{subfiles}

\begin{document}

\section{Statistical Verifications}
\label{sec_appendix_statis}

Like most recent trajectory prediction approaches, the proposed \MODELSHORT~model could generate multiple random trajectories for each ego agent.
Although metrics like Negative Log-Likelihood (NLL) have been introduced to measure how the generated trajectory-distribution matches their original ones, most current approaches still cannot better describe how well the diversities are of their multiple predicted trajectories.
In addition, ADE and FDE are still the first choices of most researchers to measure their prediction performance.
Therefore, this section further discusses the ADE/FDE of several recent baselines and the proposed \MODELSHORT~model, thus verifying potential statistical properties of these models straightforwardly.

In the main manuscript, we refer to ``minADE$_K$/minFDE$_K$'' as ``ADE/FDE''.
Differently, in this section, these metrics will be strictly distinguished, where ``ADE/FDE'' only indicates metrics computed from one generated trajectory, \IE, the ``minADE$_1$/minFDE$_1$''.
We will report detailed results in the form of ``meanADE$_{K}$ $\pm$ stdADE$_{K}$'' and ``meanFDE$_{K}$ $\pm$ stdFDE$_{K}$'' to verify models' statistical performance.
For ADE, these metrics are defined as
\begin{align}
    \mbox{meanADE}_{K} &= \operatorname*{mean}_{k \in \{1, 2, ..., K\}}
    \frac{1}{t_f} \sum_{t = t_h + 1}^{t_h + t_f} \left\|
        \hatbf{p}^i_{t_k} - \mathbf{p}^i_t
    \right\|_2, \\
    \mbox{stdADE}_{K} &= \operatorname*{std}_{k \in \{1, 2, ..., K\}}
    \frac{1}{t_f} \sum_{t = t_h + 1}^{t_h + t_f} \left\|
        \hatbf{p}^i_{t_k} - \mathbf{p}^i_t
    \right\|_2.
\end{align}
Similar computations are applied to compute meanFDE$_{K}$ and stdFDE$_{K}$, based on the Eq. (20) in the main manuscript.


\documentclass[../../paper.tex]{subfiles}

\begin{document}

\begin{table*}[tbp]
    \caption{
        Statistical verifications of \MODELSHORT~model on zara1.
        Results are reported with ``mean $\pm$ std'' computed from 10 model runs.
    }
    \label{tab_meanstd_zara1}
    \centering
    \begin{tabular}{lccccccccccc}
        \toprule
        $K$ & minADE$_K$ & meanADE$_K$ & stdADE$_K$ & minFDE$_K$ & meanFDE$_K$ & stdFDE$_K$ \\

        \midrule
        1   & 0.7287 $\pm$ 0.2243 & 0.7654 $\pm$ 0.1903 & -                   & 1.6113 $\pm$ 0.5291 & 1.6768 $\pm$ 0.4327 & -                   \\
        2   & 0.5354 $\pm$ 0.1020 & 0.7541 $\pm$ 0.0826 & 0.3704 $\pm$ 0.0428 & 1.1541 $\pm$ 0.2369 & 1.6528 $\pm$ 0.2004 & 0.8522 $\pm$ 0.1152 \\
        4   & 0.3678 $\pm$ 0.0483 & 0.7977 $\pm$ 0.0341 & 0.4395 $\pm$ 0.0475 & 0.7587 $\pm$ 0.1081 & 1.7557 $\pm$ 0.0771 & 1.0262 $\pm$ 0.1230 \\
        8   & 0.2724 $\pm$ 0.0137 & 0.7666 $\pm$ 0.0288 & 0.4338 $\pm$ 0.0354 & 0.5330 $\pm$ 0.0264 & 1.6807 $\pm$ 0.0678 & 1.0086 $\pm$ 0.0850 \\
        12  & 0.2145 $\pm$ 0.0036 & 0.7331 $\pm$ 0.0279 & 0.4201 $\pm$ 0.0173 & 0.3934 $\pm$ 0.0090 & 1.6067 $\pm$ 0.0649 & 0.9764 $\pm$ 0.0420 \\
        16  & 0.1924 $\pm$ 0.0021 & 0.7577 $\pm$ 0.0089 & 0.4398 $\pm$ 0.0048 & 0.3343 $\pm$ 0.0061 & 1.6628 $\pm$ 0.0195 & 1.0228 $\pm$ 0.0101 \\
        20  & 0.1730 $\pm$ 0.0002 & 0.7482 $\pm$ 0.0002 & 0.4348 $\pm$ 0.0001 & 0.2836 $\pm$ 0.0009 & 1.6413 $\pm$ 0.0005 & 1.0122 $\pm$ 0.0005 \\
        40  & 0.1682 $\pm$ 0.0001 & 0.7481 $\pm$ 0.0001 & 0.4292 $\pm$ 0.0001 & 0.2693 $\pm$ 0.0006 & 1.6414 $\pm$ 0.0001 & 0.9993 $\pm$ 0.0002 \\
        60  & 0.1656 $\pm$ 0.0001 & 0.7481 $\pm$ 0.0001 & 0.4275 $\pm$ 0.0001 & 0.2620 $\pm$ 0.0001 & 1.6414 $\pm$ 0.0001 & 0.9952 $\pm$ 0.0001 \\
        100 & 0.1630 $\pm$ 0.0001 & 0.7481 $\pm$ 0.0000 & 0.4260 $\pm$ 0.0001 & 0.2544 $\pm$ 0.0003 & 1.6415 $\pm$ 0.0001 & 0.9918 $\pm$ 0.0001 \\
        \bottomrule
    \end{tabular}
\end{table*}

\end{document}

Take zara1 (a subset in ETH-UCY) as an example, \TABLE{tab_meanstd_zara1} has reported several statistical ADE/FDE results, under different number of generations $K$.
Note that all reported results are averaged over 10 runs to eliminate run-time randomness, therefore each single metric will be also reported in the form of ``mean $\pm$ std''.

From \TABLE{tab_meanstd_zara1}, we observe that the proposed \MODELSHORT~model exhibits stable and convergent statistical behaviors as the number of generations $K$ increases.
Specifically, both meanADE$_K$ and meanFDE$_K$ remain nearly constant when $K \ge 8$, with marginal variations smaller than $0.02$, indicating that the overall distribution of generated trajectories is statistically stable even under multi-sample stochastic generation.
Meanwhile, the minimum metrics (minADE$_K$, minFDE$_K$) consistently decrease as $K$ grows, verifying that the model can effectively produce multiple plausible trajectories and select more accurate ones when larger $K$ values are available.
This trend reflects a reasonable sampling diversity within the learned distribution rather than mere noise perturbation.

The standard deviations (stdADE$_K$, stdFDE$_K$) further quantify such diversity.
When $K$ increases from $2$ to $12$, stdADE$_K$ and stdFDE$_K$ gradually increase and then stabilize around $0.43$ and $1.00$, respectively.
This indicates that the prediction spread among different generations is bounded and statistically consistent, suggesting that the diversity originates from meaningful variations in planning hypotheses rather than random fluctuations.
Moreover, the extremely small deviations (on the order of $10^{-4}$) for large $K$ (e.g., $K \ge 20$) demonstrate that the \MODELSHORT~model reaches an equilibrium sampling state where additional generations no longer change the statistical expectation of prediction performance.

Overall, these results confirm that the proposed reverberation-based formulation not only improves the best-case forecasting accuracy (minADE/minFDE) but also maintains a stable and interpretable trajectory distribution across multiple stochastic generations.
The statistical convergence of ADE/FDE validates the robustness of the learned latency representations and the reliability of the \MODELSHORT~model under random sampling conditions.

\end{document}

\documentclass[../../paper.tex]{subfiles}

\begin{document}

\section{Efficiency Analyses}
\label{sec_appendix_efficiency}


\documentclass[../../paper.tex]{subfiles}

\begin{document}

\begin{table}[tbp]
    \caption{
        Inference Times and Memory Usages of \MODELSHORT~Models ($t_h = 8$, $t_f = 12$, forecasting 20 trajectories for each ego, on Apple M1 Mac mini).
    }
    \label{tab_efficiency}
    \centering
    \begin{tabular}{lcccccccc}
        \toprule
        Model & Parameters &
        \makecell{ADE/FDE\\(zara1)} &
        \makecell{Batch\\Size} &
        \makecell{Time\\(ms)} &
        \makecell{Memory\\(MB)} \\

        \midrule
        \multirow{3}{*}{\MODELSHORT} &
        \multirow{3}{*}{3,156,220} &
        \multirow{3}{*}{0.172/0.283}
        & 1         & 30    & 37.14 \\
        &&& 100     & 47    & 115.09 \\
        &&& 1000    & 190   & 535.30 \\

        \midrule
        \multirow{3}{*}{\MODELSHORT-a6} &
        \multirow{3}{*}{\makecell{1,216,670\\(w/o $\Delta \hatbf{Y}^i_\NON$)}} &
        \multirow{3}{*}{0.174/0.294}
        & 1         & \C{17}    & \CC{10.08} \\
        &&& 100     & 31        & 78.03 \\
        &&& 1000    & 159       & 395.99 \\

        \midrule
        \multirow{3}{*}{\MODELSHORT-a7} &
        \multirow{3}{*}{\makecell{2,079,710\\(w/o $\Delta \hatbf{Y}^i_\SOC$)}} &
        \multirow{3}{*}{0.176/0.297}
        & 1         & \CC{15}   & \C{16.21} \\
        &&& 100     & 22        & 26.29 \\
        &&& 1000    & 45        & 116.70 \\
        
        \bottomrule
    \end{tabular}
\end{table}

\end{document}

Trajectory prediction is a time-sensitive task, as it may be required to handle rapidly changing scenes, such as downtown roads or highways, in a timely manner.
This section discusses the time/memory consumption of the proposed \MODELSHORT~models.
All these efficiency tests are conducted on an Apple Mac mini (M1, 8GB memory, 2020) to simulate low-power mobile environments, rather than high-performance GPU servers.

As reported in \TABLE{tab_efficiency}, \MODELSHORT~models present competitive computation efficiency.
The full model could forecast trajectories for up to about 5263 ego agents per second, with each agent having 20 stochastic trajectories predicted.
Meanwhile, its memory usage is still acceptable\footnote{
    Memory usage reported in \TABLE{tab_efficiency} only includes memory allocated by neural networks and variables, using the MPS-backend PyTorch environment.
}, even though there are two Transformer backbones used to compute latencies within non-interactive and social trajectory-changing events.

Notably, variations a6 and a7 show better efficiencies, forecasting up to about 22,222 agents per second with lower memory needed.
Although these variations may sacrifice some performance (about 2\%/5\% worse ADE/FDE on zara1) as a compromise, they offer alternatives for prediction scenarios with higher efficiency requirements, like some intelligent transportation systems and in-vehicle driving systems with limited computation resources and power delivery, demonstrating the efficiency of the proposed \MODELSHORT~models.

\end{document}

\section*{Authors' Note}
The opening quotations, ``\emph{To live is to imagine ourselves in the future, and there we inevitably arrive.}'' and ``\emph{Though the future is a product of every present that precedes it, tomorrow does not belong to today.}'' are adapted from Kobo Abe's \textit{In Search of the Present}. 
They resonate with the theme of this manuscript: forecasting as a process of bridging past experiences and imagined futures.

This study continues the conceptual thread of our previous works \SCMODEL~\cite{wong2023socialcircle,wong2024socialcircle+} and \REMODEL~\cite{wong2024resonance}, but it is not a direct extension of either of these papers.
Rather, it follows the same philosophical path that views trajectory prediction as a scan-reflect-echo process, as illustrated in the first figure of \SCMODEL~paper \SCCITE, which is a gradual experience from perception to resonance and then to reverberation.
In detail, \SCMODEL~regards that forecasting trajectories can be interpreted as \emph{echolocation}, \IE, scanning the surrounding space to perceive others.
\REMODEL~then emphasized how agents internally and socially reflect those perceived signals, aligning their \emph{spectrums} through mutual responses.
Building upon these ideas, \MODEL~turns to the final stage, modeling how such reflections persist, decay, and re-emerge as temporal echoes across time, where finally the echolocation finishes.
Through this trilogy, the focus of our research evolves from where interactions happen (\SCMODEL), to how they co-vibrate across time (\REMODEL), and finally to when and for how long their influences live (\MODEL), pondering the question: how perception, causality, and anticipation are intertwined.

\end{document}

\bibliographystyle{IEEEtran}
\bibliography{ref.bib}